\documentclass[10pt,twocolumn,letterpaper]{article}

\usepackage{iccv}
\usepackage{times}
\usepackage{epsfig}
\usepackage{graphicx}
\usepackage{amsmath}
\usepackage{amssymb}

\usepackage[dvipsnames]{xcolor}
\usepackage[shortlabels]{enumitem}
\usepackage{booktabs}
\usepackage{tablefootnote}

\usepackage[breaklinks=true,bookmarks=false,colorlinks]{hyperref}

\iccvfinalcopy %

\def\iccvPaperID{3787} %

\ificcvfinal\fi

\def\HideCuts{1} %
\def\HideTodo{0} %

\newcommand{\s}[1]{{\text{#1}}}

\newcommand{\iou}{\operatorname{IOU}}  %
\newcommand{\mean}{\operatorname{M}}  %

\newcommand{\todo}[1]{\if\HideTodo0{\color{red} [todo: #1]}\fi}
\newcommand{\cut}[1]{\if\HideCuts0{\color{gray} [cut (AB): #1]}\fi}

\begin{document}

\title{Local Metrics for Multi-Object Tracking}

\author{Jack Valmadre, Alex Bewley, Jonathan Huang, Chen Sun, Cristian Sminchisescu, Cordelia Schmid\\
Google Research
}

\maketitle
\ificcvfinal\thispagestyle{empty}\fi

\begin{abstract}

This paper introduces temporally local metrics for Multi-Object Tracking.
These metrics are obtained by restricting existing metrics based on track matching to a finite temporal horizon, and provide new insight into the ability of trackers to maintain identity over time.
Moreover, the horizon parameter offers a novel, meaningful mechanism by which to define the relative importance of detection and association, a common dilemma in applications where imperfect association is tolerable.
It is shown that the historical Average Tracking Accuracy (ATA) metric exhibits superior sensitivity to association, enabling its proposed local variant, ALTA, to capture a wide range of characteristics.
In particular, ALTA is better equipped to identify advances in association independent of detection.
The paper further presents an error decomposition for ATA that reveals the impact of four distinct error types and is equally applicable to ALTA.
The diagnostic capabilities of ALTA are demonstrated on the MOT 2017 and Waymo Open Dataset benchmarks.

\end{abstract}

\section{Introduction}

Multi-Object Tracking (MOT) algorithms aim to predict a set of tracks corresponding to the unique instances of a given class in a video.
Trackers must be evaluated for their ability to both detect and associate objects.\footnote{Even if two trackers start from an identical per-frame detector, their detection characteristics will depend on their use of temporal information.}
While detection errors can be measured by simply counting false positives and false negatives in each frame, there are multiple possible ways to treat association errors during evaluation.
Moreover, the significance of association errors depends on the application.
For example, in surveillance applications, an association error represents a critical failure, whereas for action recognition or trajectory forecasting, 
it may be sufficient for tracks to be only locally correct.

Existing MOT metrics can thus be divided into two distinct categories: \emph{strict metrics}, which require the association to be correct for the entire sequence, and \emph{non-strict metrics}, which seek to award \emph{partial credit} for tracks that are only correct within a subset of frames.
Strict metrics establish one-to-one correspondence between entire tracks, whereas non-strict metrics typically adopt a time-varying correspondence and use a soft association metric to quantify the confusion between tracks.
Examples of strict metrics are Identification $F_{1}$-score (IDF1)~\cite{ristani2016performance} and Average Tracking Accuracy (ATA)~\cite{manohar2006performance}, while examples of non-strict metrics are Multi-Object Tracking Accuracy (MOTA)~\cite{bernardin2008evaluating} and Higher Order Tracking Accuracy (HOTA)~\cite{luiten2020hota}.

\begin{figure}
\centering
\includegraphics[width=\columnwidth]{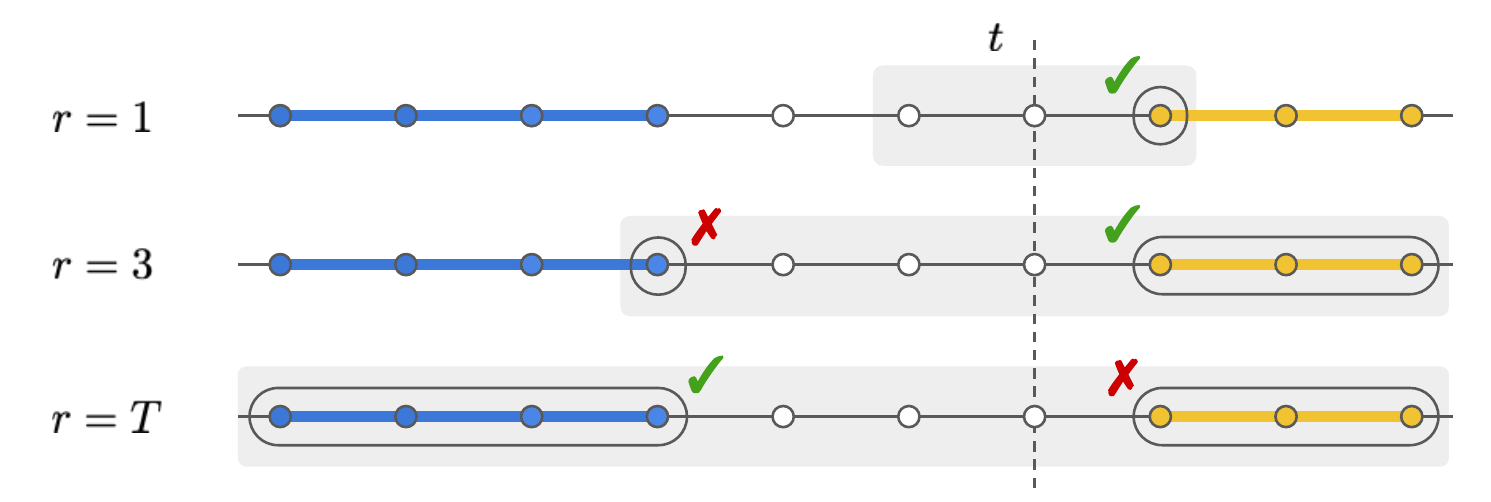}
\caption{The proposed local metrics establish one-to-one correspondence between tracks within a finite horizon~$r$ of each frame~$t$. Thin black lines represent ground-truth tracks while coloured lines denote distinct predicted tracks.}
\label{fig:overview}
\end{figure}

While the application governs which type of metric is suitable, each has associated drawbacks.
For non-strict metrics, it is unclear how best to define a scalar measure of association quality, leading to a proliferation of non-strict metrics~\cite{smith2005evaluating, bernardin2008evaluating, kao2009information, leichter2013monotonicity, luiten2020hota, feng2020samot}.
Moreover, when combined with a detection metric, each association metric induces an \emph{implicit} trade-off between detection and association which may not be well understood.
Strict metrics, on the other hand, generally lack error-type differentiability~\cite{leichter2013monotonicity} as a result of their monolithic track comparison. %
Furthermore, there is a distinct absence of temporal information in metrics of either type. %
In fact, with the exception of identity switches in MOTA~\cite{bernardin2008evaluating}, existing metrics are invariant to permutation of the frames, effectively evaluating the association of detections into tracks as a clustering problem.
This paper seeks to address these deficiencies through several contributions.

\textbf{Contribution 1.}
We introduce local metrics that generalise the existing strict metrics IDF1 and ATA.
The local metrics, designated LIDF1 and ALTA, are parametrised by a temporal horizon and thereby reveal the temporal ranges at which association errors occur,  see Figure \ref{fig:overview}.
Furthermore, the local metrics themselves constitute non-strict metrics which circumvent the need to choose a soft measure of track confusion and which possess an explicit, meaningful mechanism to set the balance between association and detection. %
The horizon(s) of interest depend on the application.
Empirically, we verify that the local metrics capture a wide range of trade-offs compared to existing metrics and show that ATA better represents association errors than IDF1.

\textbf{Contribution 2.}
We demonstrate that error-type differentiability can be achieved in ATA using independent per-frame correspondence.
The overall tracking error in ATA is approximately decomposed into four distinct error types. %
This is valuable both to developers and users of tracking algorithms, who need to understand how algorithms differ with respect to error modes.
The error decomposition applies equally well to the local metrics, revealing the shifting distribution of error types as the horizon is varied.

\textbf{Contribution 3.}
We use ALTA to re-evaluate the predictions of state-of-the-art trackers on the MOT 2017 and Waymo Open Dataset benchmarks.
Analysis of the metric at different horizons clearly shows that trackers which account for longer-term dependencies such as MAT~\cite{han2020mat} and Lif\_T~\cite{hornakova2020lifted} are better able to maintain identity over time.
Conversely, trackers which can only capture short-term dependencies such as TubeTK~\cite{pang2020tubetk} experience a rapid drop in accuracy as the horizon increases, despite achieving excellent detection accuracy.

\section{Existing metrics and related work}

Strict metrics establish one-to-one correspondence between ground-truth and predicted tracks, usually to maximise a measure of track overlap.
The two key strict metrics in the literature are IDF1~\cite{ristani2016performance} and ATA~\cite{manohar2006performance}.
The main difference is that IDF1 measures overlap by the number of frames in which two tracks overlap, whereas ATA uses spatiotemporal IOU (Intersection Over Union).
That is, IDF1 gives equal weight to each detection and ATA gives equal weight to each track.
These metrics will be described in greater detail in the next section.
One noted weakness of IDF1 is its invariance to the association of unmatched tracks~\cite{luiten2020hota, feng2020samot}.
While ATA seems to have been abandoned due to its inability to provide a separate measure of localisation error~\cite{stiefelhagen2006clear}, we will consider it as a viable alternative. %

Strict metrics have also been proposed for problems like video object detection, where a class must be predicted for each track.
Here, it is standard to use a binary criterion for \emph{track} overlap and obtain per-class precision-recall curves using confidence scores~\cite{ILSVRC16VID, kalogeiton2017action, dave2020tao}.
This approach focuses on detection rather than diagnosing tracking errors.  %

Non-strict metrics generally permit the correspondence between ground-truth and predicted tracks to vary over time and adopt some measure of confusion between tracks.
Since it is difficult to optimise for a time-varying correspondence, most approaches either use independent, per-frame correspondence or an approximate algorithm.

The most prevalent metric in recent history is MOTA~\cite{bernardin2008evaluating}, which quantifies association error using the number of identity switches. %
These are added to detection errors to obtain
\begin{equation}
\s{MOTA} = 1 - (\s{DetFN} + \s{DetFP} + \s{IDSw})/N
\end{equation}
where $N$ is the number of ground-truth boxes.
Amongst various issues with MOTA~\cite{luiten2020hota}, the most critical is that detection errors overwhelm identity switches, rendering it almost a pure detection metric~\cite{feng2019multi, maksai2019eliminating, hung2020soda, luiten2020hota}.
Weng \etal~\cite{weng20203d} consider the integral of MOTA with respect to recall by varying a per-box confidence threshold, however this remains primarily a measure of detection.
While improvement in detection can serve as a proxy for tracker performance~\cite{Dendorfer2020}, it ignores the quality of the tracks.

Several historical works have proposed alternative ways to quantify association error in a non-strict manner.
Smith \etal~\cite{smith2005evaluating} measured the average fraction of each ground-truth and predicted track covered by its best match, achieving partial credit by not enforcing one-to-one correspondence.
Kao \etal~\cite{kao2009information} quantified confusion using conditional entropy, similar to the $V$-measure in clustering~\cite{rosenberg2007v}.
Leichter and Krupka~\cite{leichter2013monotonicity} instead measured confusion between tracks using the fraction of frame pairs in each track with the same correspondence, similar to the Rand index in clustering~\cite{rand1971objective}.
More recently, Feng \etal~\cite{feng2020samot} quantified track confusion using the %
two-norm of overlap between all tracks.
The diversity of proposals illustrates the inherent subjectivity of quantifying association in MOT. %

Luiten \etal~\cite{luiten2020hota} recently proposed HOTA, which measures association accuracy using the average fraction of frames in the same track with the same association.
This is similar to the pair-based metric of Leichter and Krupka except that HOTA adopts a definition of association accuracy that includes false-positive and false-negative detections to ensure that adding a true-positive detection increases the tracking metric regardless of its association. %
The overall HOTA metric is obtained as the geometric mean of detection and association accuracy, averaged over multiple thresholds for spatial overlap.
Since it is computationally infeasible to find a time-varying correspondence which exactly maximises HOTA, a near-optimal correspondence is instead sought by maximising a surrogate objective.  %

\begin{figure}
\centering
\includegraphics[width=75mm]{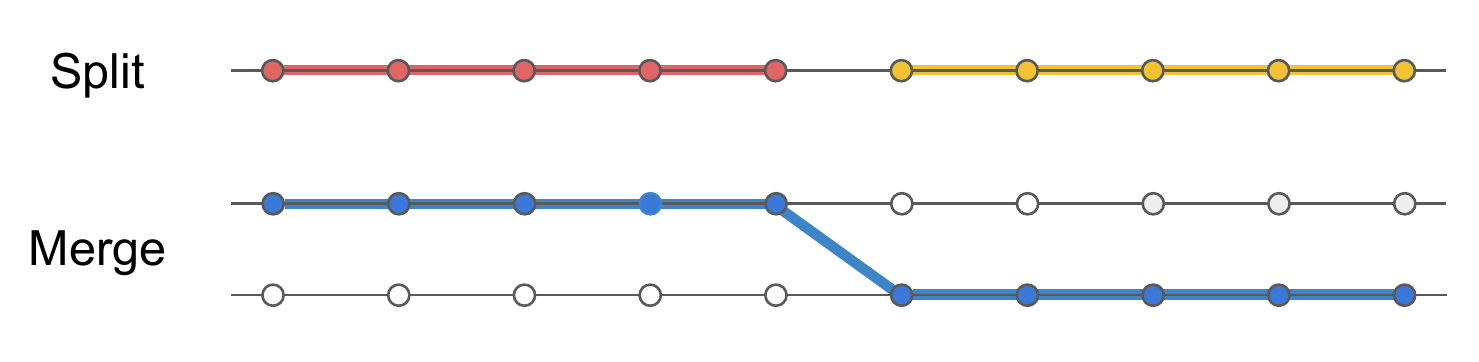}
\caption{
  Association errors can be divided into two types.
  \emph{Split} errors occur when one ground-truth track is associated to two predicted tracks (top) and, conversely, \emph{merge} errors occur when one predicted track is associated to two ground-truth tracks (bottom).
}
\label{fig:split_merge_error_types}
\end{figure}

Leichter and Krupka~\cite{leichter2013monotonicity} argued that metrics should possess error-type differentiability and enumerated the different error types in MOT.
Besides inaccurate localisation, these are characterised as under-detection (false negative), over-detection (false positive), under-association (split) and over-association (merge).
The two association errors are illustrated in Figure~\ref{fig:split_merge_error_types}.
The same authors proposed that metrics should be monotonic, meaning that eliminating one error without introducing another does not degrade the metric.
Unfortunately, this places no constraint on the metric's behaviour when a missed detection is corrected but the new detection is incorrectly associated.

Finally, Kim \etal~\cite{kim2020video} recently introduced a metric for panoptic video segmentation that evaluates tracks in a sliding window, similar to our proposed approach.
Since their primary focus is segmentation accuracy, they do not consider the role of association errors and restrict their analysis to a maximum window of 15 frames.

\section{Local metrics}

This section defines IDF1 and ATA in a common framework and develops the extension of each to finite horizons.

\subsection{Preliminaries}

Consider a video with $T$ frames and $K$ ground-truth tracks.
Let $t \in [T] \triangleq \{1, \dots, T\}$ denote the frames and $i \in [K]$ denote the ground-truth tracks.
For each ground-truth track~$i$, let $\mathcal{V}_{i} \subseteq [T]$ denote the subset of frames in which it is visible and let the rectangle $y_{i}(t)$ be its position in the image at each time $t \in \mathcal{V}_{i}$.
Similarly, let $\hat{K}$ denote the number of predicted tracks, $\hat{\mathcal{V}}_{j} \subseteq [T]$ denote the subset of frames in which predicted track~$j \in [\hat{K}]$ is visible and $\hat{y}_{j}(t)$ be its position at time~$t \in \hat{\mathcal{V}}_{j}$.

For each frame~$t$, let $B(t)$ denote a $K \times \hat{K}$ binary overlap matrix indicating the overlap of each ground-truth track~$i$ and predicted track~$j$
\begin{equation}
B_{i j}(t) = \begin{cases}
  1, & t \in \mathcal{V}_{i} \cap \hat{\mathcal{V}}_{j} \; \wedge \; \iou(y_{i}(t), \hat{y}_{j}(t)) \ge 0.5 \\
  0 & \text{otherwise.}
\end{cases} \nonumber
\end{equation}
The sequence of binary matrices $B(t)$ represents sufficient information to compute IDF1 and ATA.
The use of a per-frame, binary overlap criterion facilitates the study of detection and association errors as discrete events and ensures that the tracking metric is not affected by imprecise localisation.
If sensitivity to localisation is desired, the metric can be averaged over multiple thresholds, \eg~\cite{luiten2020hota}.

\subsection{Identification F1-score}

IDF1~\cite{ristani2016performance} establishes correspondence between entire tracks to maximise the total number of frames in which corresponding tracks overlap, referred to as the Identification True Positives (IDTP).
In our formulation, the number of frames in which each ground-truth track~$i$ and predicted track~$j$ overlap can be obtained by taking the sum of binary overlap matrices $B = \sum_{t = 1}^{T} B(t)$.

Let us represent the set of one-to-one matchings between the two sets of tracks by the binary matrices
\begin{equation}
\mathcal{M} = \big\{ A \in \{0, 1\}^{K \times \hat{K}} \mid A 1 \le 1, A^{T} 1 \le 1 \big\}
\end{equation}
where the element of~$A_{i j}$ indicates the selection of edge $(i, j)$ in the bipartite graph.
The maximum number of overlapping frames between matched tracks is then obtained by solving the linear assignment problem
\begin{equation}
\s{IDTP}
= \max_{A \in \mathcal{M}} \langle A, B \rangle
\end{equation}
where $\langle A, B \rangle = \operatorname{tr}(A^{T} B) = \sum_{i, j} A_{i j} B_{i j}$.
The $F_{1}$-score is defined in terms of the recall and precision of this subset
\begin{equation}
\textstyle \s{IDF1} = \mean_{-1}(\s{IDR}, \s{IDP}) = \s{IDTP} / [\frac{1}{2}(N + \hat{N})]
\end{equation}
where $N = \sum_{i} |\mathcal{V}_{i}|$ and $\hat{N} = \sum_{j} |\hat{\mathcal{V}}_{j}|$ denote the total number of boxes in the ground-truth and predicted tracks, the recall and precision are defined $\s{IDR} = \s{IDTP} / N$ and $\s{IDP} = \s{IDTP} / \hat{N}$ and $\operatorname{M}_{p}$ is the $p$-mean.

\subsection{Local Identification F1-score}

To generalise IDF1 to a local metric, we instead consider the identification true positives within a temporal interval $[a, b] \triangleq \{t \in [T] : a \le t \le b\}$ %
\begin{equation}
\s{IDTP}_{[a, b]} = \max_{A \in \mathcal{M}} \textstyle \Big\langle A, \sum_{t \in [a, b]} B(t) \Big\rangle \; .
\end{equation}
Importantly, each interval can adopt different correspondence between the two sets.
The terms $N$ and $\hat{N}$ can, likewise, be written as the sum of an indicator function $\mathbf{1}_{\mathcal{X}}(x) = 1[x \in \mathcal{X}]$ and restricted to the interval $[a, b]$
\begin{align}
N_{[a, b]} & = \textstyle \sum_{t \in [a, b]} \sum_{i} \mathbf{1}_{\mathcal{V}_{i}}(t)
\; .
\end{align}

To obtain a local metric with temporal horizon $r \ge 0$, we take the mean of these quantities over all intervals $[t - r, t + r]$ for $t \in [T]$
\begin{align}
\s{LIDF1}(r) & = \frac{\frac{1}{T} \sum_{t = 1}^{T} \s{IDTP}_{[t-r, t+r]}}{\frac{1}{2 T} \sum_{t = 1}^{T} \big(N_{[t-r, t+r]} + \hat{N}_{[t-r, t+r]}\big)} \; .
\end{align}
The mean numerator and denominator are used rather than the mean fraction to gracefully handle empty intervals.
The same approach of accumulating numerators and denominators is used to obtain the metrics for multiple sequences, where the factor $1 / T$ is important if there are sequences of different length.

\subsection{Average Tracking Accuracy}

ATA~\cite{manohar2006performance} is similar to IDF1, establishing per-video correspondence between entire tracks.
The key difference is that ATA measures the fraction of correct tracks.

Let $Q$ denote the temporal IOU matrix whose elements are the number of frames in which two tracks overlap divided by the number of frames in which either is present
\begin{equation}
Q_{i j} = B_{i j} / |\mathcal{V}_{i} \cup \hat{\mathcal{V}}_{j}| \; .
\end{equation}
The average track accuracy is then defined
\begin{equation}
\s{ATA} = \s{TrackTP} / [\tfrac{1}{2} (K + \hat{K})]
\end{equation}
where $\s{TrackTP} = \max_{A \in \mathcal{M}} \langle A, Q \rangle$ is the maximum sum of track overlap in a one-to-one correspondence between the two sets of tracks.
Similarly to IDF1, the metric can be understood as the harmonic mean of recall $\s{ATR} = \s{TrackTP} / K$ and precision $\s{ATP} = \s{TrackTP} / \hat{K}$. %

\subsection{Average Local Tracking Accuracy}

The local metric is similarly obtained by restricting the numerator and denominator to intervals with horizon~$r$ and taking the mean over intervals
\begin{align}
\s{ALTA}(r) = \frac
  {\frac{1}{T} \sum_{t = 1}^{T} \s{TrackTP}_{[t-r, t+r]}}
  {\frac{1}{2 T} \sum_{t = 1}^{T} \big(K_{[t-r, t+r]} + \hat{K}_{[t-r, t+r]}\big)} \; .
\end{align}
The interval-restricted numerator is obtained by finding the optimal correspondence between tracks within the interval
\begin{equation}
\s{TrackTP}_{[a, b]} = \max_{A \in \mathcal{M}} \langle A, Q_{[a, b]} \rangle
\end{equation}
where $Q_{[a, b]}$ is the temporal IOU matrix with elements
\begin{align}
\left(Q_{[a, b]}\right)_{i j} = \frac{ \sum_{t \in [a, b]} B_{i j}(t) }{ \sum_{t \in [a, b]} \mathbf{1}_{\mathcal{V}_{i} \cup \hat{\mathcal{V}}_{j}}(t) } \; .
\end{align}
The denominator terms $K_{[a, b]}$ and $\hat{K}_{[a, b]}$ are found by counting the number of tracks present within the interval.

\begin{figure}
\hypersetup{hidelinks}  %
\centering
\includegraphics[height=54mm]{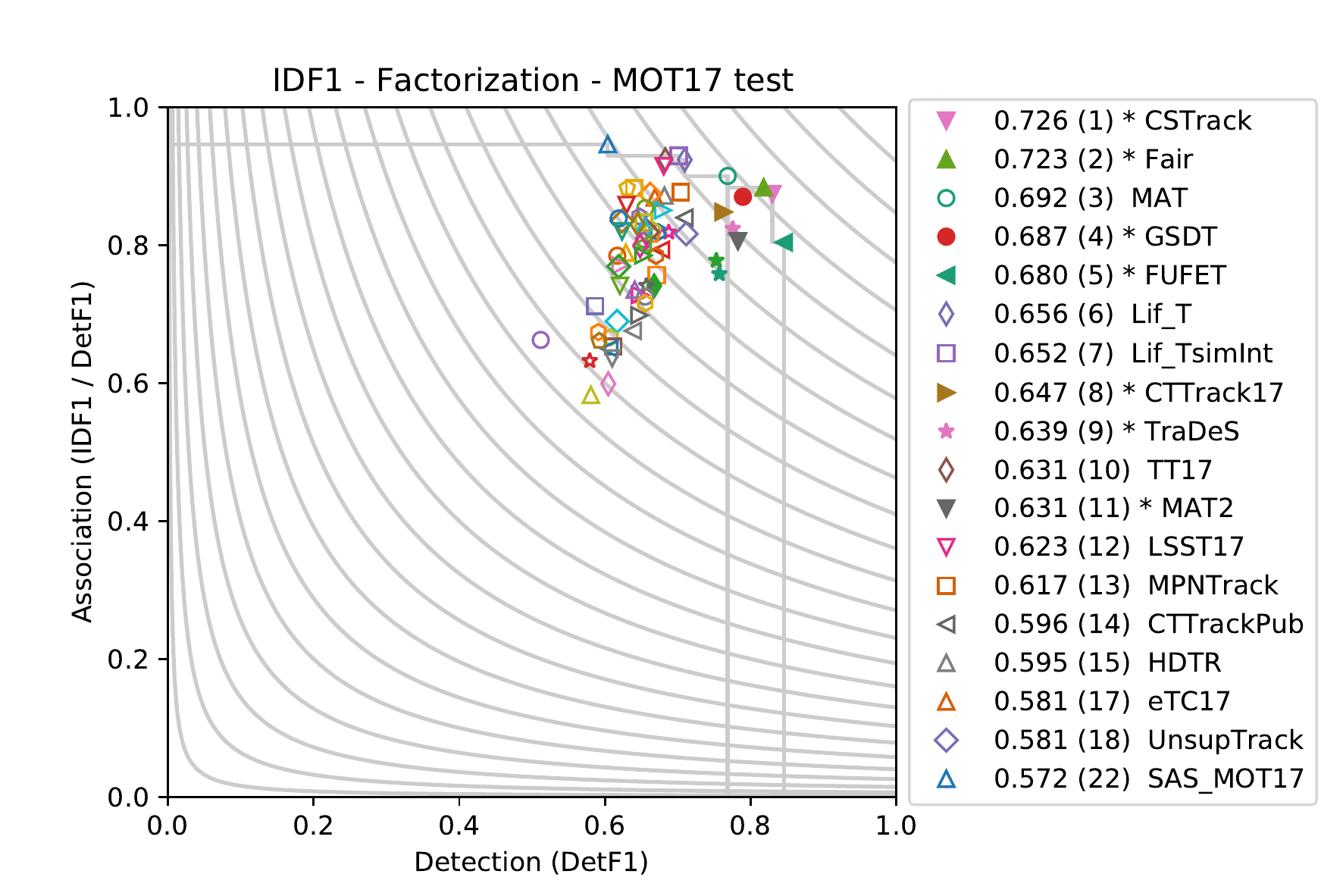}
\includegraphics[height=54mm]{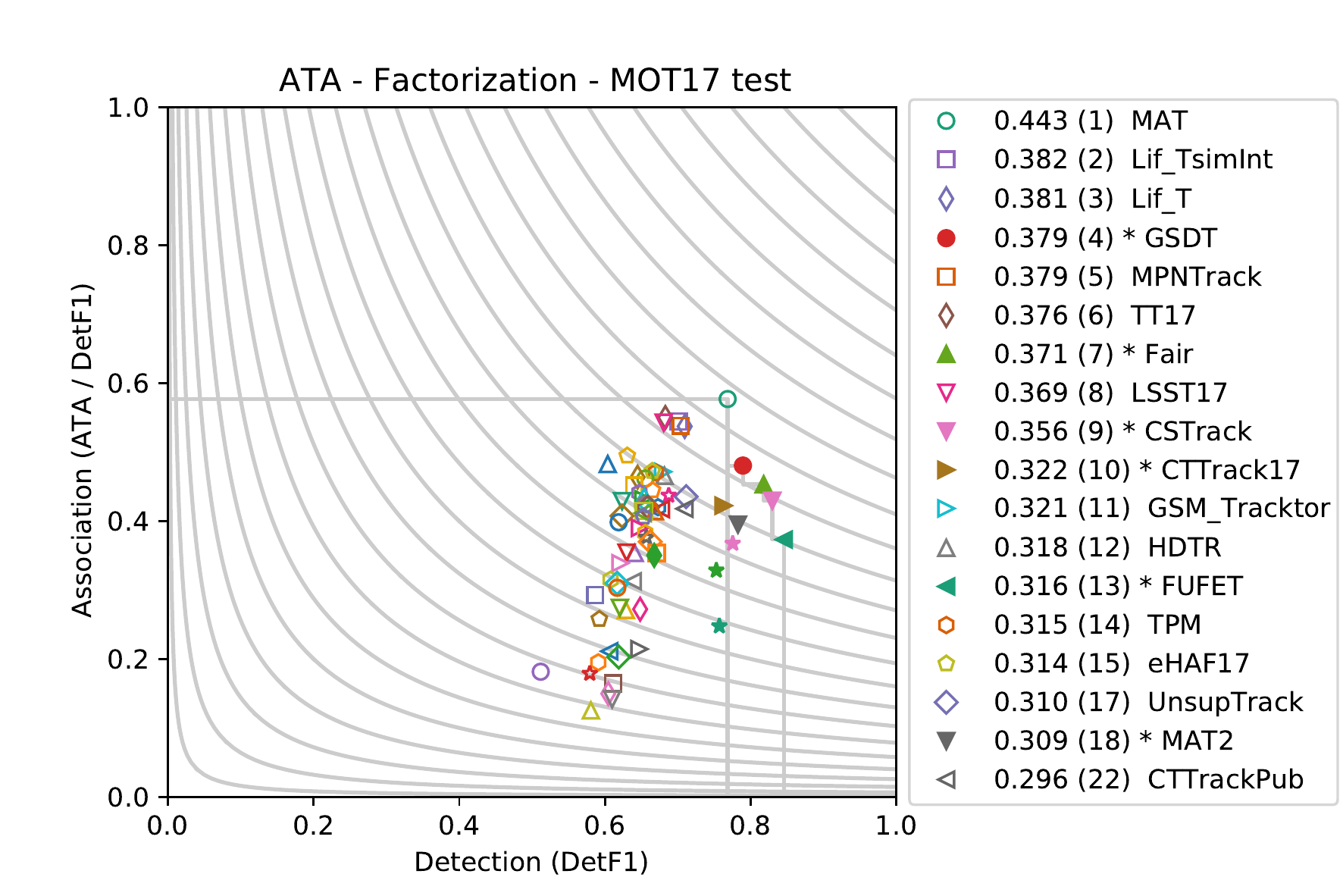}
\caption{
  Factorisation of IDF1 and ATA into detection and association for the 70 published submissions to the MOT 2017 Challenge.
  The implicit association metric of ATA has a much greater effect on the overall score, hence the use of a private detector (solid marker) provides a greater advantage under IDF1.
  The Pareto front is shown separately for all methods and for those that use public detections.
  The legend lists the union of top ten trackers and top five per dimension.
  The curves depict level sets of the metric.
}
\label{fig:factorize}
\end{figure}

\begin{figure*}
\centering
\includegraphics[height=48mm]{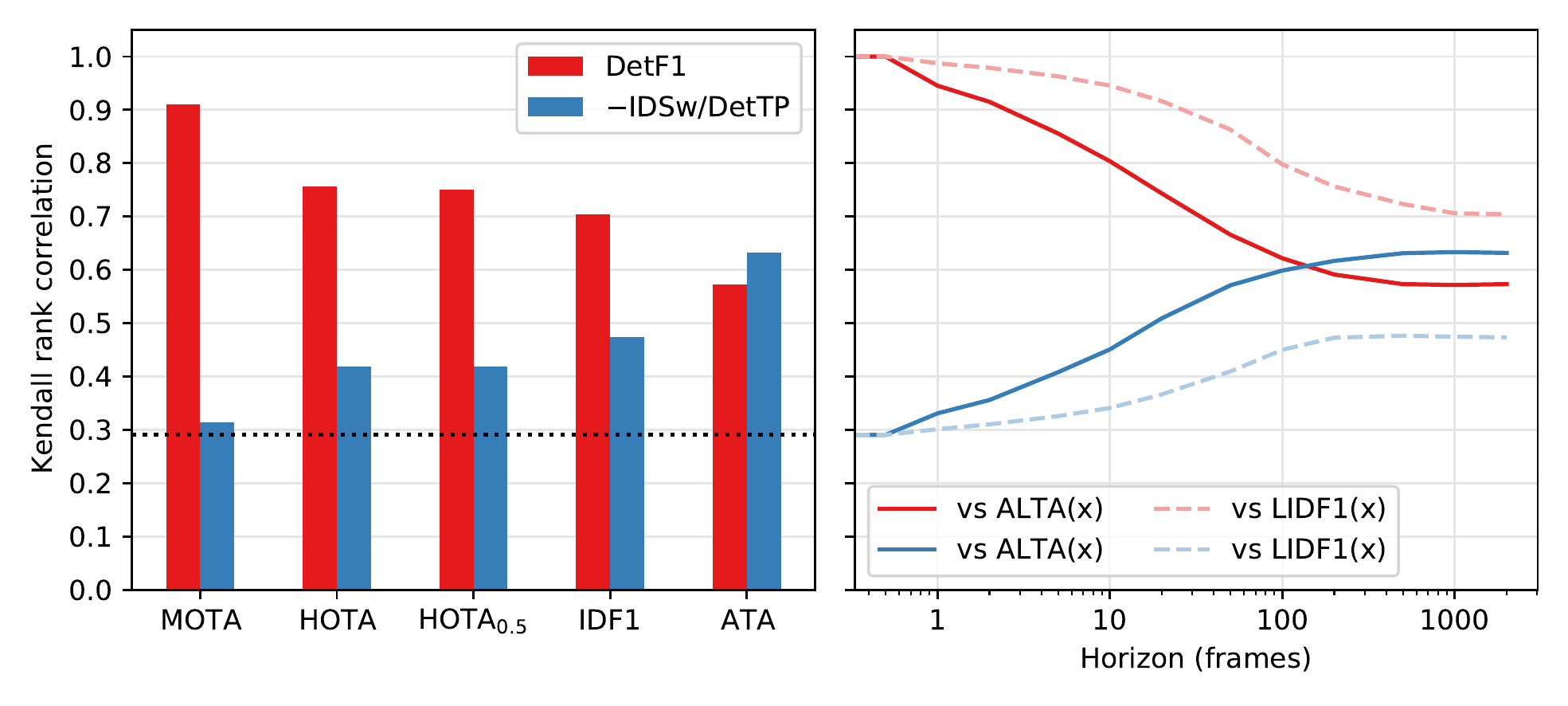}
\includegraphics[height=48mm]{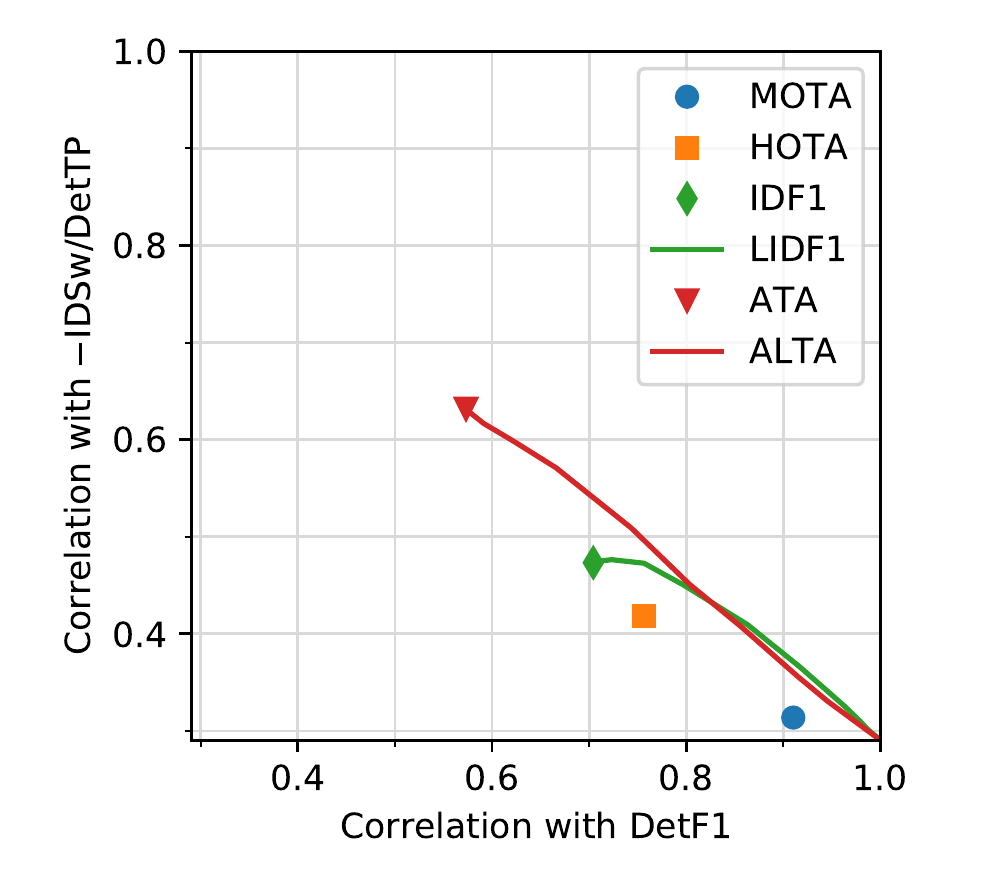}
\caption{
   \textbf{Left:}~Rank correlation of tracking metrics with detection $F_{1}$-score and normalised identity switches based on the 125 submissions to the MOT 2017 Challenge~\cite{MOT16}.
  ATA is more correlated with association and less correlated with detection than IDF1, HOTA and MOTA.
  The black dotted line illustrates the ``baseline'' correlation between detection $F_{1}$-score and identity switches.
  \textbf{Middle:}~Our local versions of IDF1 and ATA trace a spectrum between pure detection towards the strict tracking metrics as the temporal horizon increases.
  \textbf{Right:}~The local metrics trace a path in the detection-association spectrum while other metrics correspond to a particular trade-off.
}
\label{fig:rank-corr}
\end{figure*}

\subsection{Properties of local metrics}

It is straightforward to verify that, for $r \ge T - 1$, each local metric reduces to its strict counterpart, $\s{LIDF1}(r) = \s{IDF1}$ and $\s{ALTA}(r) = \s{ATA}$.
Conversely, for $r < 1$ (single-frame windows), both reduce to the detection $F_{1}$-score
\begin{equation}
\s{LIDF1}(r) = \s{ALTA}(r) = \s{DetF1} \triangleq \frac{\tfrac{1}{T} \, \s{DetTP}}{\tfrac{1}{2 T} (N + \hat{N})} \; .
\end{equation}
The detection metric is an upper bound of the local metrics $\s{DetF1} \ge \s{IDF1}(r), \s{ALTA}(r)$ since it can choose the optimal correspondence independently for each frame.
This means that comparing the local tracking metric to the detection metric reveals the impact of association errors.

Computing the local metrics requires solving $\mathcal{O}(T)$ bipartite matching problems per horizon.
Assuming that $\hat{K}$ is $\mathcal{O}(K)$, each can be solved in $\mathcal{O}(K^3)$ by the Hungarian algorithm~\cite{kuhn1955hungarian} and the cost matrix of each problem can be constructed in $\mathcal{O}(K^{2})$ time using summed area tables, which themselves are obtained in $\mathcal{O}(T K^2)$ time total.
Therefore, evaluating the local metric for $H$ different horizons requires $\mathcal{O}(H T K^{3})$ time, roughly $H$ times as expensive as MOTA and HOTA, which also solve an assignment problem per frame.
One advantage of the local metrics is that they use the optimal correspondence, whereas MOTA and HOTA depend on approximations.
Furthermore, there may exist a more efficient algorithm (like Toroslu and {\"U}{\c{c}}oluk~\cite{toroslu2007incremental}) which re-use effort from interval $t$ to solve interval $t + 1$, seeing as the edge weights change gradually.

\section{Comparative study of metrics}

This section studies the relative impact of detection and association on existing metrics and the proposed local metrics.
It will be established that, for trackers submitted to the MOT 2017 Challenge, association plays a more significant role in ATA than in IDF1.
The range of trade-offs captured by the local metrics will further be characterised relative to existing non-strict metrics.

As described in the previous section, the impact of association on ATA and IDF1 can be measured by considering each as a fraction of DetF1, corresponding to single-frame windows.
Figure~\ref{fig:factorize} presents the factorisation of IDF1 and ATA into DetF1 and this association fraction for submissions to the MOT 2017 Challenge~\cite{MOT16}.
Association has a greater impact on ATA than on IDF1, since the tracking metric is a smaller fraction of the detection metric.
Notably, four of the top five trackers under IDF1 make use of an external detector, whereas this statistic is reversed under ATA.

To characterise the degree to which different overall metrics capture detection and association, we compare the ranking that each metric induces on the submissions to the MOT 2017 Challenge~\cite{MOT16}.
Specifically, in Figure~\ref{fig:rank-corr}, we measure the Kendall rank correlation of each metric with (i) the detection $F_{1}$-score and (ii) the rate of identity switches per correct detection\footnote{We do not argue that the number of identity switches is the best way to quantify association error, we simply adopt it as an established reference. We normalise by the number of true-positive detections since trackers that detect more objects have greater potential for identity switches.}.
It is immediately apparent that ATA exhibits the strongest correlation with identity switches and the least with detection.
Conversely, MOTA is highly correlated with detection and barely represents identity switches better than the detection metric itself (the dotted line).
While IDF1 and ATA are both strict metrics, IDF1 exhibits greater dependence on detection.
HOTA is somewhat less strict than IDF1 with respect to association, and the integral over localisation thresholds has little effect on these particular correlation measures.
The center and right plots of Figure~\ref{fig:rank-corr} mark the range of metrics that are attained by varying the horizon in our local variants.
ALTA achieves a greater range of characteristics than LIDF1. %

Given that (i) we are most interested in track accuracy and (ii) ALTA can express a greater range of trade-offs, %
we will concentrate on ALTA for the remainder of the paper.
LIDF1 may still be preferred for applications where short, incorrect tracks are of minor importance.

\section{Error type decomposition}

To determine a tracker's suitability for an application, it is crucial to have transparency with respect to different error types.
Here we develop a decomposition of ATA (and by extension ALTA) error into the two types of detection error (false positive and false negative) and the two types of association error (split and merge), see Figure~\ref{fig:split_merge_error_types}).

The fundamental idea behind the decomposition is to use a per-frame correspondence to categorise the different errors that occur within each track.
Let $C(t) \in \mathcal{M}$ denote the independent, per-frame correspondence which is used to measure detection error.
We can obtain a good approximation to ATA by replacing the number of frames in which two tracks overlap, $B$, with the number in which they are matched in the per-frame correspondence, $C = \sum_{t = 1}^{T} C(t)$,
\begin{equation}
\widetilde{\s{TrackTP}} = \max_{A \in \mathcal{M}} \langle A, \widetilde{Q} \rangle \; , \quad
\widetilde{Q}_{i j} = \frac{C_{i j}}{|\mathcal{V}_{i} \cup \hat{\mathcal{V}}_{j}|} \; .
\label{eq:approx-track-tp}
\end{equation}
This ensures per-frame exclusivity and is a good approximation since the majority of detections have only a single candidate with $\iou \ge 0.5$.

We now present the error decomposition for track recall using this approximation.
For track precision, error types are simply exchanged.
Let $\pi(i)$ denote the corresponding predicted track for each ground-truth track~$i$ in the optimal matching~$A^{\star}$ from eq.~\ref{eq:approx-track-tp}.
We rewrite the recall as the mean temporal IOU of ground-truth track~$i$ with its partner~$\pi(i)$
\begin{align}
\widetilde{\s{ATR}}
= \frac{1}{K} \sum_{i = 1}^{K} \widetilde{Q}_{i, \pi(i)} \; , \quad
\widetilde{Q}_{i, \pi(i)} = \frac{C_{i, \pi(i)}}{| \mathcal{V}_{i} \cup \hat{\mathcal{V}}_{\pi(i)} |}
\end{align}
If track~$i$ is unmatched, we set $\hat{\mathcal{V}}_{\pi(i)} = \emptyset$ and $C_{i, \pi(i)} = 0$.
The recall error is then equal to the mean per-track error,
\begin{align}
1 - \widetilde{\s{ATR}}
= \textstyle \frac{1}{K} \sum_{i = 1}^{K} \big(1 - \widetilde{Q}_{i, \pi(i)}\big) \enspace .
\end{align}
To decompose the error for track~$i$, we consider the fraction of each track which is covered by its assigned partner, by its best possible partner, and by any other track.
These define a series of upper bounds on $\widetilde{Q}_{i, \pi(i)} =$
\begin{equation}
\frac{C_{i, \pi(i)}}{|\mathcal{V}_{i} \cup \hat{\mathcal{V}}_{\pi(i)}|}
\le \underbrace{ \frac{C_{i, \pi(i)}}{|\mathcal{V}_{i}|} }_{\rho^{\pi}_{i}}
\le \underbrace{ \frac{\max_{j} C_{i j}}{|\mathcal{V}_{i}|} }_{\rho^{\s{best}}_{i}}
\le \underbrace{ \frac{\sum_{j} C_{i j}}{|\mathcal{V}_{i}|} }_{\rho^{\s{det}}_{i}}
\le 1 \; .
\nonumber
\end{equation}
The gaps between these inequalities provide a decomposition
$1 - \widetilde{Q}_{i, \pi(i)} = (1 - \rho^{\s{det}}_{i}) +  (\rho^{\s{det}}_{i} - \rho^{\s{best}}_{i}) + (\rho^{\s{best}}_{i} - \rho^{\pi}_{i}) + (\rho^{\pi}_{i} - \widetilde{Q}_{i, \pi(i)})$
where each term is caused by a specific error type.
Specifically, within ground-truth track~$i$,
\begin{enumerate}[1.,itemsep=4pt,topsep=4pt,parsep=0pt,partopsep=0pt,leftmargin=1.5em]
\item $(1 - \rho^{\s{det}}_{i})$ represents frames that were not matched to any predicted track, \ie false-negative detection errors,
\item $(\rho^{\s{det}}_{i} - \rho^{\s{best}}_{i})$ represents frames which were matched to some track other than the best possible partner, which only occurs if the ground-truth track~$i$ was matched to multiple predicted tracks, \ie split errors, and
\item $(\rho^{\s{best}}_{i} - \rho^{\pi}_{i})$ is the difference between the best possible partner and the actual assigned partner, which only exists if the best partner was matched to multiple ground-truth tracks, \ie merge errors.
\end{enumerate}
The remaining gap, due to the union denominator, is
\begin{align}
\rho_{i}^{\pi} - \widetilde{Q}_{i, \pi(i)}
& = \Big( 1 - \tfrac{|\mathcal{V}_{i}|}{| \mathcal{V}_{i} \cup \hat{\mathcal{V}}_{\pi(i)} |} \Big) \tfrac{C_{i, \pi(i)}}{|\mathcal{V}_{i}|}
\end{align}
and represents the frames $\hat{\mathcal{V}}_{\pi(i)} \setminus \mathcal{V}_{i}$ in which the matching predicted track~$\pi(i)$ is present but track~$i$ itself is not.
These frames of the predicted track can be partitioned into those which are not associated with any ground-truth track (\ie false-positive detection errors) and those which are associated to a different ground-truth track (\ie track merge errors), thus completing the decomposition.
To obtain a decomposition of the overall error, we note
\begin{align}
& 1 - \s{ATA}
= \textstyle \frac{1}{K + \hat{K}} [K (1 - \s{ATR}) + \hat{K}(1 - \s{ATP} )] \nonumber \\
& = \textstyle \frac{1}{K + \hat{K}} [ \sum_{i} (1 - Q_{i, \pi(i)}) + \sum_{j} (1 - Q_{\pi^{-1}(j), j}) ]
\label{eq:overall-error-decomposition}
\end{align}
and then use the approximate decompositions of each error.

Figure~\ref{fig:ata_purity_coverage} illustrates the decomposition of recall and precision for several trackers from the MOT 2017 Challenge. %

\section{Tracker analysis with ALTA}

To demonstrate the utility of ALTA and its error decomposition we compare a diverse set of state-of-the-art trackers from the MOT 2017 Challenge~\cite{MOT16} and Waymo Open Dataset~\cite{sun2020scalability} tracking benchmark. %
Metrics for the private test sets were obtained by providing code to the dataset maintainers.
Extended results can be found in the appendix.  %

\begin{figure}
\centering
\includegraphics[height=60mm]{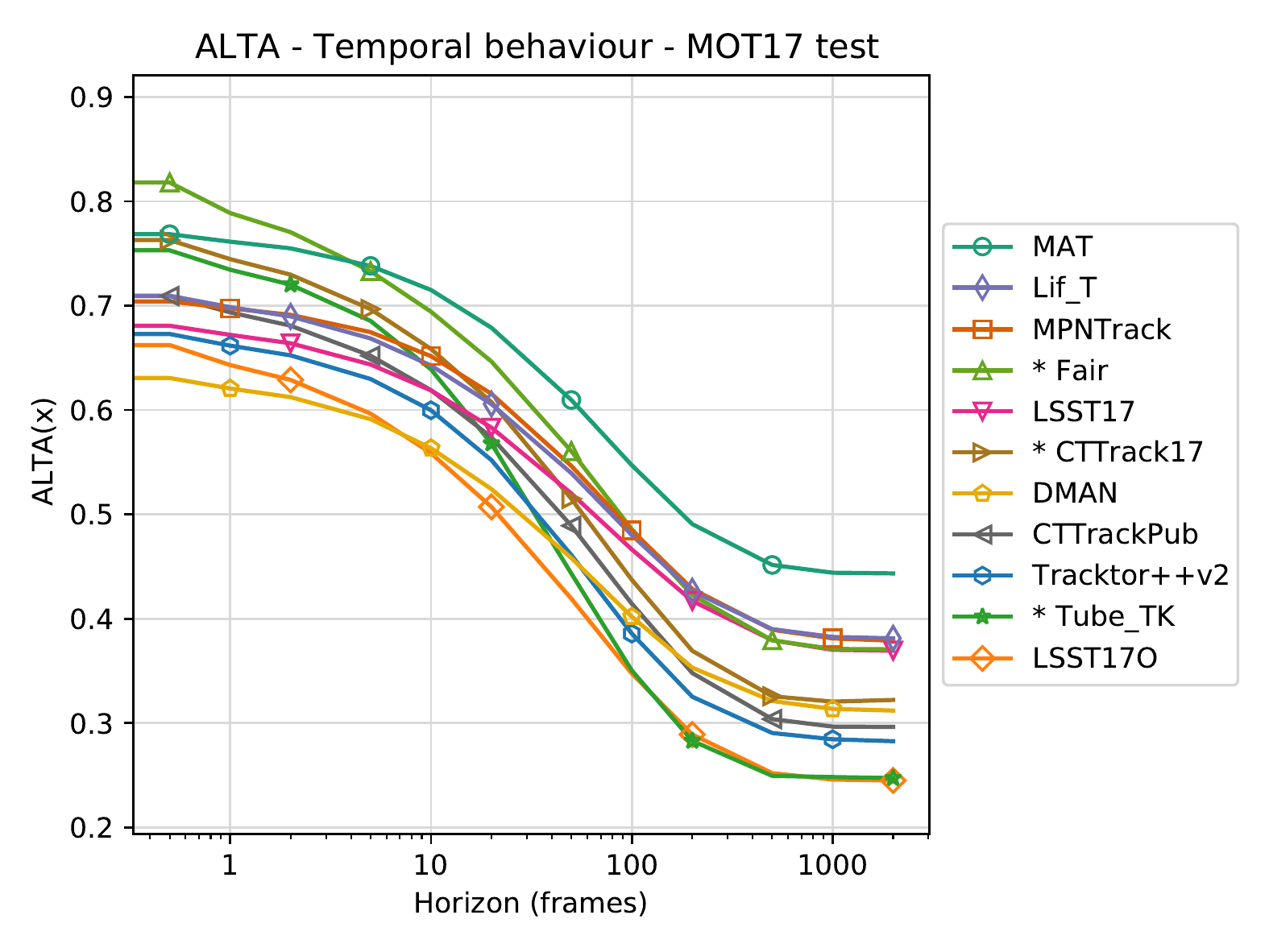}
\caption{ALTA versus the horizon parameter for high-performing trackers on the MOT 2017 Challenge test set.
The * denotes trackers that use detections other than provided by the benchmark.}
\label{fig:alta}
\end{figure}

\begin{table*}
\centering
\caption{
  Evaluation of representative trackers in the MOT 2017 Challenge using key existing metrics and ALTA at horizons of one and five seconds.
  Asterisks * denote the use of a detector besides that provided by the benchmark. 
  The rank within each metric is indicated in parentheses with the top five in bold.
  The mean of three horizons is adopted as an illustrative example of a general purpose metric.
}
\scalebox{0.8}{\begin{tabular}{l r@{~~}l r@{~~}l r@{~~}l r@{~~}l r@{~~}l | r@{~~}l r@{~~}l r@{~~}l r@{~~}l}
\toprule
Tracker & \multicolumn{2}{c}{DetF1} & \multicolumn{2}{c}{MOTA} & \multicolumn{2}{c}{HOTA} & \multicolumn{2}{c}{HOTA$_{0.5}$} & \multicolumn{2}{c |}{IDF1} & \multicolumn{2}{c}{ALTA(1s)} & \multicolumn{2}{c}{ALTA(5s)} & \multicolumn{2}{c}{ATA} & \multicolumn{2}{c}{$\uparrow$ mean} \\ %
\midrule
                    MAT \cite{han2020mat} & .769 & {\bf (2)} &  .671 & {\bf (3)} &        .560 &       {\bf (2)} &       .677 &      {\bf (2)} &  .692 & {\bf (2)} &     .660 &    {\bf (1)} &     .520 &    {\bf (1)} & .443 & {\bf (1)} &   .541 &  {\bf (1)} \\
           * Fair \cite{zhang2020fairmot} & .818 & {\bf (1)} &  .737 & {\bf (1)} &        .593 &       {\bf (1)} &       .719 &      {\bf (1)} &  .723 & {\bf (1)} &     .623 &    {\bf (2)} &     .455 &    {\bf (3)} & .371 & {\bf (4)} &   .483 &  {\bf (2)} \\
        MPNTrack \cite{braso2020learning} & .704 &       (7) &  .588 &       (7) &        .490 &       {\bf (5)} &       .603 &      {\bf (5)} &  .617 &       (6) &     .596 &    {\bf (3)} &     .456 &    {\bf (2)} & .379 & {\bf (3)} &   .477 &  {\bf (3)} \\
        Lif\_T \cite{hornakova2020lifted} & .710 & {\bf (5)} &  .605 &       (6) &        .513 &       {\bf (4)} &       .638 &      {\bf (4)} &  .656 & {\bf (3)} &     .585 &    {\bf (4)} &     .451 &    {\bf (4)} & .381 & {\bf (2)} &   .473 &  {\bf (4)} \\
              LSST17 \cite{feng2019multi} & .681 &       (8) &  .547 &       (9) &        .471 &             (8) &       .589 &            (7) &  .623 & {\bf (5)} &     .565 &          (6) &     .447 &    {\bf (5)} & .369 & {\bf (5)} &   .461 &  {\bf (5)} \\
      * CTTrack17 \cite{zhou2020tracking} & .763 & {\bf (3)} &  .678 & {\bf (2)} &        .522 &       {\bf (3)} &       .651 &      {\bf (3)} &  .647 & {\bf (4)} &     .583 &    {\bf (5)} &     .406 &          (6) & .322 &       (6) &   .437 &        (6) \\
       CTTrackPub \cite{zhou2020tracking} & .709 &       (6) &  .615 & {\bf (5)} &        .482 &             (6) &       .595 &            (6) &  .596 &       (7) &     .552 &          (7) &     .385 &          (7) & .296 &       (8) &   .411 &        (7) \\
                DMAN \cite{zhu2018online} & .631 &      (11) &  .482 &      (11) &        .425 &            (11) &       .532 &           (11) &  .557 &      (10) &     .506 &         (10) &     .378 &          (8) & .312 &       (7) &   .399 &        (8) \\
 Tracktor++v2 \cite{bergmann2019tracking} & .673 &       (9) &  .563 &       (8) &        .448 &             (9) &       .550 &           (10) &  .551 &      (11) &     .525 &          (9) &     .351 &          (9) & .283 &       (9) &   .386 &        (9) \\
         * Tube\_TK \cite{pang2020tubetk} & .753 & {\bf (4)} &  .630 & {\bf (4)} &        .480 &             (7) &       .586 &            (8) &  .586 &       (8) &     .529 &          (8) &     .312 &         (11) & .247 &      (10) &   .363 &       (10) \\
             LSST17O \cite{feng2019multi} & .662 &      (10) &  .527 &      (10) &        .443 &            (10) &       .554 &            (9) &  .579 &       (9) &     .479 &         (11) &     .315 &         (10) & .245 &      (11) &   .346 &       (11) \\

\bottomrule
\end{tabular}}
\label{tab:compare}
\end{table*}

\subsection{MOT 2017 Challenge}

We briefly describe the trackers considered.
\textbf{Tracktor++v2}~\cite{bergmann2019tracking} uses the second stage of Faster R-CNN~\cite{ren2015faster} to find the new location of each track, compensating for camera motion and performing re-ID with a Siamese network, subject to a strict overlap criterion.
\textbf{LSST17}~\cite{feng2019multi} combines a single-object tracker~\cite{li2018high}, a deep re-ID embedding and a tree-based, ``switcher-aware'' classifier.
The default variant uses agglomerative clustering to construct tracks and the online variant, \textbf{LSST17O}, does not.
\textbf{DMAN}~\cite{zhu2018online} extends a single-object tracker~\cite{danelljan2017eco} using a Siamese network and a bi-directional LSTM to compare detections to tracklets.
\textbf{CTTrack17} is CenterTrack~\cite{zhou2019objects}, which performs auto-regressive offset regression, taking the previous image and prediction as input.
\textbf{CTTrackPub} is a variant constrained to only start tracks using the standard, public detections.
\textbf{Fair}~\cite{zhang2020fairmot} is an anchor-free CNN detector that is jointly trained to predict instance embeddings.
\textbf{TubeTK}~\cite{pang2020tubetk} uses a CNN to predict tubes as temporal splines and then links these tubes using their overlap and velocity.
\textbf{MPNTrack}~\cite{braso2020learning} trains a Message Passing Network to operate on the graph of detections and predict the validity of putative edges.
\textbf{Lif\_T}~\cite{hornakova2020lifted} adds long-term cues to existing paths in a network flow formulation with simple motion-based features, with Tracktor-style interpolation~\cite{bergmann2019tracking} as post-processing.
\textbf{MAT}~\cite{han2020mat} (Motion Aware Tracker) is an enhanced SORT~\cite{bewley2016simple}, accounting for camera motion and permitting re-association at greater temporal distances. %

Figure~\ref{fig:alta} shows ALTA as a function of horizon.
As the horizon increases, the ordering of the trackers undergoes significant change, as some are much better at associating tracks over greater temporal distances. %
Most of the trackers with the best detection accuracy experience a severe drop in ALTA as the horizon increases, especially CTTrack17 and TubeTK, which lack a mechanism to perform longer-term association.
This suggests that powerful detectors have effectively compensated for poor association in the past, whereas ATA makes this more difficult.
One exception to this trend is MAT~\cite{han2020mat}, which has good detection accuracy while still associating tracks robustly over longer time-scales by extrapolating tracks with a Kalman filter and camera motion compensation. %
MAT uses the public set of detections and, surprisingly, does not use appearance features in re-association.
On the other hand, some trackers with only reasonable detection accuracy are particularly effective at maintaining association over time, in particular MPNTrack, LSST17 and Lif\_T.
These trackers all incorporate longer-term cues for association.
DMAN also achieves reasonable ATA despite starting with particularly poor detection accuracy.
Comparing LSST17 to its online variant LSST17O, which does not use agglomerative clustering, reveals a marked difference in association quality over time.

Table~\ref{tab:compare} compares DetF1, MOTA, HOTA and IDF1 to ATA and ALTA at one and five second horizons.
While we do not claim to propose a universal scalar metric, since no single metric will be suitable for all applications, we order trackers in this table by the mean ALTA at three horizons as an illustrative example of a general purpose metric.
One advantage of a mean ALTA metric is that the contribution of each time-scale is known.
The trackers which best maintain associations over longer time intervals, MAT, MPNTrack, and Lif\_T, %
are ranked higher under ATA than other metrics.
The methods which depend on detection, Fair, CTTrack17 and Tube\_TK, drop in the ranking under ALTA.

\begin{figure}
\centering
\makebox[\columnwidth][c]{\includegraphics[width=1.05\columnwidth]{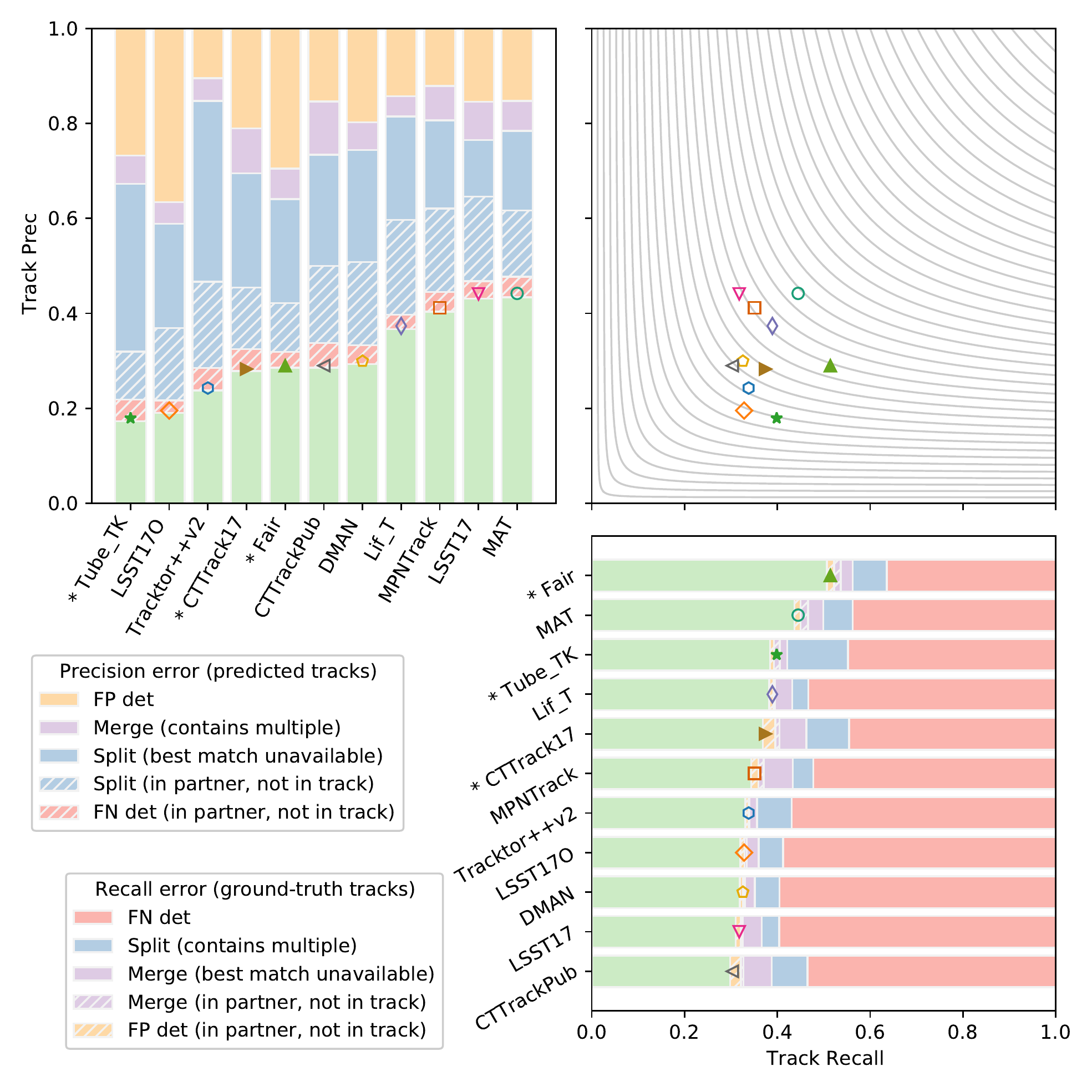}}
\caption{
    ATA is computed as the harmonic mean (top right) of track precision and track recall.
    The bar graphs show the decomposition of precision error (top left) and recall error (bottom right). %
}
\label{fig:ata_purity_coverage}
\end{figure}

Figure~\ref{fig:ata_purity_coverage} depicts the error decomposition of precision and recall. %
This reveals that, for these trackers, recall is dominated by false-negative detections, while precision is most affected by split errors.
Split errors affect the precision both in that (a) the best match for a predicted track is often unavailable in a one-to-one correspondence and (b) the ground-truth track with which a predicted track is matched often contains frames where the predicted track is not present.
This is indicative of over-segmentation of the ground-truth track into multiple predictions.
Examining individual trackers, we first observe that Fair achieves excellent recall by reducing false-negative detections, %
however its precision suffers from a significant %
fraction of false-positive detections.
While Tube\_TK and CTTrack17 also reduce the rate of false-negative detections using an external detector, their recall is significantly affected by association errors.
The trackers with the best track precision are MAT, LSST17 and MPNTrack, while only MAT achieves this at a high recall.
Despite having the least false-positive detections, the precision of Tracktor++v2 is limited by its high prevalence of split errors.

\begin{figure}
\centering
\includegraphics[height=50mm]{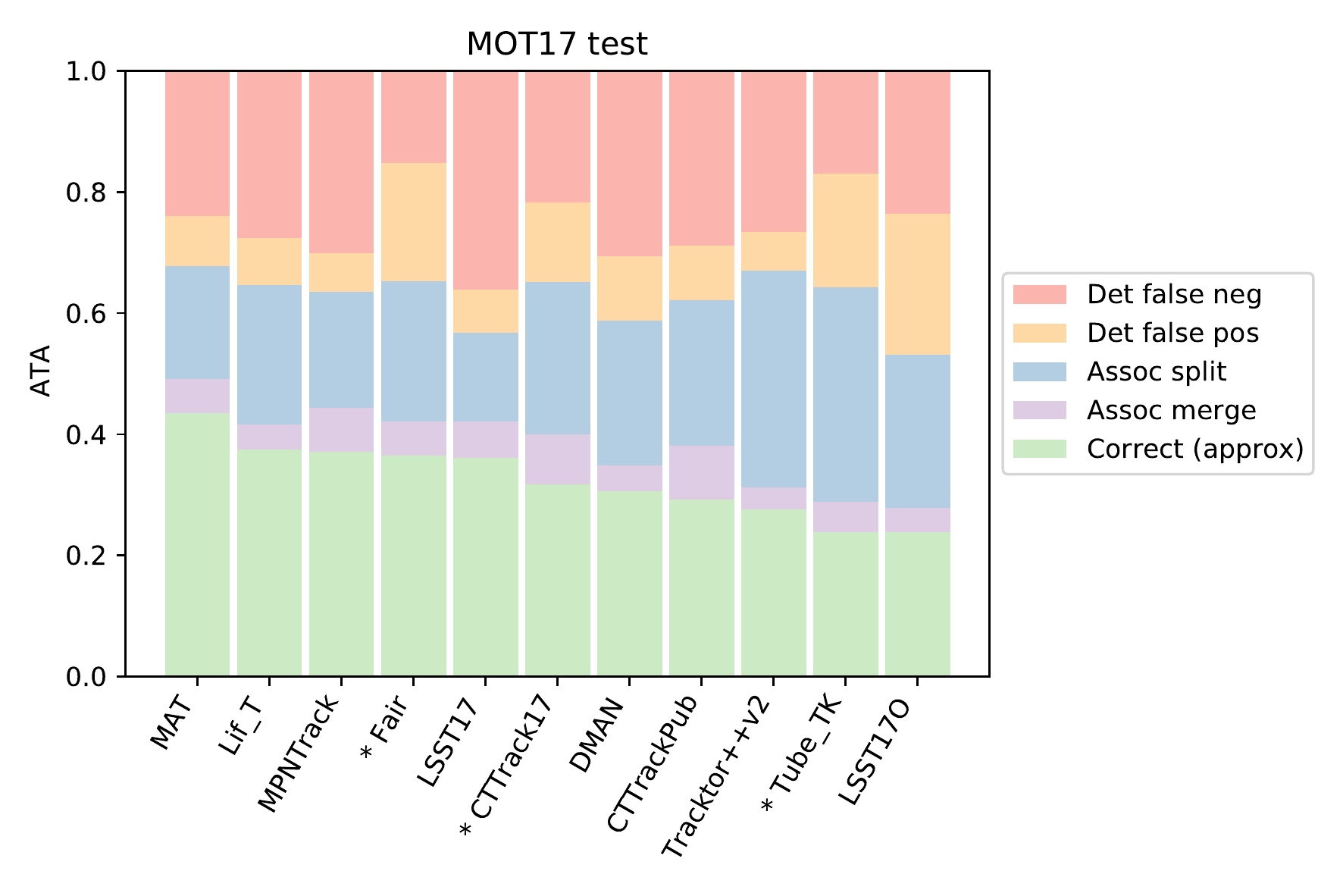}
\caption{The relative impact of each error type on ATA per tracker.}
\label{fig:decompose}
\end{figure}

Figure~\ref{fig:decompose} shows the relative contribution of each error type to the overall score.
This reveals that split errors account for a much larger component than merge errors and that the trackers with the highest ATA generally have the smallest association error component.
LSST17 in particular achieves an impressive ATA score given the large contribution of false-negative detection errors.
Note that eq.~\ref{eq:overall-error-decomposition} gives equal weight to predicted and ground-truth tracks, therefore the relative importance of precision and recall in the overall decomposition depends on the number of predicted tracks.

\begin{figure}
\centering
\includegraphics[width=40mm,trim={5mm 0mm 5mm 0mm},clip]{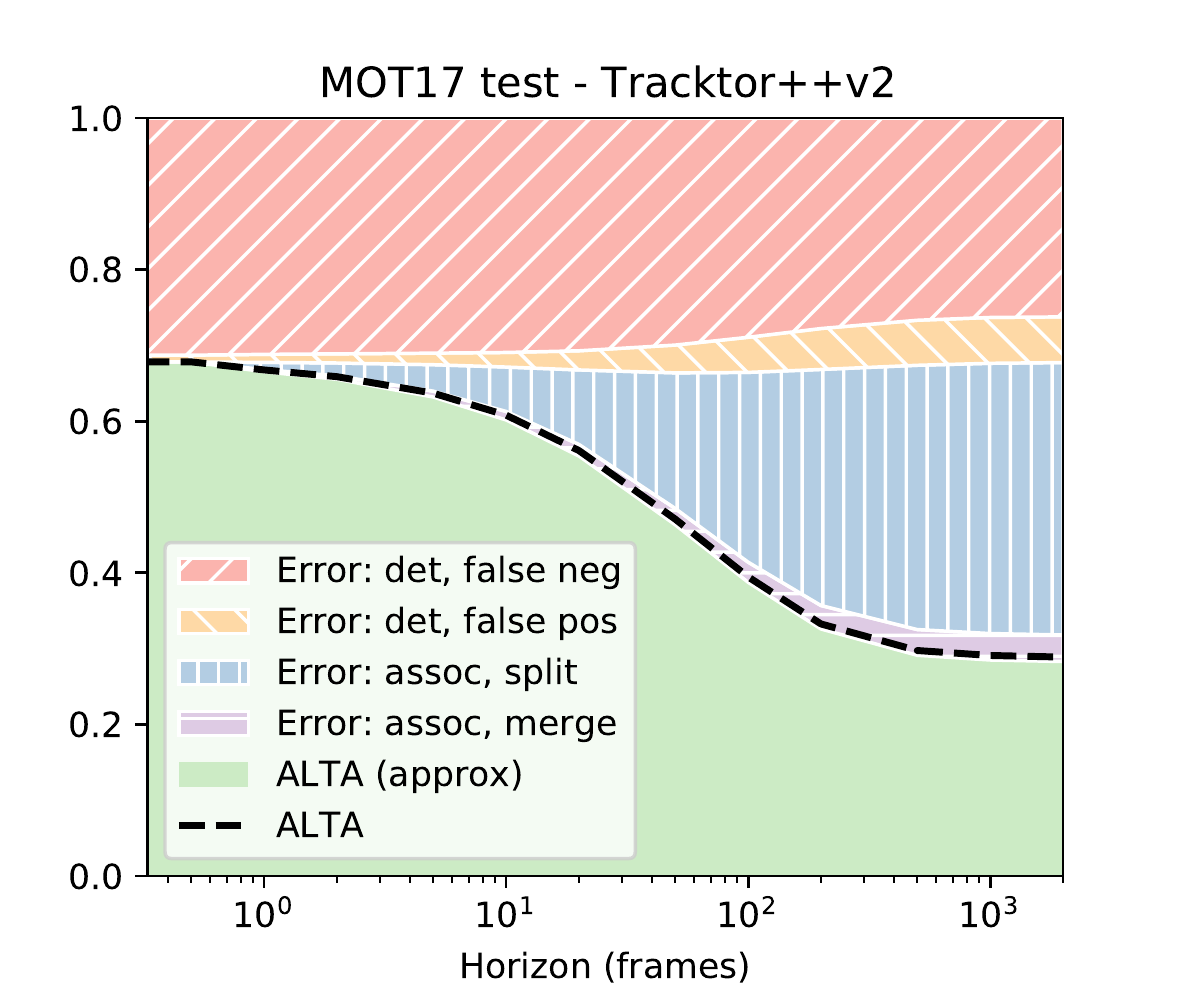}
\includegraphics[width=40mm,trim={5mm 0mm 5mm 0mm},clip]{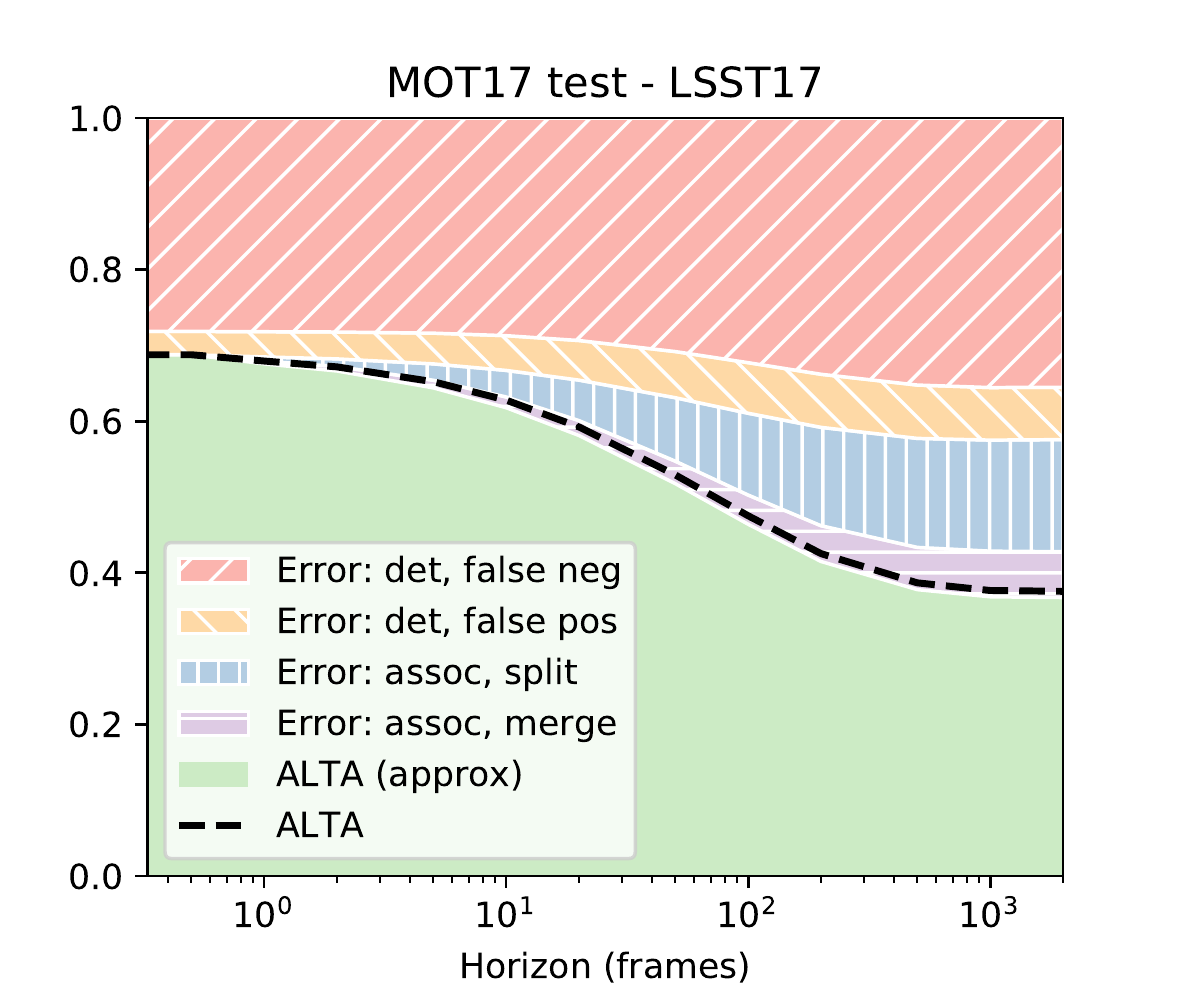}
\caption{Relative impact of each error type at different temporal horizons.
The approximate ALTA is close to the true value.}
\label{fig:decompose-vs-horizon}
\end{figure}

Figure~\ref{fig:decompose-vs-horizon} visualises the varying distribution of error types with respect to horizon.
As the horizon approaches zero, the association errors disappear.
As the horizon increases, so does the fraction of association errors, which is dominated by split errors for both trackers.

\subsection{Waymo Open Dataset}

\begin{figure}
\centering
\includegraphics[height=54mm]{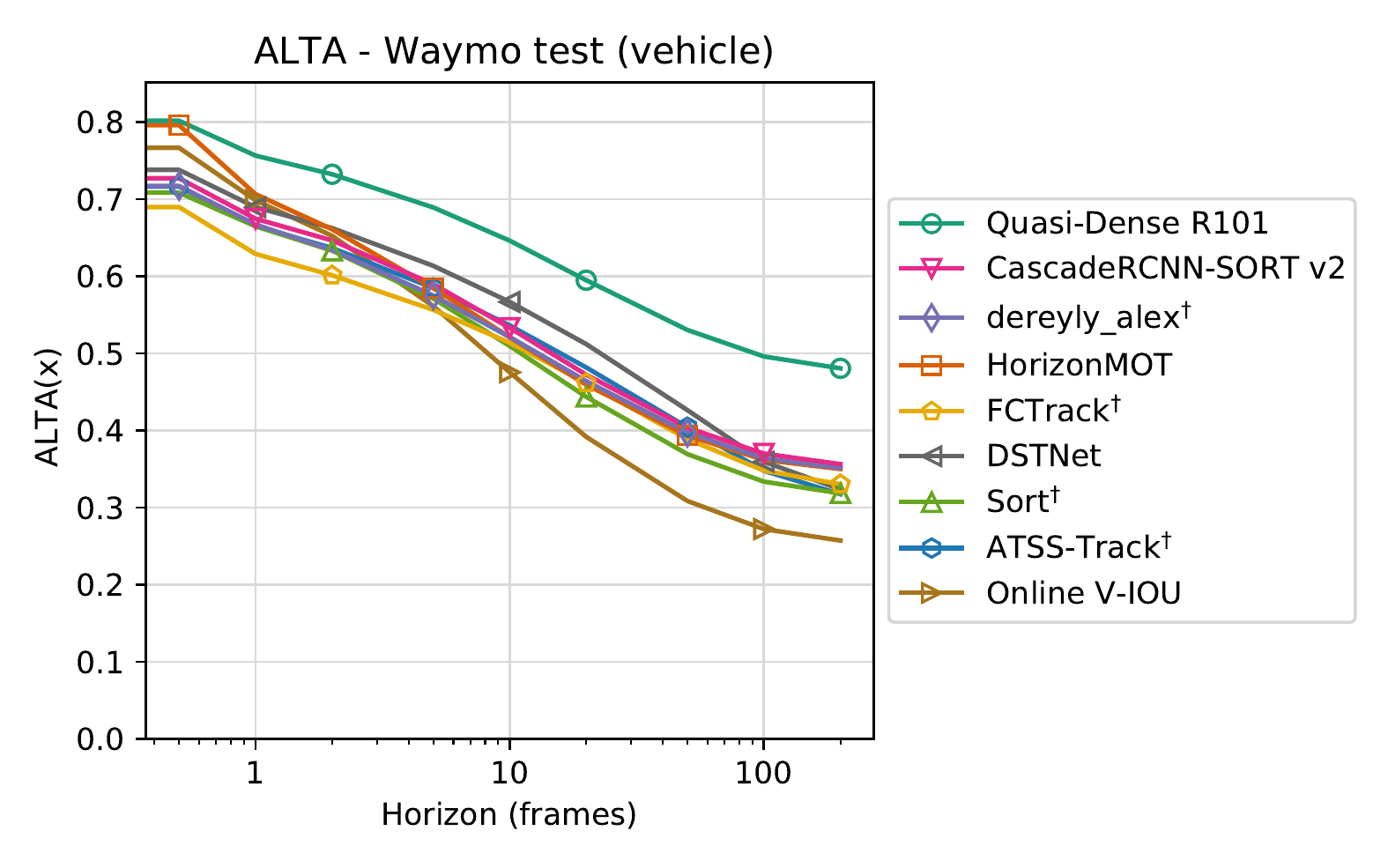}
\caption{ALTA versus horizon for submissions to the Waymo Open Dataset 2D Tracking competition. $\dagger$ indicates trackers with either missing publication or not submitted by an original author.}
\label{fig:alta_waymo}
\end{figure}

We also evaluate submissions to the Waymo Open Dataset~\cite{sun2020scalability} for vehicle tracking, comprising 20 sec sequences at 10Hz.
Autonomous driving further motivates the temporal horizon, as its requirements differ greatly from applications such as surveillance or object counting.
Specifically, knowing the path of a car in the previous seconds is more informative for trajectory prediction than the ability to recognise a car from several minutes ago.
Different types of association error also incur different costs: split errors may mean the re-initialisation of a motion model while merge errors may result in unreliable velocity estimates.

Figure~\ref{fig:alta_waymo} plots ALTA at different horizons for a subset of trackers listed on the public benchmark.
Two interesting trackers to compare are \textbf{Quasi-Dense R101} and \textbf{Online V-IOU}.
While both have high detection accuracy, the latter shows a steeper drop in association over just one second (ten frames).
Extended results are found in the appendix.

\section{Conclusion}

This paper has highlighted the dilemma posed by association in non-strict metrics for MOT.
The proposed local metrics provide an intuitive and meaningful way for benchmarks to specify the desired balance between detection and association, as well as providing insight into the temporal distribution of association errors.
Compared to current popular metrics, ATA and ALTA are able to place greater importance on association, diminishing the advantage of using an external detector.
Coupling the horizon parameter with error-type transparency makes ALTA a powerful tool for both quantifying and understanding tracker accuracy.
We have demonstrated its expressivity using the MOT 2017 and Waymo Open Dataset benchmarks %
and believe ALTA is a valuable metric for MOT researchers and practitioners alike.
Furthermore, the adoption of a finite maximum horizon may simplify video annotation and help protect individual privacy by distinguishing tracking from re-identification. %
Code will be made available online.

\paragraph{Acknowledgements}
We extend an enormous thanks to Patrick Dendorfer (MOT Challenge) and Pei Sun (Waymo Open Dataset) for running our code on the private test set. %
We also thank Jonathan Luiten, Vivek Rathod and Zhichao Lu for valuable discussions and feedback.

\clearpage
\appendix

\section{Extended results: MOT17}

\subsection{Comparison of more trackers}

\begin{figure*}
\centering
\includegraphics[width=0.48\textwidth]{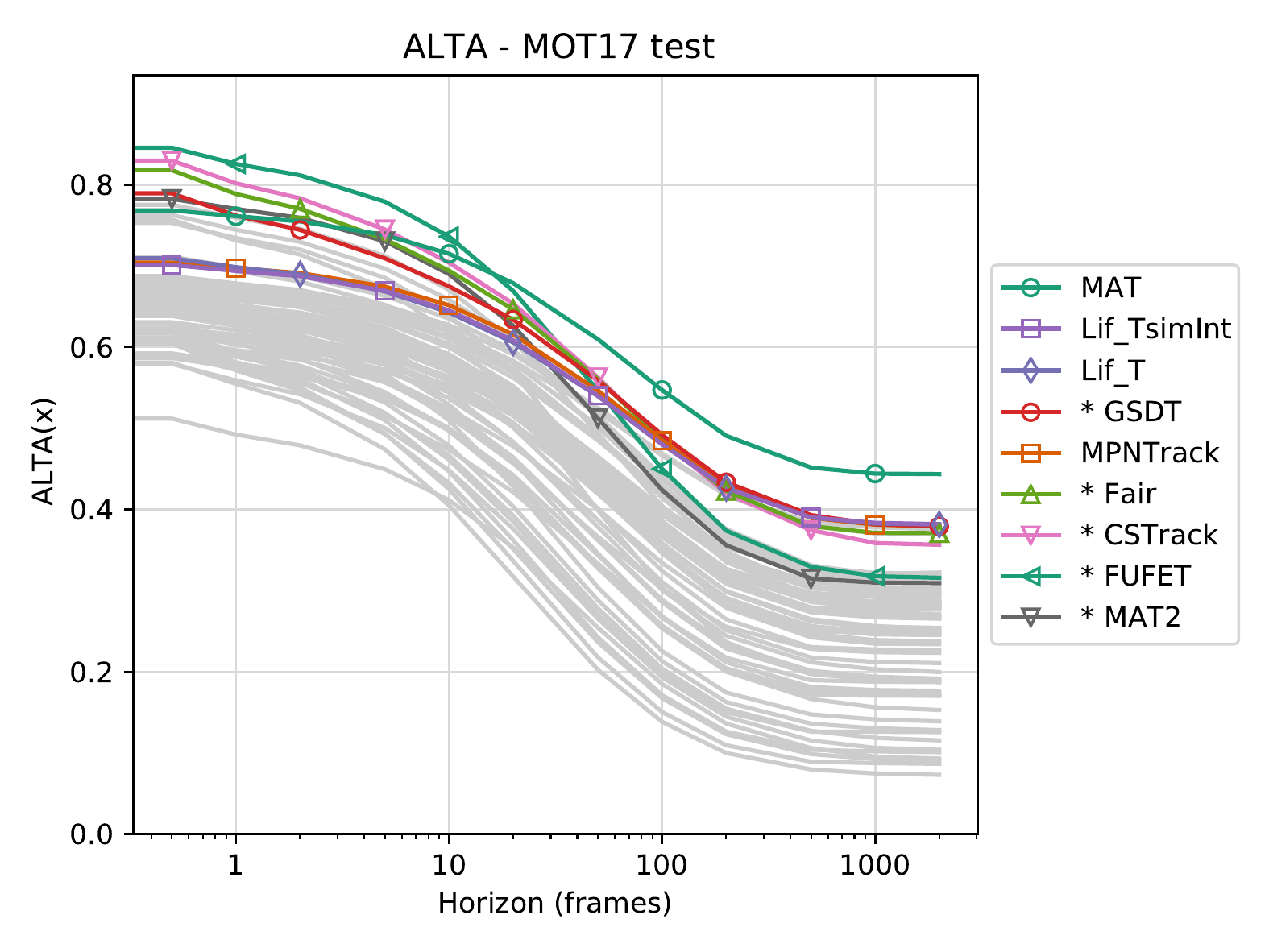}
\hspace{1em}
\includegraphics[width=0.48\textwidth]{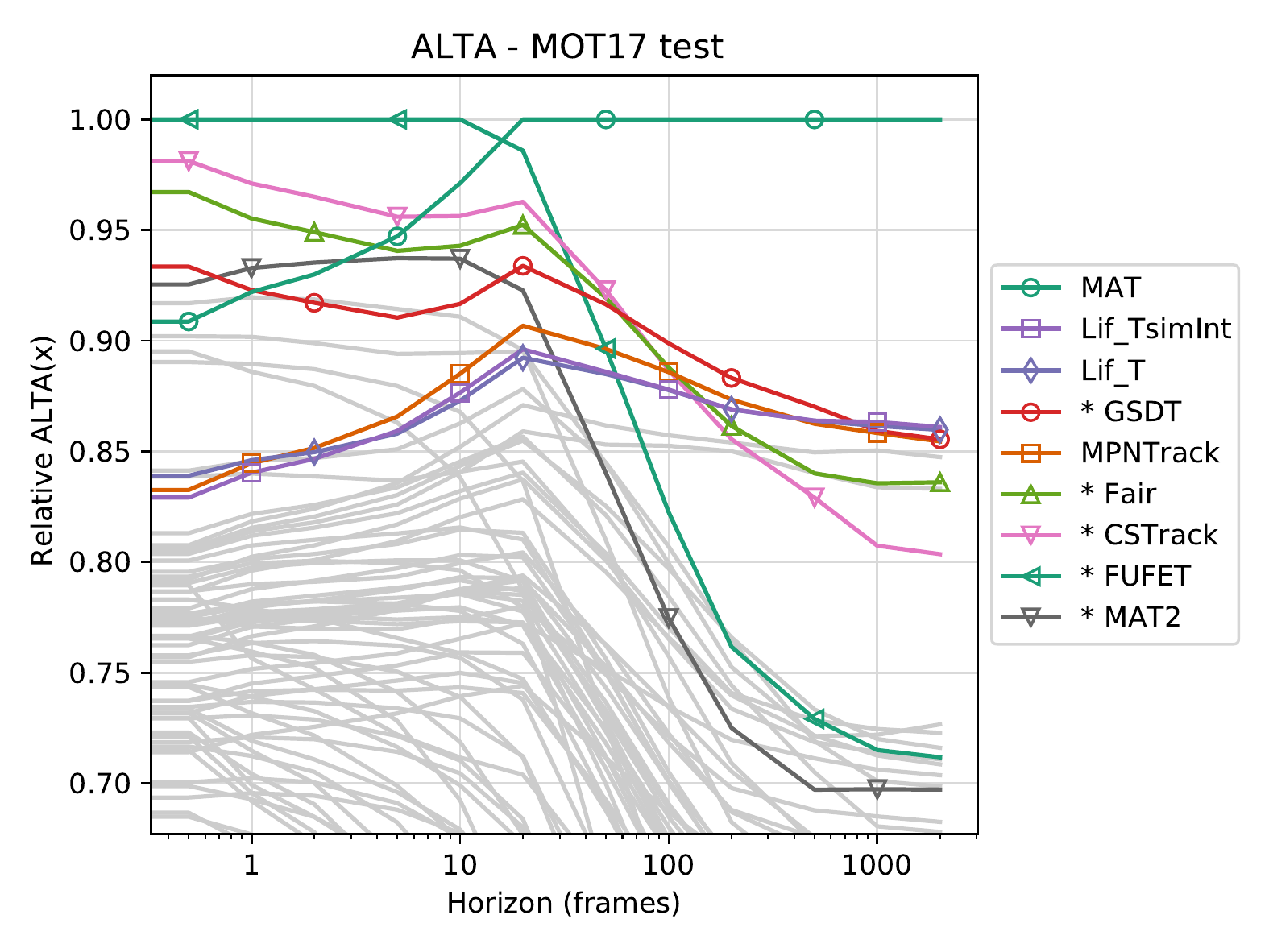}
\caption{
  The published trackers which achieve the highest ALTA at any horizon (top 5 highlighted).
  Other trackers are shown in grey.
  The right figure shows the value relative to the best tracker at each horizon for easier comparison.
}
\label{fig:mot17-extended-pub}
\end{figure*}

\begin{figure*}
\centering
\includegraphics[width=0.48\textwidth]{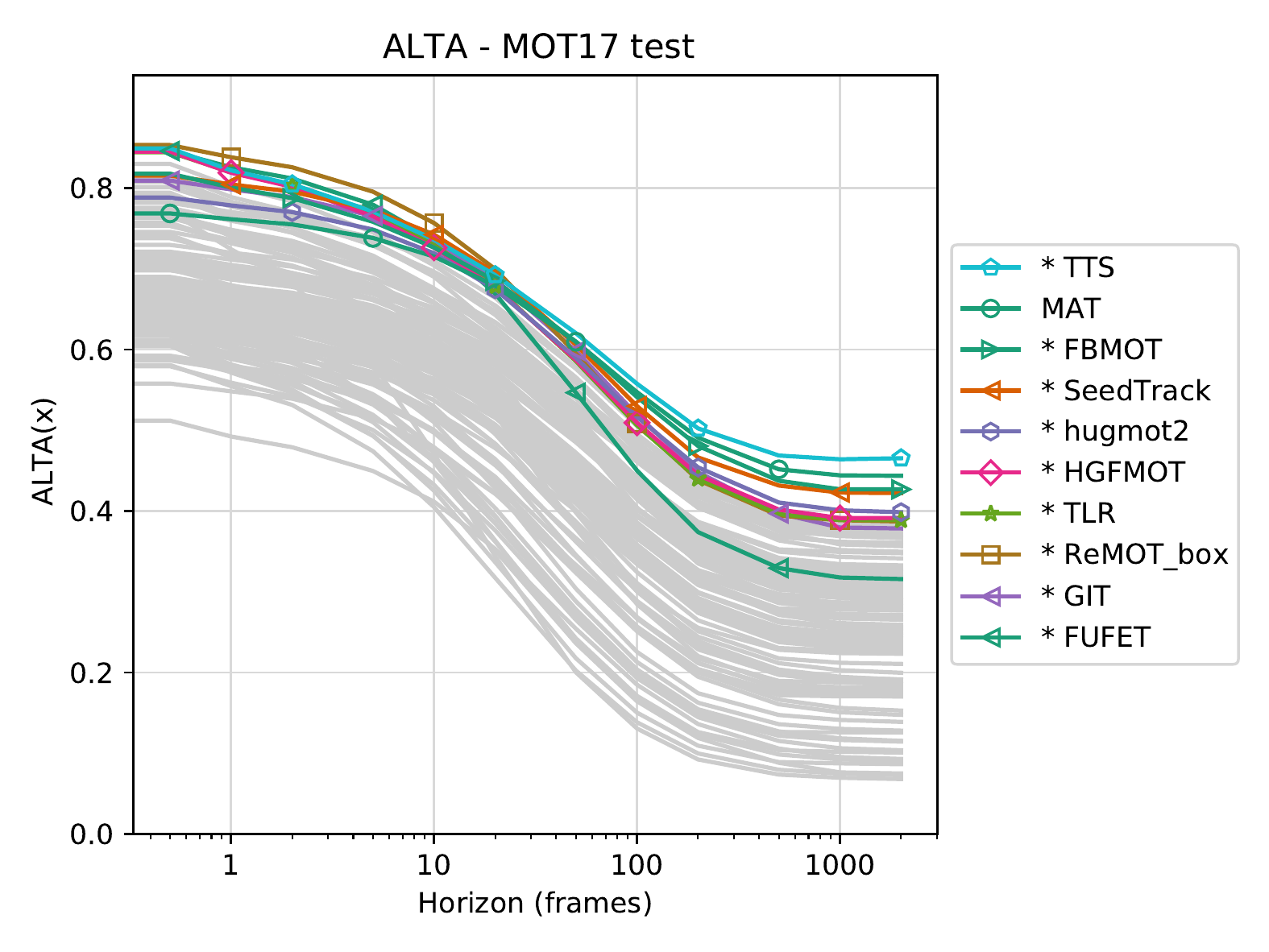}
\hspace{1em}
\includegraphics[width=0.48\textwidth]{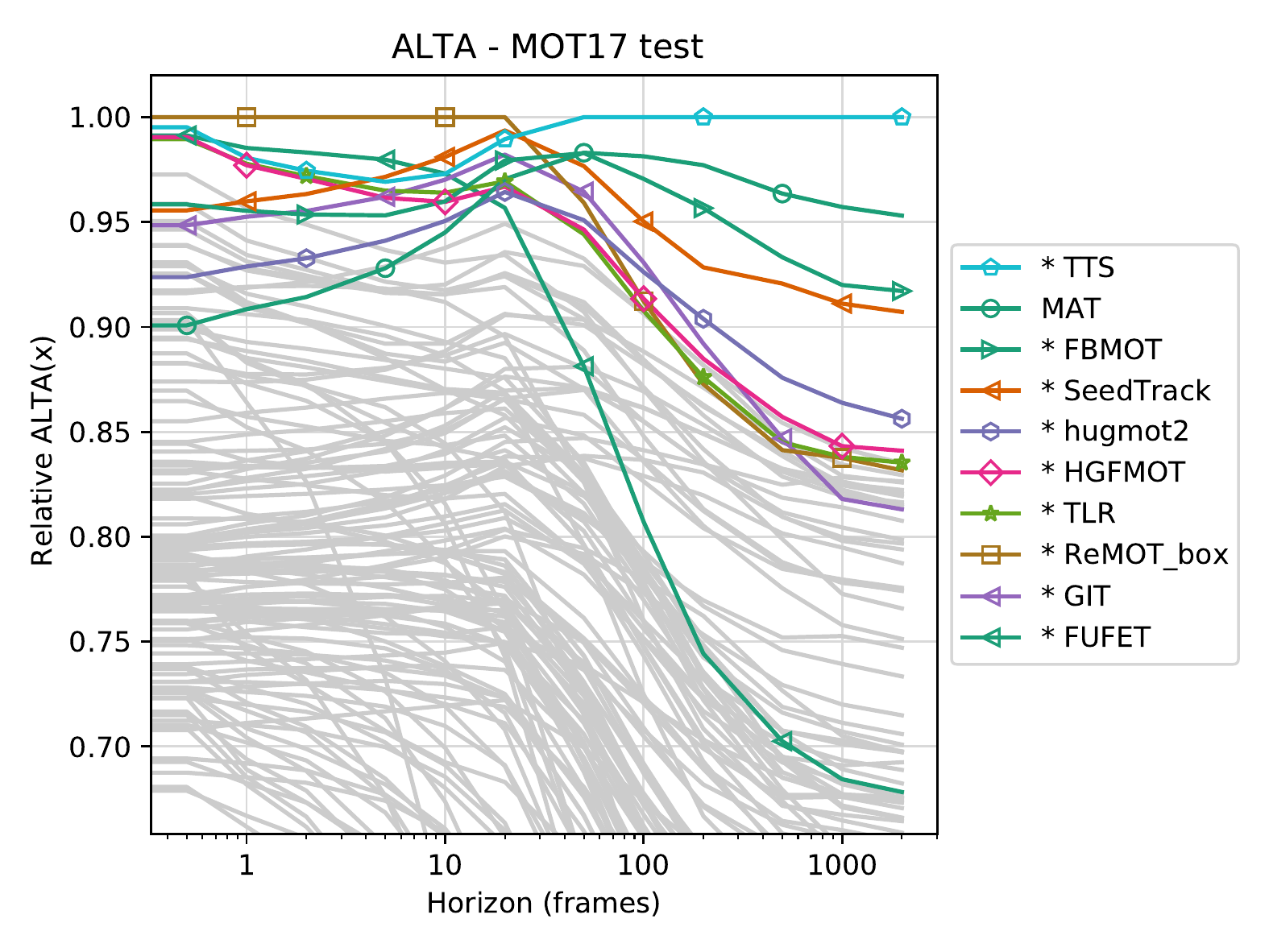}
\caption{
  The trackers (published or unpublished) which achieve the highest ALTA at any horizon (top 5 highlighted).
  Other trackers are shown in grey.
  The right figure shows the value relative to the best tracker at each horizon for easier comparison.
}
\label{fig:mot17-extended-unpub}
\end{figure*}

Table~\ref{tab:mot17-extended-pub} and Figure~\ref{fig:mot17-extended-pub} gives an extended comparison of the various metrics for the top published trackers.
Table~\ref{tab:mot17-extended-unpub} and Figure~\ref{fig:mot17-extended-unpub} are the same but include unpublished trackers, \ie~submissions for which no publication or technical report is available.
A few key observations are as follows.
\begin{itemize}[\textendash]
\item
  DetF1 has a stronger influence on MOTA, HOTA and IDF1 than on ATA: the trackers with the best DetF1 are typically ranked higher under those metrics.
\item
  Identity switches are a temperamental predictor of the association fraction $\s{ATA}/\s{DetF1}$.
  The methods with the least switches often have a high association fraction but the converse is not necessarily true.
  For example, `GSDT' (4th in published) and `TTS' (1st in unpublished) have a large number of switches.
  This may be explained by switch-and-switch-back events.
\item
  Even under ATA, it is possible for improved detection to ``compensate'' for weaker association.
  Amongst the unpublished trackers with the highest ATA, 8 of the top 10 algorithms use private detectors.
  In particular, the trackers `HGFMOT', `TLR' and \mbox{`ReMOT\_box'} are in the top 10 despite a poor ratio of $\s{ATA}/\s{DetF1}$.
\end{itemize}

\begin{figure*}
\centering
\includegraphics[height=150mm]{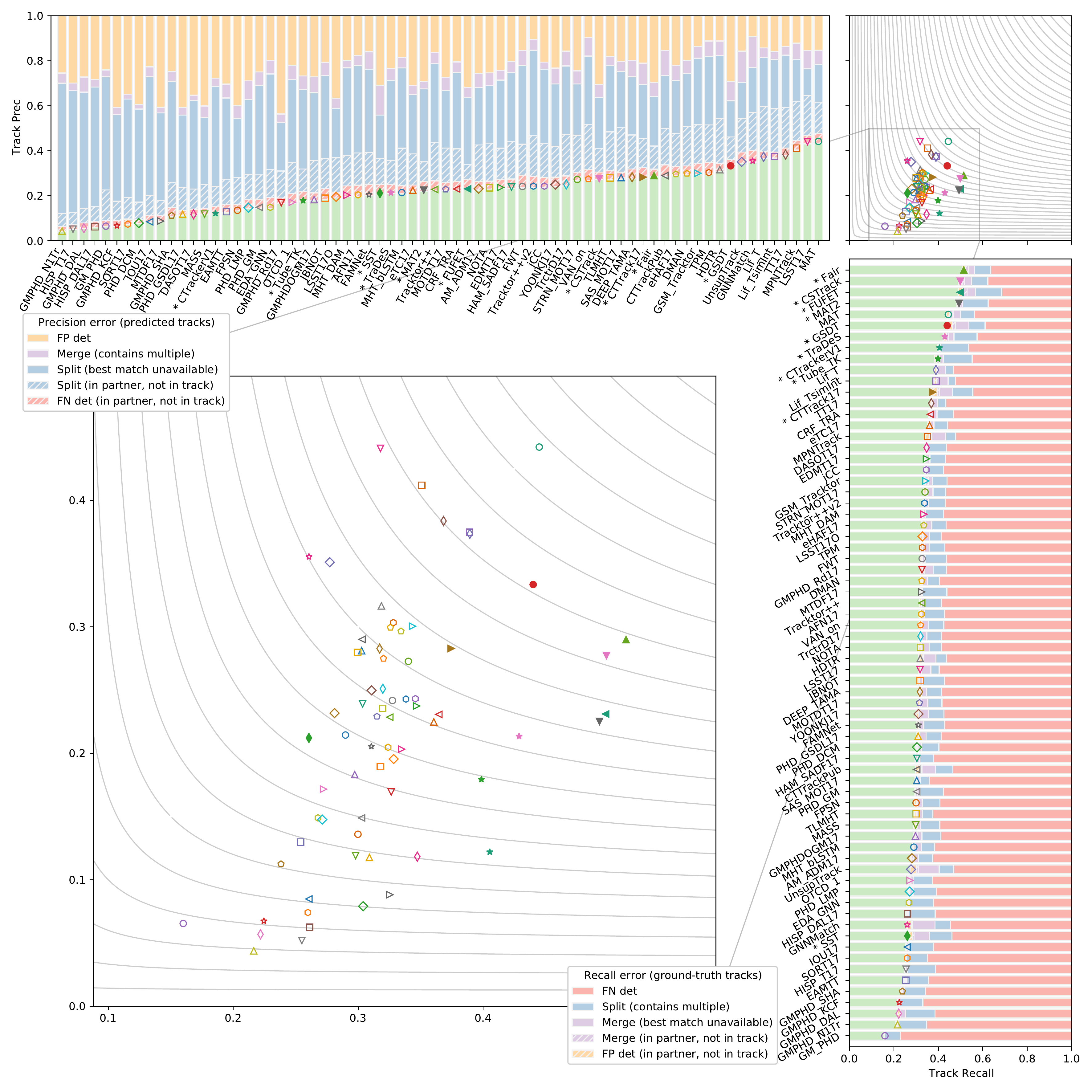}
\caption{
    Decomposition of track precision (ATP) and recall (ATR) for the 70 published trackers on the MOT 2017 benchmark.
    Trackers that use a private detector are distinguished by a filled marker.
}
\label{fig:mot17-decompose-bars-extended-pub}
\end{figure*}

\begin{figure*}
\centering
\includegraphics[height=50mm]{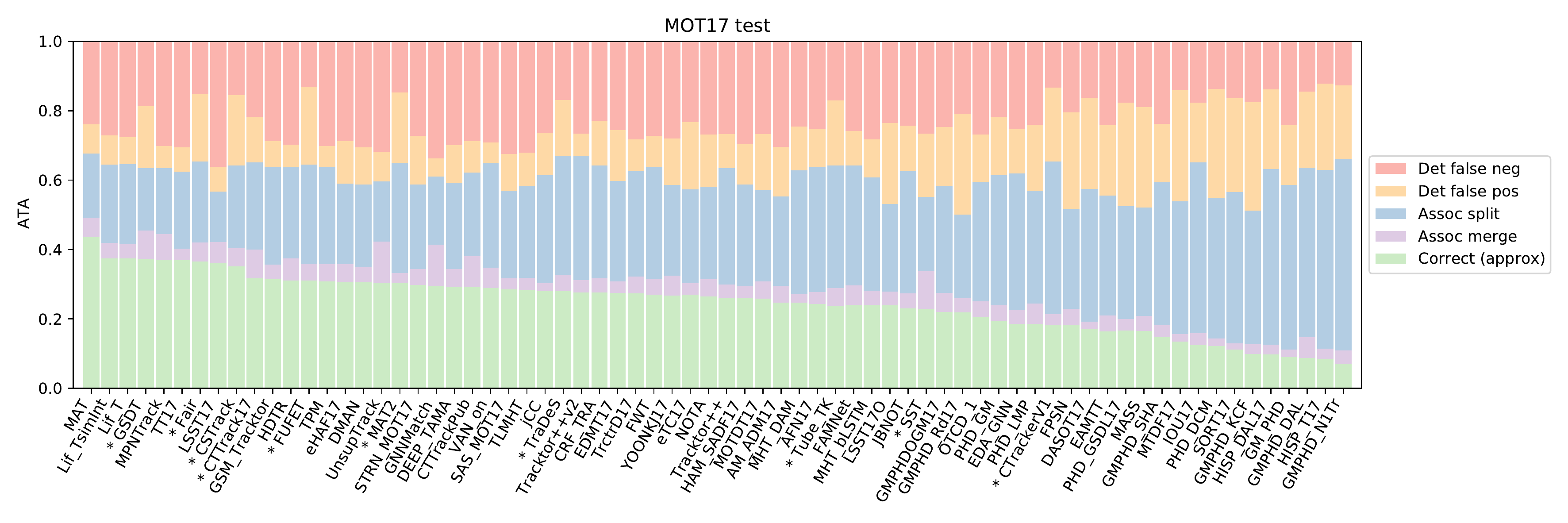}
\caption{Distribution of error types in ATA for the 70 published trackers in the MOT 2017 benchmark.}
\label{fig:mot17-decompose-extended-pub}
\end{figure*}

Figure~\ref{fig:mot17-decompose-bars-extended-pub} shows the decomposition of ATR and ATP error into different types for all published trackers and Figure~\ref{fig:mot17-decompose-extended-pub} shows the overall error decomposition of ATA.
A few key observations are as follows.
\begin{itemize}[\textendash]
\item The top tracker `MAT' stands apart from the next-best published methods.
\item As in the main paper, splits have a greater effect than merges, particularly for track precision.
\item Trackers using private detectors generally achieve higher track recall and are clustered towards the right of the scatter plot.
\item Amongst the trackers which achieve the highest ATA score, `LSST17' has a particularly large fraction of false negative errors, while this quantity is small for `GSDT', `Fair' and `CTTrack17'.
\item For track precision, the worst methods have a larger fraction of split errors due to unmatched predicted tracks (``best match unavailable'') than due to not finding the full extent of the ground-truth track (``in partner, not in track'').
This suggests that the ground-truth track has been split into many predicted tracks.
\end{itemize}

Several trackers have two variants submitted to the challenge and it can be informative to contrast these pairs.
`MAT2' (using private detections) achieves better ATR than its public counterpart `MAT', but seems to suffer from a much higher rate of split errors.
Similarly, `CTTrackPub' is a version of `CTTrack17' (CenterTrack) that is constrained to use the public set of detections for track initialisation.
As observed in the main paper, `CTTrackPub' achieves similar precision but significantly worse recall.
`UnsupTrack', which adds an embedding to CenterTrack, achieves a relative increase in ATP but not ATR, as expected.
`Tracktor++v2' achieves a slight boost in both precision and recall over the earlier implementation `Tracktor++'.
Finally, `Lif\_TsimInt' is a variant of `Lif\_T' that uses simple linear interpolation instead of Tracktor-style box regression.
This modification is shown to have a marginal impact on the metrics.

\clearpage

\begin{table*}
\centering
\caption{
  Metric comparison for MOT 2017 benchmark (9 Mar 2021) for top~50 (of~70) published trackers by ATA.
  The top~5 for each metric are shown in bold.
}
\label{tab:mot17-extended-pub}
\scalebox{0.85}{
\begin{tabular}{l r@{~~}l r@{~~}l r@{~~}l r@{~~}l |  r@{~~}l r@{~~}l r@{~~}l r@{~~}l r@{~~}l}
\toprule
Tracker & \multicolumn{2}{c}{MOTA} & \multicolumn{2}{c}{IDSw} & \multicolumn{2}{c}{HOTA$_{0.5}$} & \multicolumn{2}{c}{IDF1} & \multicolumn{2}{c}{DetF1} & \multicolumn{2}{c}{ALTA(1s)} & \multicolumn{2}{c}{ALTA(5s)} & \multicolumn{2}{c}{\textbf{ATA}} & \multicolumn{2}{c}{ATA$/$DetF1} \\
\midrule
                      MAT \cite{han2020mat} &  .671 &       (7) &          1279 &               (6) &       .677 &      {\bf (4)} &  .692 & {\bf (3)} & .769 &       (7) &     .660 &    {\bf (1)} &     .520 &    {\bf (1)} & .443 & {\bf (1)} &       .577 &      {\bf (1)} \\
    Lif\_TsimInt \cite{hornakova2020lifted} &  .582 &      (15) &          1022 &         {\bf (1)} &       .626 &           (10) &  .652 &       (7) & .701 &      (15) &     .587 &          (8) &     .451 &          (6) & .382 & {\bf (2)} &       .545 &      {\bf (3)} \\
          Lif\_T \cite{hornakova2020lifted} &  .605 &      (13) &          1189 &         {\bf (4)} &       .638 &            (9) &  .656 &       (6) & .710 &      (12) &     .585 &          (9) &     .451 &    {\bf (5)} & .381 & {\bf (3)} &       .537 &            (6) \\
                * GSDT \cite{wang2020joint} &  .662 &       (9) &          3318 &              (44) &       .662 &      {\bf (5)} &  .687 & {\bf (4)} & .789 & {\bf (4)} &     .612 &    {\bf (5)} &     .462 &    {\bf (2)} & .379 & {\bf (4)} &       .481 &            (9) \\
          MPNTrack \cite{braso2020learning} &  .588 &      (14) &          1185 &         {\bf (3)} &       .603 &           (12) &  .617 &      (13) & .704 &      (14) &     .596 &          (6) &     .456 &    {\bf (3)} & .379 & {\bf (5)} &       .538 &      {\bf (5)} \\
                  TT17 \cite{zhang2020long} &  .549 &      (20) &          1088 &         {\bf (2)} &       .601 &           (13) &  .631 &      (10) & .683 &      (17) &     .571 &         (13) &     .443 &          (9) & .376 &       (6) &       .550 &      {\bf (2)} \\
             * Fair \cite{zhang2020fairmot} &  .737 & {\bf (3)} &          3303 &              (43) &       .719 &      {\bf (2)} &  .723 & {\bf (2)} & .818 & {\bf (3)} &     .623 &    {\bf (4)} &     .455 &    {\bf (4)} & .371 &       (7) &       .453 &           (16) \\
                LSST17 \cite{feng2019multi} &  .547 &      (21) &          1243 &         {\bf (5)} &       .589 &           (15) &  .623 &      (12) & .681 &      (19) &     .565 &         (14) &     .447 &          (8) & .369 &       (8) &       .543 &      {\bf (4)} \\
       * CSTrack \cite{liang2020rethinking} &  .749 & {\bf (2)} &          3567 &              (47) &       .721 &      {\bf (1)} &  .726 & {\bf (1)} & .830 & {\bf (2)} &     .627 &    {\bf (3)} &     .450 &          (7) & .356 &       (9) &       .429 &           (24) \\
        * CTTrack17 \cite{zhou2020tracking} &  .678 &       (6) &          3039 &              (38) &       .651 &            (7) &  .647 &       (8) & .763 &       (8) &     .583 &         (10) &     .406 &         (13) & .322 &      (10) &       .422 &           (27) \\
            GSM\_Tracktor \cite{liu2020gsm} &  .564 &      (17) &          1485 &               (8) &       .565 &           (19) &  .578 &      (20) & .680 &      (20) &     .548 &         (18) &     .391 &         (15) & .321 &      (11) &       .472 &           (11) \\
             HDTR \cite{babaee2018multiple} &  .541 &      (23) &          1895 &              (15) &       .566 &           (18) &  .595 &      (15) & .682 &      (18) &     .559 &         (16) &     .410 &         (11) & .318 &      (12) &       .466 &           (13) \\
           * FUFET \cite{shan2020tracklets} &  .762 & {\bf (1)} &          3237 &              (42) &       .700 &      {\bf (3)} &  .680 & {\bf (5)} & .846 & {\bf (1)} &     .630 &    {\bf (2)} &     .406 &         (12) & .316 &      (13) &       .373 &           (42) \\
                     TPM \cite{peng2020tpm} &  .542 &      (22) &          1824 &              (11) &       .518 &           (35) &  .526 &      (38) & .670 &      (24) &     .544 &         (19) &     .385 &         (17) & .315 &      (14) &       .471 &           (12) \\
       eHAF17 \cite{sheng2018heterogeneous} &  .518 &      (32) &          1834 &              (12) &       .535 &           (28) &  .547 &      (28) & .665 &      (28) &     .538 &         (20) &     .387 &         (16) & .314 &      (15) &       .472 &           (10) \\
                  DMAN \cite{zhu2018online} &  .482 &      (50) &          2194 &              (24) &       .532 &           (30) &  .557 &      (26) & .631 &      (48) &     .506 &         (33) &     .378 &         (20) & .312 &      (16) &       .495 &            (7) \\
        UnsupTrack \cite{karthik2020simple} &  .617 &      (11) &          1864 &              (13) &       .582 &           (17) &  .581 &      (18) & .711 &      (11) &     .576 &         (11) &     .415 &         (10) & .310 &      (17) &       .435 &           (21) \\
                   * MAT2 \cite{han2020mat} &  .695 & {\bf (4)} &          2844 &              (37) &       .649 &            (8) &  .631 &      (11) & .783 & {\bf (5)} &     .591 &          (7) &     .384 &         (19) & .309 &      (18) &       .395 &           (38) \\
           STRN\_MOT17 \cite{xu2019spatial} &  .509 &      (38) &          2397 &              (31) &       .537 &           (26) &  .560 &      (25) & .656 &      (32) &     .507 &         (31) &     .364 &         (23) & .303 &      (19) &       .462 &           (15) \\
       GNNMatch \cite{papakis2020gcnnmatch} &  .573 &      (16) &          1911 &              (16) &       .557 &           (20) &  .563 &      (24) & .688 &      (16) &     .562 &         (15) &     .406 &         (14) & .301 &      (20) &       .437 &           (20) \\
           DEEP\_TAMA \cite{yoon2020online} &  .503 &      (42) &          2192 &              (23) &       .520 &           (34) &  .535 &      (36) & .645 &      (44) &     .513 &         (27) &     .361 &         (25) & .299 &      (21) &       .464 &           (14) \\
         CTTrackPub \cite{zhou2020tracking} &  .615 &      (12) &          2583 &              (34) &       .595 &           (14) &  .596 &      (14) & .709 &      (13) &     .552 &         (17) &     .385 &         (18) & .296 &      (22) &       .418 &           (30) \\
                  VAN\_on \cite{lee2020van} &  .552 &      (19) &          2220 &              (26) &       .537 &           (25) &  .542 &      (31) & .665 &      (27) &     .522 &         (23) &     .363 &         (24) & .296 &      (23) &       .445 &           (18) \\
    SAS\_MOT17 \cite{maksai2019eliminating} &  .442 &      (64) &          1529 &               (9) &       .535 &           (29) &  .572 &      (22) & .604 &      (64) &     .485 &         (44) &     .352 &         (26) & .291 &      (24) &       .483 &            (8) \\
            TLMHT \cite{sheng2018iterative} &  .506 &      (41) &          1407 &               (7) &       .547 &           (24) &  .565 &      (23) & .640 &      (46) &     .518 &         (25) &     .374 &         (21) & .289 &      (25) &       .452 &           (17) \\
                jCC \cite{keuper2018motion} &  .512 &      (37) &          1802 &              (10) &       .535 &           (27) &  .545 &      (29) & .647 &      (42) &     .506 &         (32) &     .344 &         (32) & .286 &      (26) &       .441 &           (19) \\
                * TraDeS \cite{wu2021track} &  .691 & {\bf (5)} &          3555 &              (46) &       .652 &            (6) &  .639 &       (9) & .775 &       (6) &     .574 &         (12) &     .367 &         (22) & .285 &      (27) &       .368 &           (44) \\
   Tracktor++v2 \cite{bergmann2019tracking} &  .563 &      (18) &          1987 &              (18) &       .550 &           (23) &  .551 &      (27) & .673 &      (22) &     .525 &         (22) &     .351 &         (27) & .283 &      (28) &       .420 &           (29) \\
               CRF\_TRA \cite{xiang2020end} &  .531 &      (26) &          2518 &              (33) &       .527 &           (31) &  .537 &      (35) & .677 &      (21) &     .519 &         (24) &     .346 &         (29) & .283 &      (29) &       .417 &           (31) \\
            EDMT17 \cite{chen2017enhancing} &  .500 &      (43) &          2264 &              (27) &       .509 &           (40) &  .513 &      (43) & .653 &      (38) &     .509 &         (28) &     .346 &         (30) & .282 &      (30) &       .432 &           (22) \\
                TrctrD17 \cite{xu2020train} &  .537 &      (24) &          1947 &              (17) &       .527 &           (32) &  .538 &      (34) & .654 &      (35) &     .509 &         (29) &     .344 &         (31) & .281 &      (31) &       .430 &           (23) \\
              FWT \cite{henschel2018fusion} &  .513 &      (35) &          2648 &              (36) &       .479 &           (48) &  .476 &      (52) & .655 &      (34) &     .505 &         (34) &     .340 &         (34) & .278 &      (32) &       .424 &           (26) \\
          YOONKJ17 \cite{yoon2020oneshotda} &  .514 &      (34) &          2118 &              (21) &       .454 &           (55) &  .540 &      (33) & .659 &      (30) &     .517 &         (26) &     .349 &         (28) & .277 &      (33) &       .420 &           (28) \\
               eTC17 \cite{wang2019exploit} &  .519 &      (31) &          2288 &              (29) &       .555 &           (21) &  .581 &      (17) & .669 &      (25) &     .507 &         (30) &     .335 &         (35) & .277 &      (34) &       .414 &           (33) \\
              NOTA \cite{chen2019aggregate} &  .513 &      (36) &          2285 &              (28) &       .526 &           (33) &  .545 &      (30) & .654 &      (36) &     .499 &         (36) &     .341 &         (33) & .271 &      (35) &       .415 &           (32) \\
     Tracktor++ \cite{bergmann2019tracking} &  .535 &      (25) &          2072 &              (20) &       .518 &           (36) &  .523 &      (39) & .652 &      (39) &     .503 &         (35) &     .333 &         (38) & .268 &      (36) &       .412 &           (34) \\
          HAM\_SADF17 \cite{yoon2018online} &  .483 &      (49) &          1871 &              (14) &       .500 &           (44) &  .511 &      (44) & .624 &      (51) &     .490 &         (42) &     .333 &         (37) & .267 &      (37) &       .429 &           (25) \\
                MOTDT17 \cite{chen2018real} &  .509 &      (39) &          2474 &              (32) &       .511 &           (39) &  .527 &      (37) & .653 &      (37) &     .497 &         (38) &     .333 &         (36) & .265 &      (38) &       .406 &           (36) \\
           AM\_ADM17 \cite{lee2018learning} &  .481 &      (51) &          2214 &              (25) &       .505 &           (43) &  .521 &      (40) & .623 &      (52) &     .479 &         (47) &     .326 &         (39) & .254 &      (39) &       .407 &           (35) \\
            MHT\_DAM \cite{kim2015multiple} &  .507 &      (40) &          2314 &              (30) &       .480 &           (47) &  .472 &      (53) & .648 &      (41) &     .492 &         (40) &     .318 &         (41) & .253 &      (40) &       .390 &           (39) \\
              AFN17 \cite{shen2018tracklet} &  .515 &      (33) &          2593 &              (35) &       .478 &           (49) &  .469 &      (55) & .655 &      (33) &     .497 &         (39) &     .321 &         (40) & .251 &      (41) &       .383 &           (40) \\
           * Tube\_TK \cite{pang2020tubetk} &  .630 &      (10) &          4137 &              (52) &       .586 &           (16) &  .586 &      (16) & .753 &      (10) &     .529 &         (21) &     .312 &         (46) & .247 &      (42) &       .328 &           (50) \\
                FAMNet \cite{chu2019famnet} &  .520 &      (30) &          3072 &              (40) &       .000 &           (68) &  .487 &      (48) & .657 &      (31) &     .487 &         (43) &     .312 &         (44) & .247 &      (43) &       .376 &           (41) \\
             MHT\_bLSTM \cite{kim2018multi} &  .475 &      (53) &          2069 &              (19) &       .506 &           (42) &  .519 &      (41) & .619 &      (56) &     .479 &         (46) &     .317 &         (42) & .246 &      (44) &       .398 &           (37) \\
               LSST17O \cite{feng2019multi} &  .527 &      (27) &          2167 &              (22) &       .554 &           (22) &  .579 &      (19) & .662 &      (29) &     .479 &         (45) &     .315 &         (43) & .245 &      (45) &       .370 &           (43) \\
          JBNOT \cite{henschel2019multiple} &  .526 &      (28) &          3050 &              (39) &       .508 &           (41) &  .508 &      (45) & .671 &      (23) &     .497 &         (37) &     .312 &         (45) & .237 &      (46) &       .354 &           (46) \\
                   * SST \cite{sun2019deep} &  .524 &      (29) &          8431 &              (66) &       .489 &           (45) &  .495 &      (47) & .668 &      (26) &     .454 &         (49) &     .306 &         (47) & .234 &      (47) &       .350 &           (48) \\
           GMPHDOGM17 \cite{song2019online} &  .499 &      (44) &          3125 &              (41) &       .000 &           (68) &  .471 &      (54) & .641 &      (45) &     .471 &         (48) &     .292 &         (48) & .227 &      (48) &       .354 &           (47) \\
      GMPHD\_Rd17 \cite{baisa2019occlusion} &  .468 &      (55) &          3865 &              (49) &       .518 &           (37) &  .541 &      (32) & .630 &      (49) &     .424 &         (54) &     .274 &         (50) & .223 &      (49) &       .354 &           (45) \\
                 OTCD\_1 \cite{liu2019real} &  .486 &      (48) &          3502 &              (45) &       .477 &           (50) &  .479 &      (50) & .621 &      (53) &     .454 &         (50) &     .278 &         (49) & .211 &      (50) &       .339 &           (49) \\

\bottomrule
\end{tabular}
}
\end{table*}

\begin{table*}
\centering
\caption{
  Metric comparison for MOT 2017 benchmark including unpublished trackers (9 Mar 2021) for the top~50 (of~125) trackers by~ATA.
  The top~5 for each metric are shown in bold.
}
\scalebox{0.85}{
\begin{tabular}{l r@{~~}l r@{~~}l r@{~~}l r@{~~}l |  r@{~~}l r@{~~}l r@{~~}l r@{~~}l r@{~~}l}
\toprule
Tracker & \multicolumn{2}{c}{MOTA} & \multicolumn{2}{c}{IDSw} & \multicolumn{2}{c}{HOTA$_{0.5}$} & \multicolumn{2}{c}{IDF1} & \multicolumn{2}{c}{DetF1} & \multicolumn{2}{c}{ALTA(1s)} & \multicolumn{2}{c}{ALTA(5s)} & \multicolumn{2}{c}{\textbf{ATA}} & \multicolumn{2}{c}{ATA$/$DetF1} \\
\midrule
                                     * TTS  &  .767 & {\bf (3)} &          2346 &              (53) &       .739 &      {\bf (2)} &  .752 & {\bf (1)} & .849 & {\bf (2)} &     .671 &    {\bf (1)} &     .532 &    {\bf (1)} & .465 & {\bf (1)} &       .548 &      {\bf (3)} \\
                      MAT \cite{han2020mat} &  .671 &      (25) &          1279 &               (9) &       .677 &           (16) &  .692 &      (14) & .769 &      (26) &     .660 &          (6) &     .520 &    {\bf (2)} & .443 & {\bf (2)} &       .577 &      {\bf (1)} \\
                                   * FBMOT  &  .740 &       (7) &          2166 &              (40) &       .735 &      {\bf (4)} &  .744 & {\bf (2)} & .818 &       (8) &     .666 &    {\bf (3)} &     .516 &    {\bf (3)} & .427 & {\bf (3)} &       .522 &           (10) \\
                               * SeedTrack  &  .687 &      (18) &          2571 &              (60) &       .688 &           (13) &  .707 &      (12) & .815 &       (9) &     .668 &    {\bf (2)} &     .497 &    {\bf (4)} & .422 & {\bf (4)} &       .518 &           (11) \\
                                 * hugmot2  &  .688 &      (17) &          2190 &              (42) &       .642 &           (25) &  .646 &      (30) & .788 &      (19) &     .650 &          (9) &     .483 &          (6) & .398 & {\bf (5)} &       .506 &           (14) \\
                                  * HGFMOT  &  .771 & {\bf (1)} &          3480 &              (86) &       .741 &      {\bf (1)} &  .741 & {\bf (3)} & .845 & {\bf (4)} &     .653 &          (7) &     .480 &          (7) & .391 &       (6) &       .463 &           (35) \\
                                     * TLR  &  .765 & {\bf (4)} &          3369 &              (82) &       .737 &      {\bf (3)} &  .736 & {\bf (4)} & .844 & {\bf (5)} &     .653 &          (8) &     .474 &          (9) & .389 &       (7) &       .460 &           (37) \\
                              * XJTU\_priv  &  .682 &      (19) &          2262 &              (48) &       .642 &           (26) &  .647 &      (29) & .782 &      (21) &     .639 &         (10) &     .472 &         (10) & .389 &       (8) &       .497 &           (19) \\
                              * ReMOT\_box  &  .770 & {\bf (2)} &          2853 &              (66) &       .729 &      {\bf (5)} &  .720 &      (10) & .853 & {\bf (1)} &     .664 &    {\bf (4)} &     .463 &         (13) & .387 &       (9) &       .454 &           (41) \\
                                   RGCN\_T  &  .639 &      (32) &          1774 &              (19) &       .648 &           (24) &  .661 &      (21) & .746 &      (33) &     .613 &         (18) &     .468 &         (11) & .386 &      (10) &       .518 &           (12) \\
                                  LPC\_MOT  &  .590 &      (43) &          1122 &         {\bf (3)} &       .641 &           (27) &  .668 &      (19) & .708 &      (44) &     .592 &         (24) &     .449 &         (21) & .385 &      (11) &       .543 &      {\bf (5)} \\
                            * SLA\_Tracker  &  .718 &      (13) &          2493 &              (56) &       .683 &           (14) &  .690 &      (15) & .801 &      (14) &     .633 &         (11) &     .477 &          (8) & .384 &      (12) &       .480 &           (25) \\
                                    * FMv2  &  .727 &      (10) &          3132 &              (77) &       .719 &            (8) &  .725 &       (7) & .807 &      (12) &     .625 &         (14) &     .465 &         (12) & .383 &      (13) &       .474 &           (27) \\
    Lif\_TsimInt \cite{hornakova2020lifted} &  .582 &      (48) &          1022 &         {\bf (1)} &       .626 &           (31) &  .652 &      (26) & .701 &      (47) &     .587 &         (26) &     .451 &         (19) & .382 &      (14) &       .545 &      {\bf (4)} \\
          Lif\_T \cite{hornakova2020lifted} &  .605 &      (38) &          1189 &               (6) &       .638 &           (29) &  .656 &      (24) & .710 &      (42) &     .585 &         (29) &     .451 &         (18) & .381 &      (15) &       .537 &            (9) \\
                                 * SD\_MOT  &  .732 &       (9) &          2964 &              (69) &       .719 &            (7) &  .728 & {\bf (5)} & .811 &      (10) &     .625 &         (15) &     .463 &         (14) & .380 &      (16) &       .468 &           (31) \\
                * GSDT \cite{wang2020joint} &  .662 &      (29) &          3318 &              (81) &       .662 &           (18) &  .687 &      (16) & .789 &      (18) &     .612 &         (19) &     .462 &         (15) & .379 &      (17) &       .481 &           (24) \\
          MPNTrack \cite{braso2020learning} &  .588 &      (44) &          1185 &         {\bf (5)} &       .603 &           (40) &  .617 &      (39) & .704 &      (45) &     .596 &         (23) &     .456 &         (16) & .379 &      (18) &       .538 &            (8) \\
                                     * GIT  &  .714 &      (14) &          3501 &              (87) &       .656 &           (20) &  .654 &      (25) & .809 &      (11) &     .663 &    {\bf (5)} &     .489 &    {\bf (5)} & .378 &      (19) &       .468 &           (32) \\
                  TT17 \cite{zhang2020long} &  .549 &      (62) &          1088 &         {\bf (2)} &       .601 &           (42) &  .631 &      (34) & .683 &      (53) &     .571 &         (38) &     .443 &         (24) & .376 &      (20) &       .550 &      {\bf (2)} \\
                                    hugmot  &  .648 &      (30) &          2102 &              (36) &       .613 &           (36) &  .628 &      (36) & .753 &      (31) &     .603 &         (20) &     .445 &         (23) & .372 &      (21) &       .493 &           (22) \\
             * Fair \cite{zhang2020fairmot} &  .737 &       (8) &          3303 &              (80) &       .719 &            (9) &  .723 &       (8) & .818 &       (7) &     .623 &         (16) &     .455 &         (17) & .371 &      (22) &       .453 &           (42) \\
                LSST17 \cite{feng2019multi} &  .547 &      (63) &          1243 &               (8) &       .589 &           (46) &  .623 &      (38) & .681 &      (56) &     .565 &         (42) &     .447 &         (22) & .369 &      (23) &       .543 &            (6) \\
                                       HMM  &  .545 &      (66) &          1172 &         {\bf (4)} &       .601 &           (43) &  .632 &      (33) & .680 &      (57) &     .566 &         (40) &     .436 &         (27) & .366 &      (24) &       .539 &            (7) \\
                                      STMA  &  .588 &      (45) &          1446 &              (13) &       .596 &           (44) &  .616 &      (40) & .700 &      (48) &     .578 &         (33) &     .431 &         (29) & .361 &      (25) &       .516 &           (13) \\
                                   * FMMOT  &  .722 &      (11) &          2199 &              (45) &       .719 &           (10) &  .723 &       (9) & .801 &      (13) &     .615 &         (17) &     .438 &         (26) & .360 &      (26) &       .450 &           (45) \\
                               SLA\_public  &  .597 &      (42) &          1647 &              (16) &       .608 &           (38) &  .634 &      (32) & .717 &      (39) &     .586 &         (28) &     .440 &         (25) & .360 &      (27) &       .502 &           (15) \\
       * CSTrack \cite{liang2020rethinking} &  .749 &       (6) &          3567 &              (90) &       .721 &            (6) &  .726 &       (6) & .830 &       (6) &     .627 &         (13) &     .450 &         (20) & .356 &      (28) &       .429 &           (56) \\
                                 * CLTSMOT  &  .671 &      (26) &          4983 &             (109) &       .634 &           (30) &  .651 &      (27) & .794 &      (15) &     .601 &         (21) &     .435 &         (28) & .350 &      (29) &       .440 &           (50) \\
                               ISDH\_HDAv2  &  .545 &      (65) &          3010 &              (71) &       .618 &           (33) &  .659 &      (22) & .698 &      (50) &     .555 &         (49) &     .411 &         (32) & .348 &      (30) &       .498 &           (18) \\
                              UNS20regress  &  .568 &      (51) &          1320 &              (10) &       .573 &           (50) &  .583 &      (50) & .681 &      (55) &     .562 &         (43) &     .410 &         (35) & .341 &      (31) &       .501 &           (16) \\
                                       EMT  &  .556 &      (58) &          1361 &              (11) &       .558 &           (57) &  .571 &      (59) & .667 &      (73) &     .539 &         (55) &     .406 &         (40) & .333 &      (32) &       .499 &           (17) \\
                                     ALBOD  &  .569 &      (50) &          2011 &              (31) &       .572 &           (51) &  .587 &      (47) & .721 &      (37) &     .576 &         (36) &     .407 &         (37) & .328 &      (33) &       .456 &           (39) \\
                               ISE\_MOT17R  &  .601 &      (40) &          2556 &              (59) &       .559 &           (56) &  .564 &      (64) & .729 &      (36) &     .579 &         (32) &     .400 &         (46) & .326 &      (34) &       .447 &           (46) \\
                                 * GMTrack  &  .638 &      (33) &          1893 &              (26) &       .656 &           (19) &  .665 &      (20) & .720 &      (38) &     .566 &         (41) &     .404 &         (42) & .325 &      (35) &       .451 &           (44) \\
                                  FGRNetIV  &  .566 &      (53) &          1722 &              (18) &       .557 &           (59) &  .572 &      (57) & .679 &      (59) &     .544 &         (53) &     .395 &         (47) & .324 &      (36) &       .478 &           (26) \\
                                       CMT  &  .518 &      (80) &          1217 &               (7) &       .580 &           (49) &  .607 &      (44) & .657 &      (80) &     .542 &         (54) &     .411 &         (31) & .324 &      (37) &       .494 &           (21) \\
        * CTTrack17 \cite{zhou2020tracking} &  .678 &      (22) &          3039 &              (72) &       .651 &           (22) &  .647 &      (28) & .763 &      (29) &     .583 &         (30) &     .406 &         (39) & .322 &      (38) &       .422 &           (59) \\
            GSM\_Tracktor \cite{liu2020gsm} &  .564 &      (54) &          1485 &              (14) &       .565 &           (55) &  .578 &      (54) & .680 &      (58) &     .548 &         (51) &     .391 &         (49) & .321 &      (39) &       .472 &           (29) \\
             HDTR \cite{babaee2018multiple} &  .541 &      (68) &          1895 &              (27) &       .566 &           (54) &  .595 &      (46) & .682 &      (54) &     .559 &         (47) &     .410 &         (34) & .318 &      (40) &       .466 &           (33) \\
           * FUFET \cite{shan2020tracklets} &  .762 & {\bf (5)} &          3237 &              (79) &       .700 &           (12) &  .680 &      (18) & .846 & {\bf (3)} &     .630 &         (12) &     .406 &         (38) & .316 &      (41) &       .373 &           (86) \\
                     TPM \cite{peng2020tpm} &  .542 &      (67) &          1824 &              (21) &       .518 &           (78) &  .526 &      (81) & .670 &      (68) &     .544 &         (52) &     .385 &         (51) & .315 &      (42) &       .471 &           (30) \\
                                     TrajE  &  .674 &      (24) &          4019 &              (96) &       .618 &           (34) &  .612 &      (42) & .763 &      (28) &     .579 &         (31) &     .401 &         (45) & .314 &      (43) &       .412 &           (68) \\
                                      SSAT  &  .620 &      (35) &          1850 &              (23) &       .618 &           (35) &  .626 &      (37) & .713 &      (40) &     .560 &         (45) &     .392 &         (48) & .314 &      (44) &       .441 &           (49) \\
       eHAF17 \cite{sheng2018heterogeneous} &  .518 &      (81) &          1834 &              (22) &       .535 &           (70) &  .547 &      (70) & .665 &      (75) &     .538 &         (56) &     .387 &         (50) & .314 &      (45) &       .472 &           (28) \\
                                  TrajEocc  &  .678 &      (23) &          3475 &              (85) &       .623 &           (32) &  .614 &      (41) & .767 &      (27) &     .586 &         (27) &     .404 &         (44) & .314 &      (46) &       .409 &           (70) \\
                             mcmt\_icv\_v3  &  .568 &      (52) &          2055 &              (33) &       .609 &           (37) &  .656 &      (23) & .690 &      (51) &     .521 &         (62) &     .377 &         (57) & .313 &      (47) &       .454 &           (40) \\
                                  HTracker  &  .669 &      (27) &          4806 &             (106) &       .675 &           (17) &  .704 &      (13) & .790 &      (17) &     .576 &         (35) &     .411 &         (33) & .313 &      (48) &       .397 &           (75) \\
                  DMAN \cite{zhu2018online} &  .482 &     (101) &          2194 &              (44) &       .532 &           (72) &  .557 &      (67) & .631 &     (100) &     .506 &         (75) &     .378 &         (56) & .312 &      (49) &       .495 &           (20) \\
                             * ShallowSORT  &  .599 &      (41) &          3045 &              (73) &       .558 &           (58) &  .566 &      (62) & .738 &      (35) &     .569 &         (39) &     .404 &         (43) & .310 &      (50) &       .420 &           (61) \\

\bottomrule
\end{tabular}
}
\label{tab:mot17-extended-unpub}
\end{table*}

\clearpage

\clearpage\onecolumn
\subsection{Temporal error decomposition}

The time-varying error decomposition for the 11 selected trackers are shown below.
Split errors generally occur at shorter temporal horizons than merge errors and in much greater proportion.
For several trackers, the fraction of false-positive detection errors increases with the fraction of split errors.
While it may seem counter-intuitive for the fraction of detection errors to depend on the temporal horizon, it is a natural effect of there being more predicted tracks than ground-truth tracks.

\begin{figure*}[h!]
\centering
\includegraphics[width=0.32\textwidth]{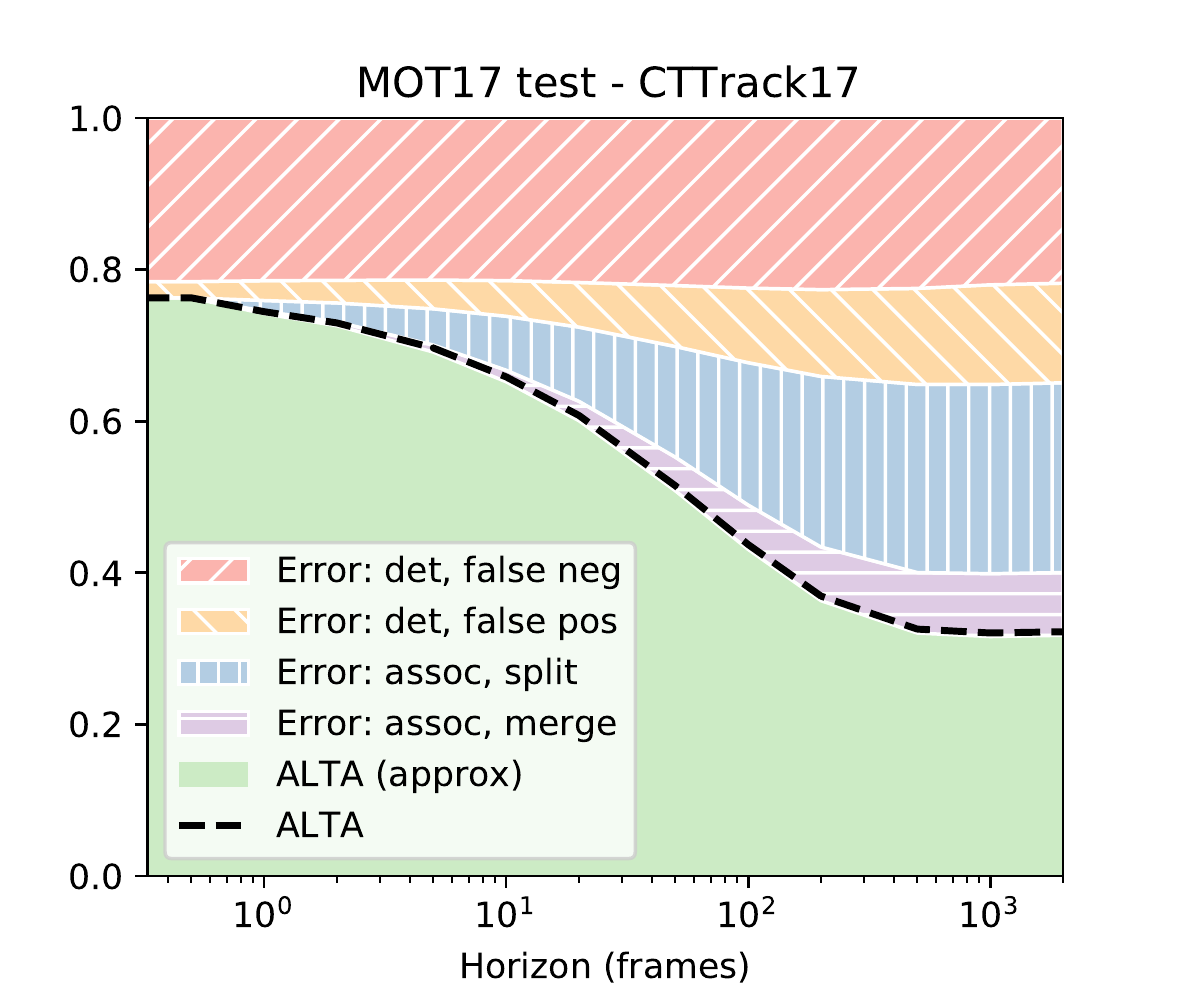}
\includegraphics[width=0.32\textwidth]{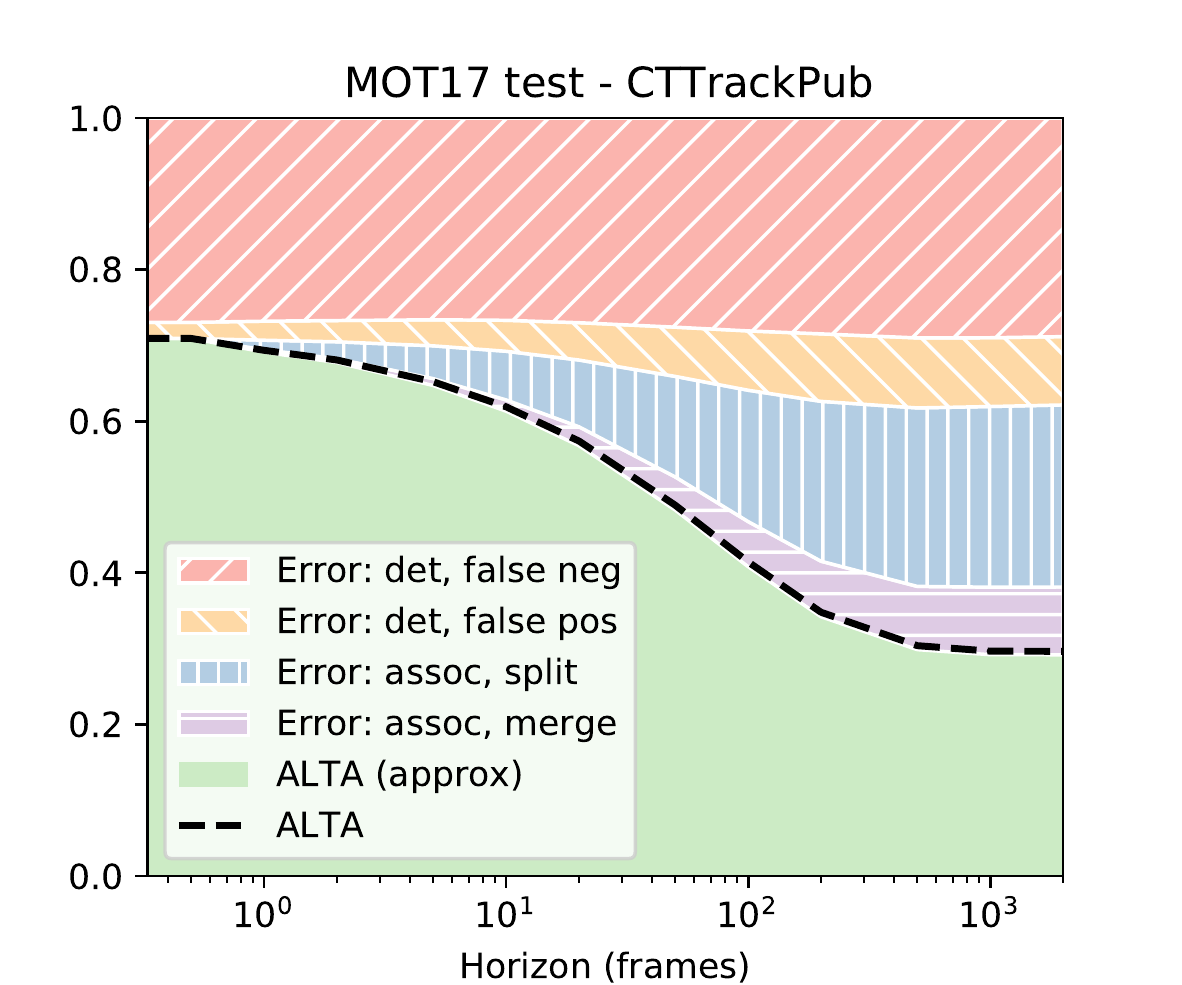}
\includegraphics[width=0.32\textwidth]{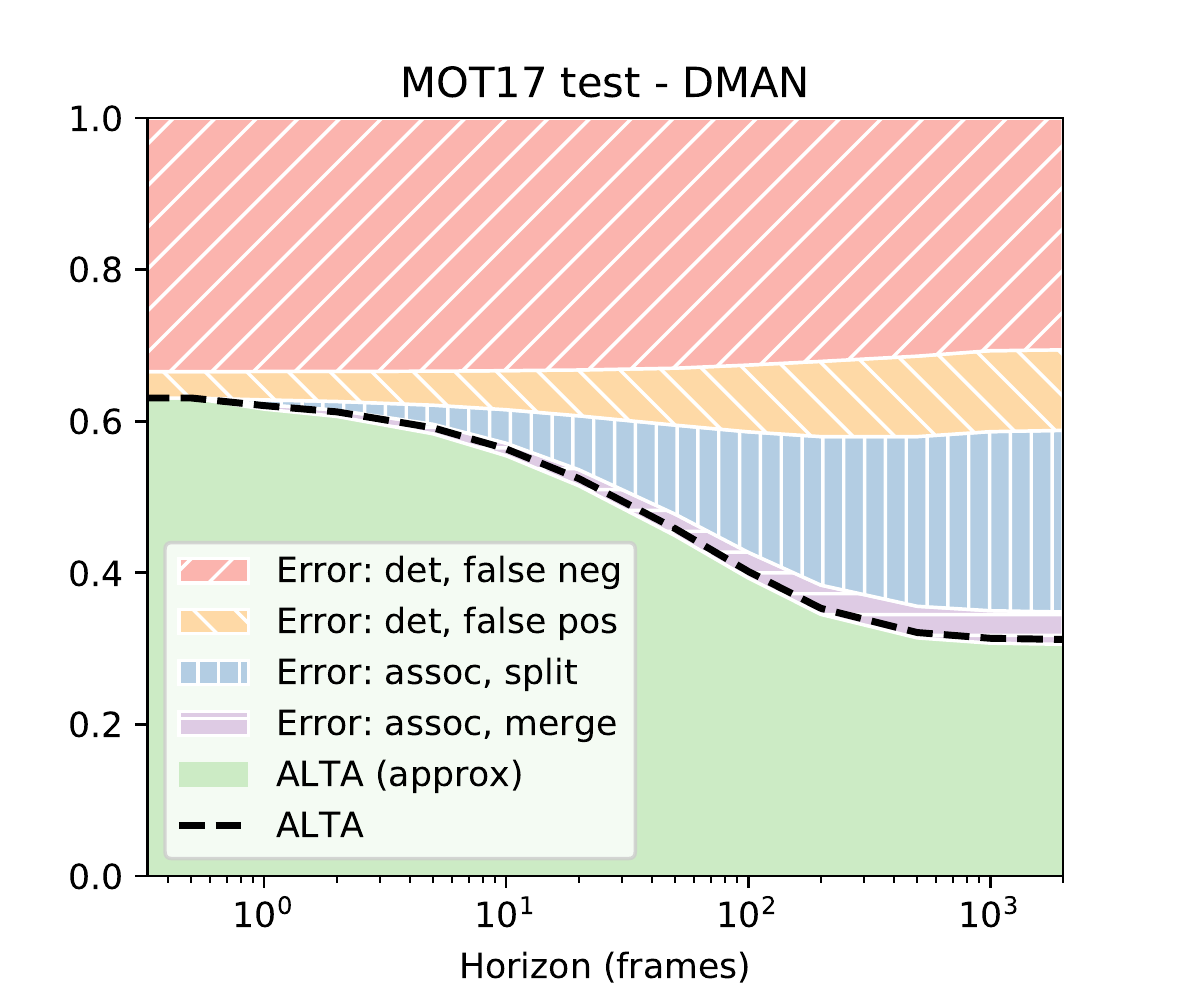}
\includegraphics[width=0.32\textwidth]{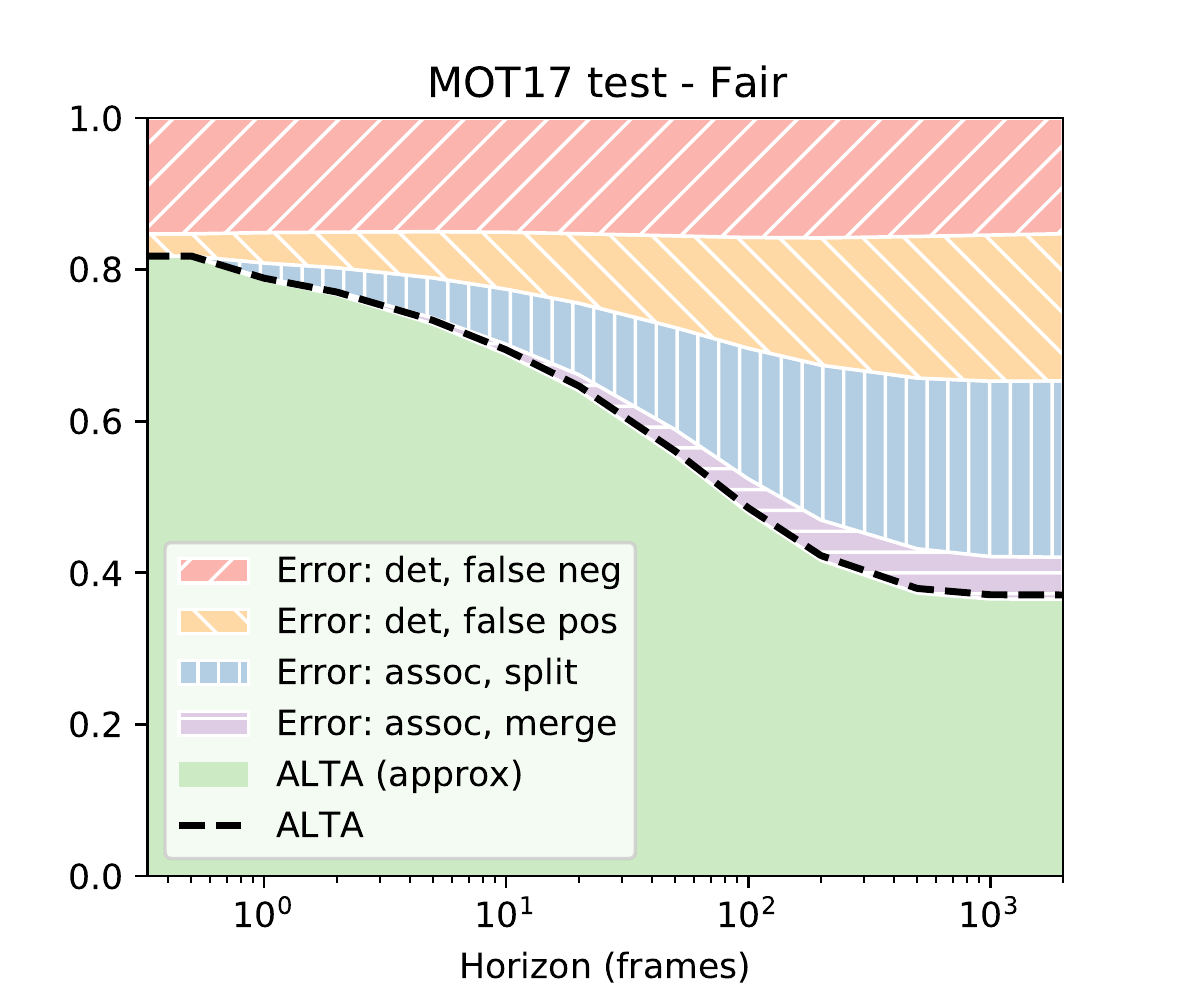}
\includegraphics[width=0.32\textwidth]{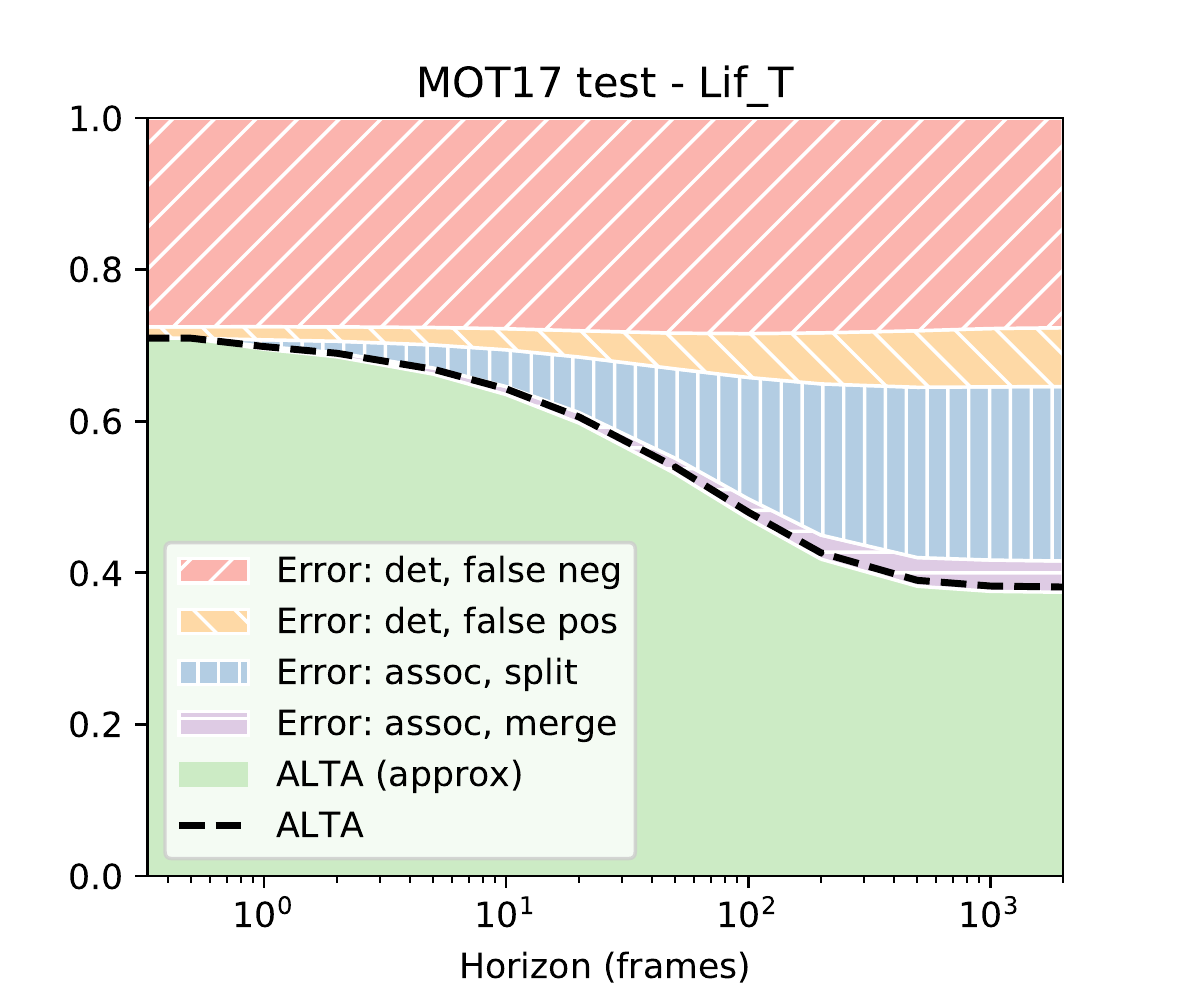}
\includegraphics[width=0.32\textwidth]{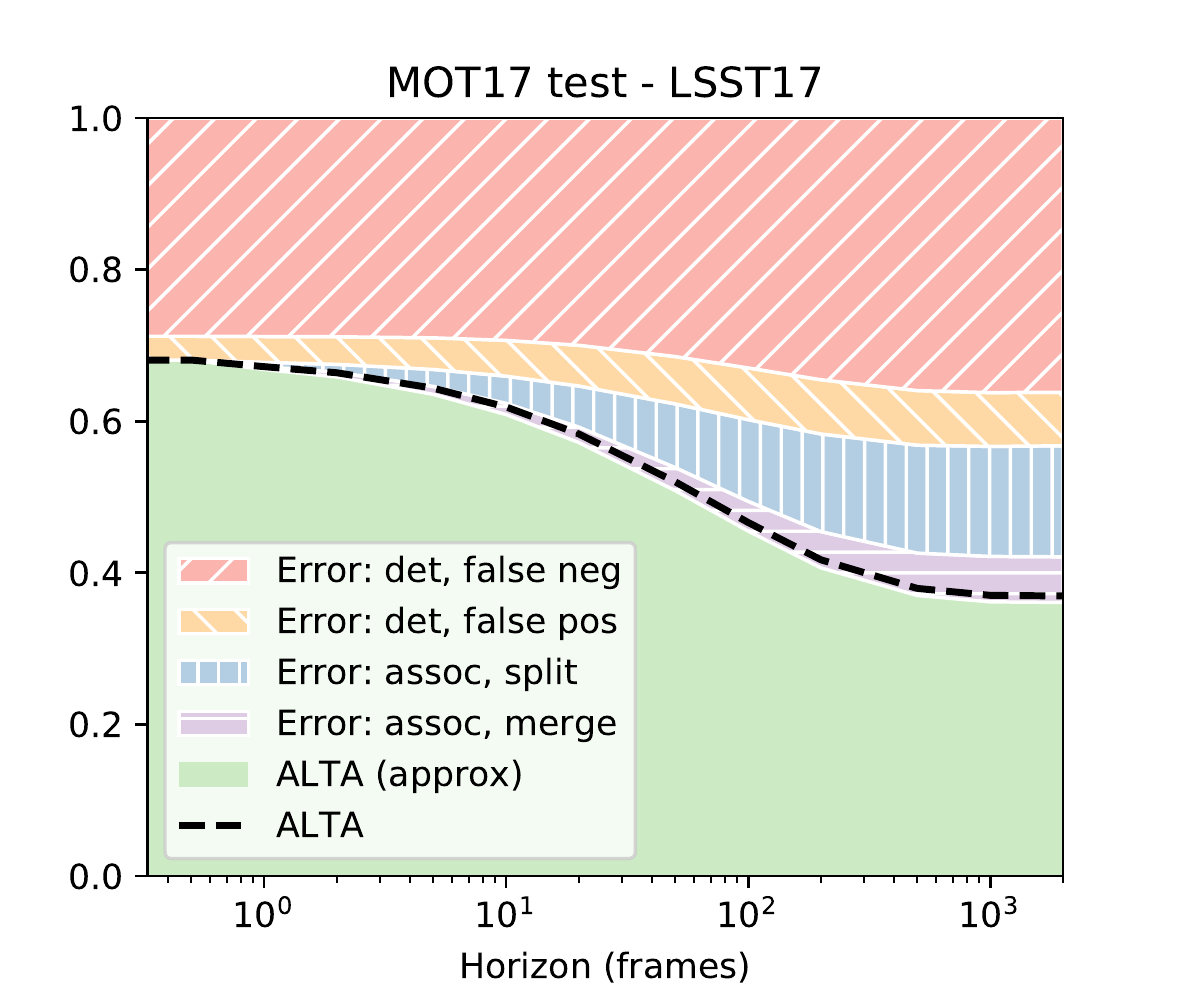}
\includegraphics[width=0.32\textwidth]{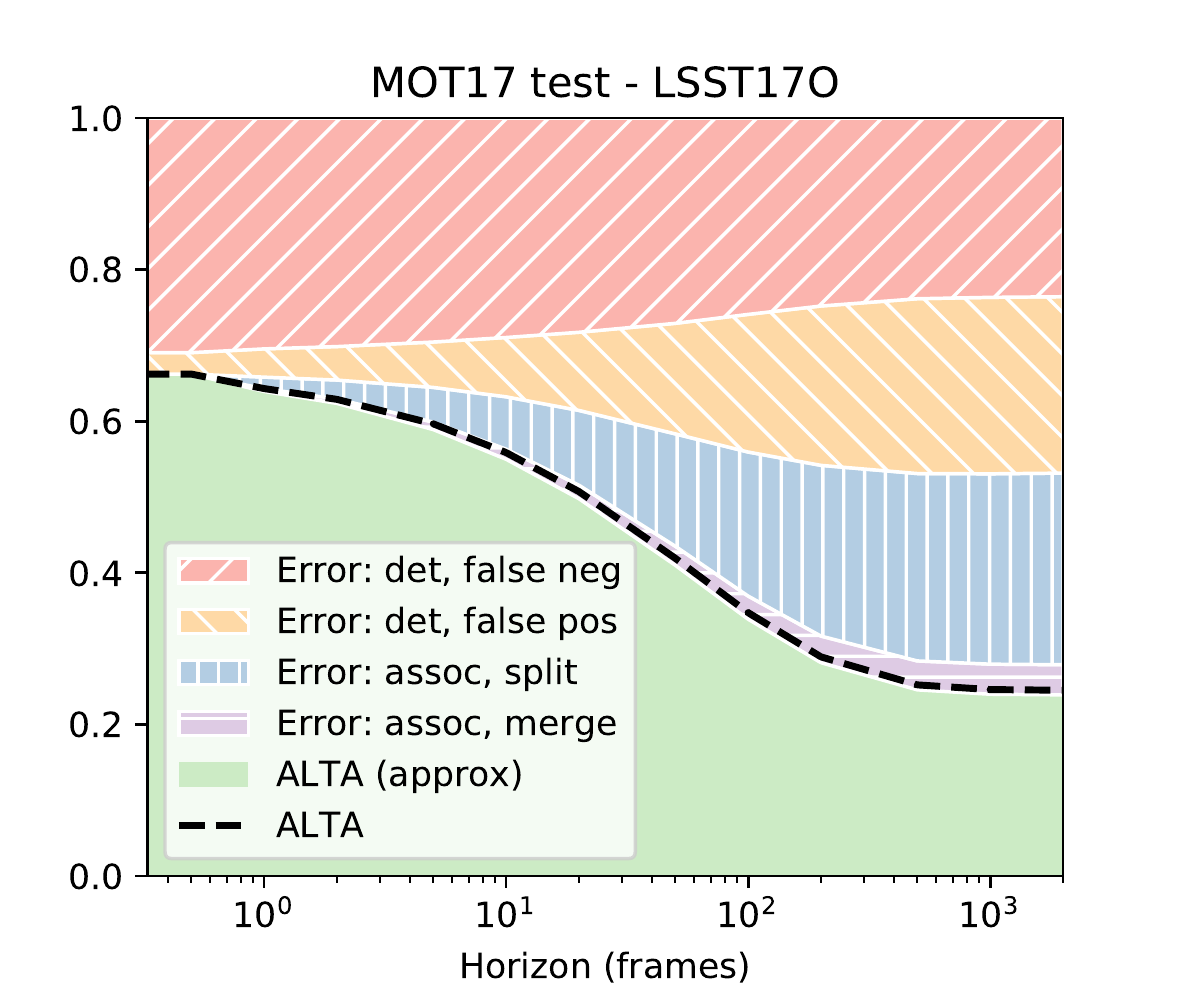}
\includegraphics[width=0.32\textwidth]{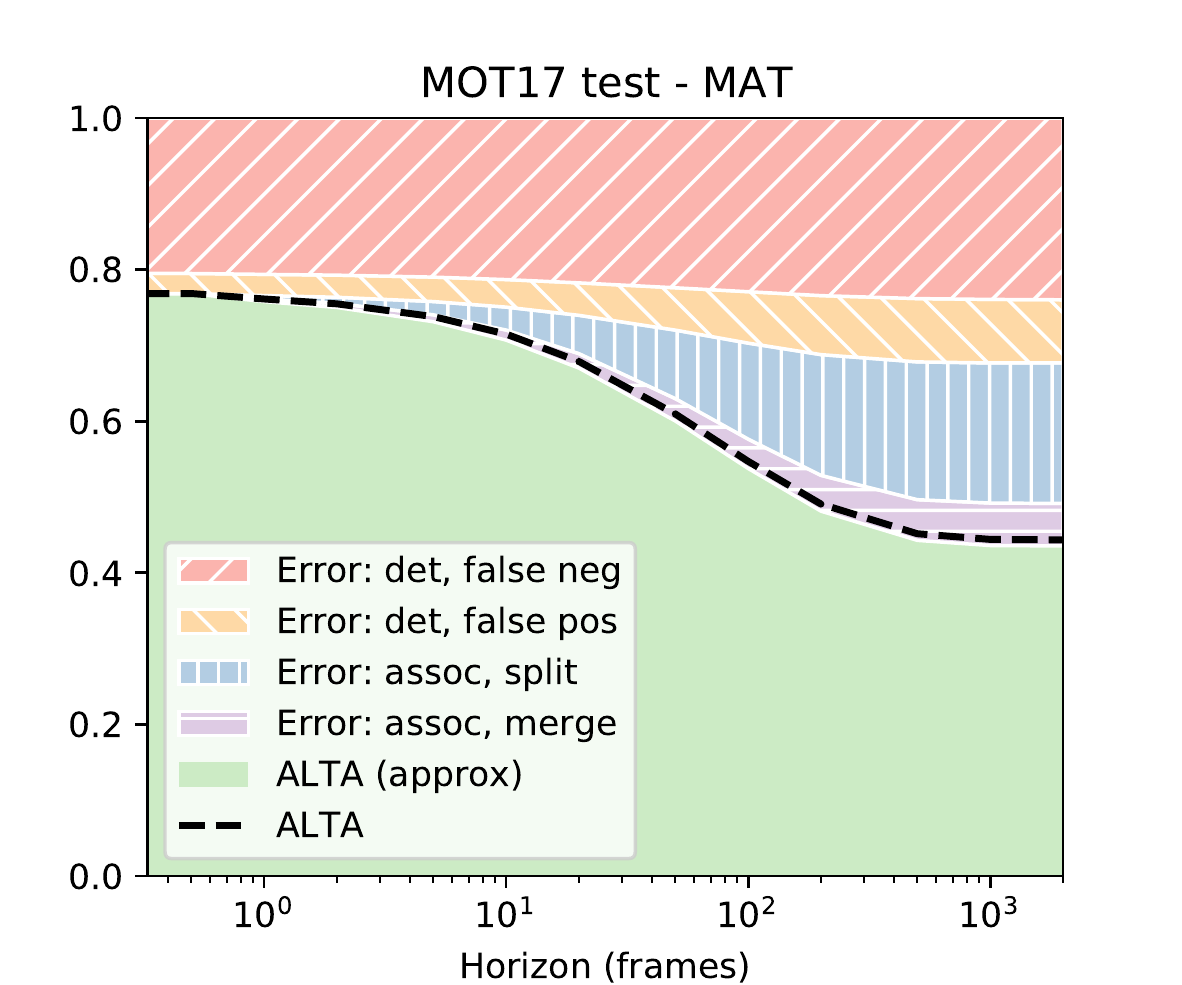}
\includegraphics[width=0.32\textwidth]{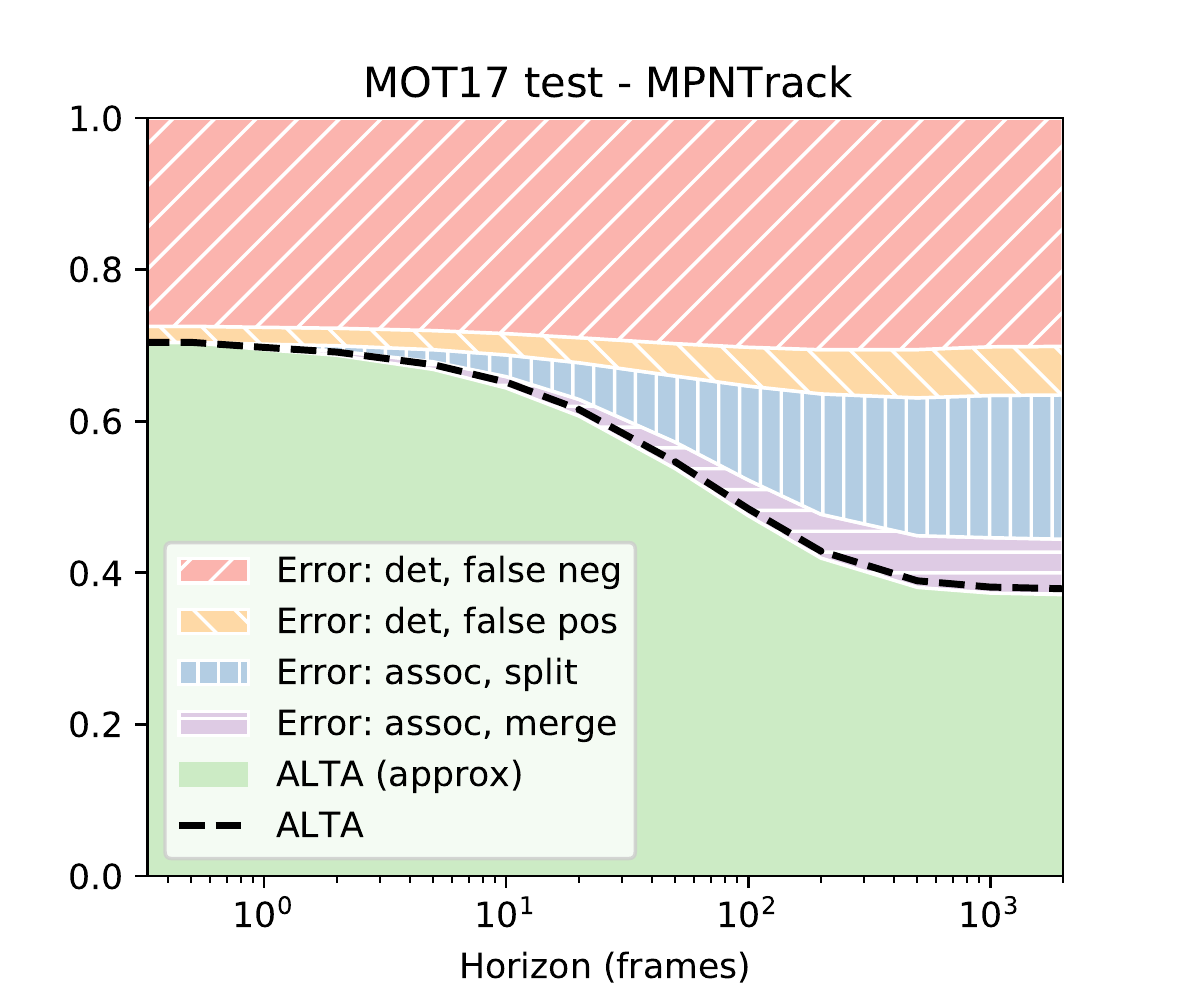}
\includegraphics[width=0.32\textwidth]{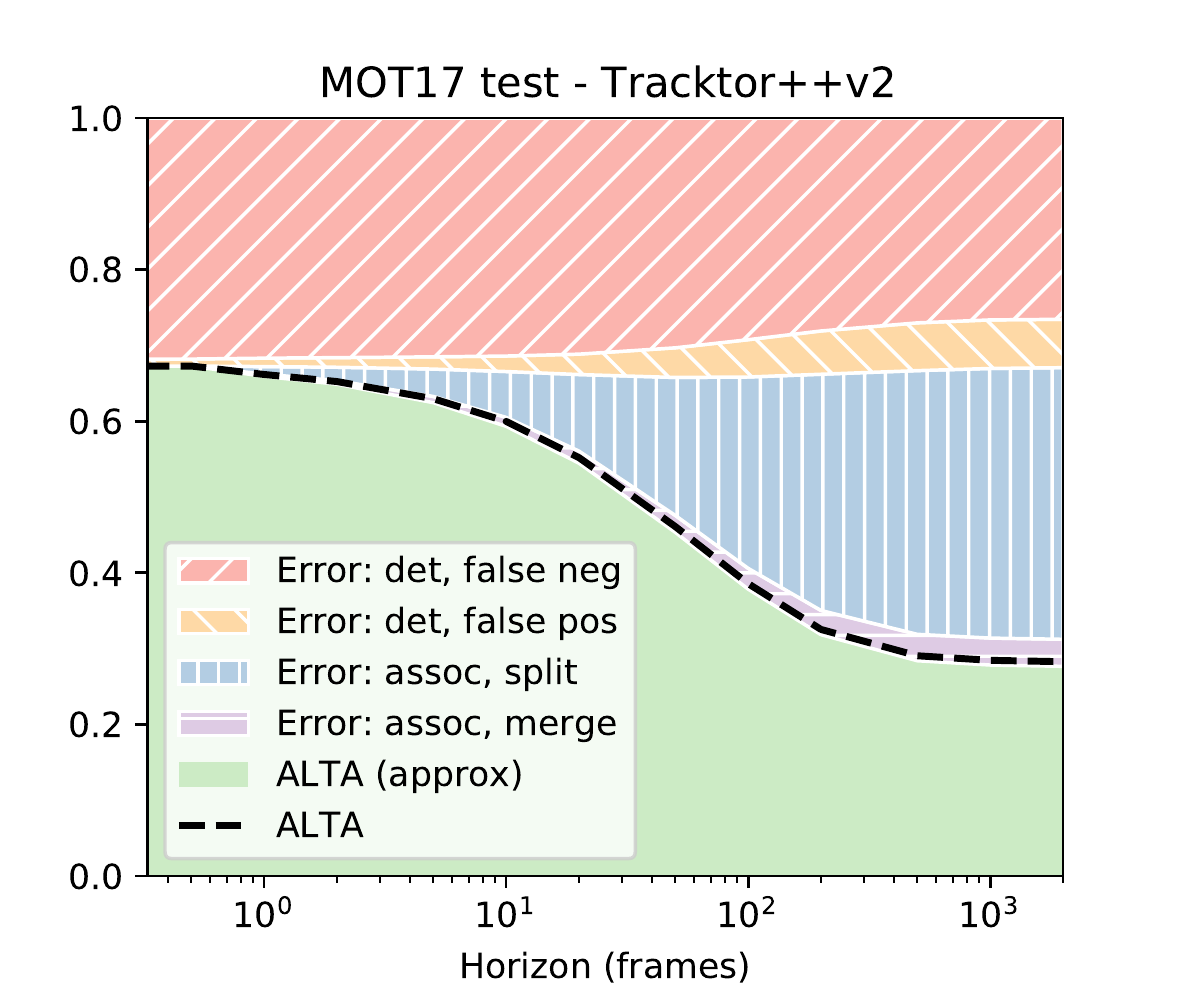}
\includegraphics[width=0.32\textwidth]{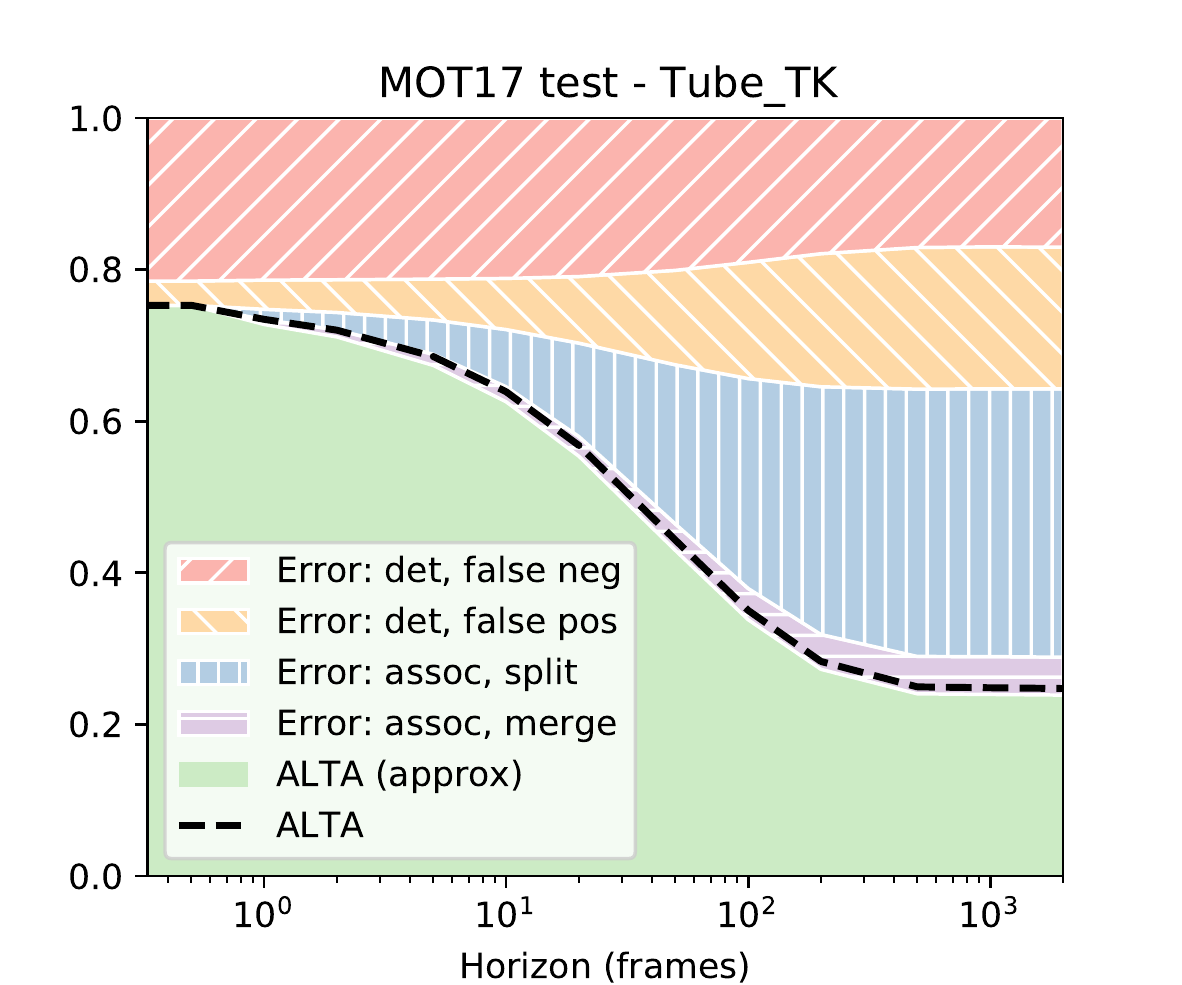}
\caption{Time-varying error decomposition of ALTA for selected trackers.}
\end{figure*}

\clearpage\onecolumn
\subsection{Per-sequence temporal analysis}

The per-sequence results for the 11 selected trackers are shown below.
The 3 columns correspond to the 3 different sets of public detections (DPM, FRCNN, SDP).
The methods which use their own, external detector have the same results for each set of detections.
Note that ALTA is not necessarily monotonically decreasing.
It may occur that a tracker achieves a higher ALTA score at a longer horizon if it is able to recover from incorrect associations.

\begin{figure*}[h!]
\centering
\includegraphics[width=0.4\textwidth]{figures/horizon_frames_ata_value_MOT17_test_selected.pdf}
\\
\includegraphics[width=0.32\textwidth]{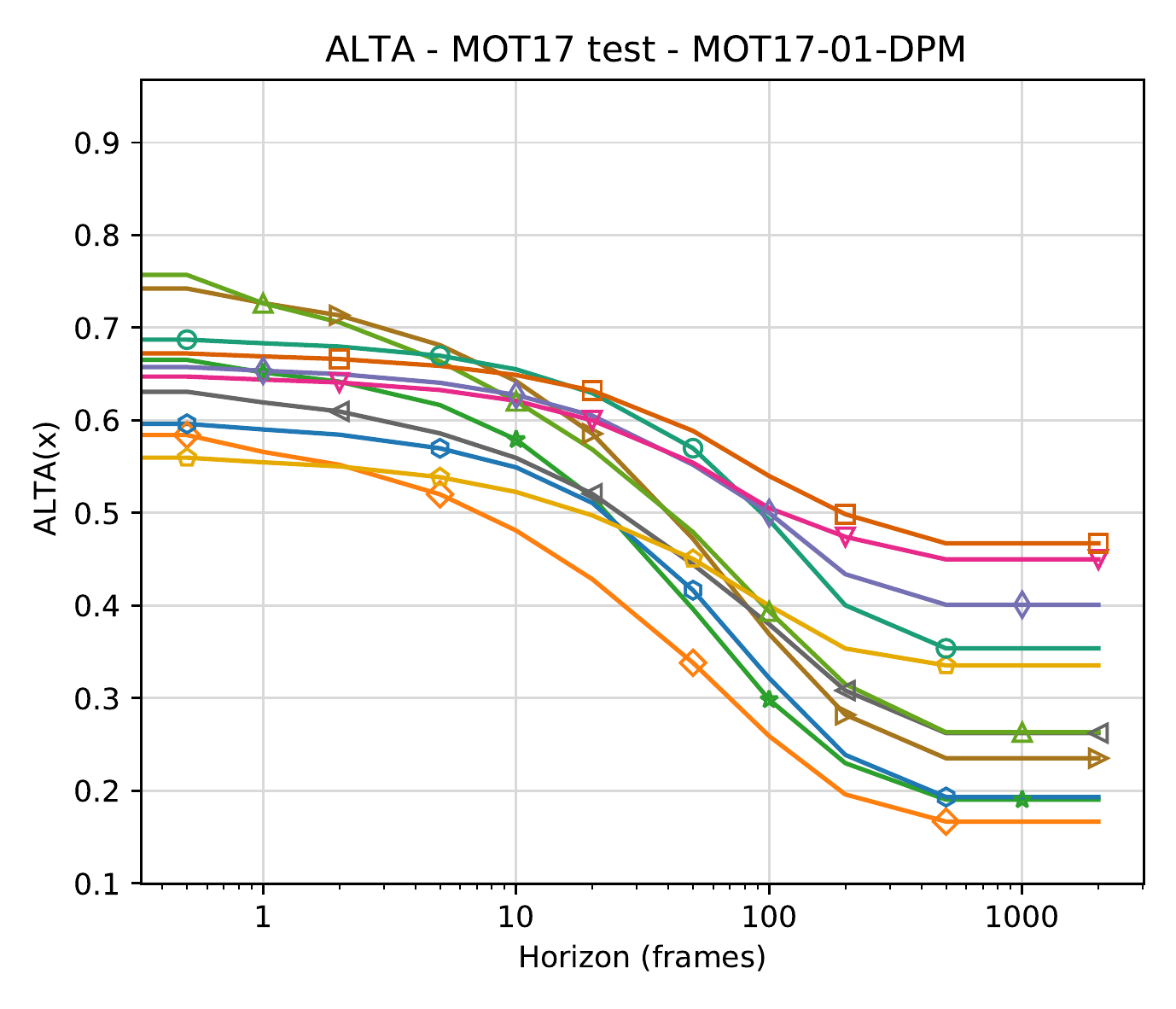}
\includegraphics[width=0.32\textwidth]{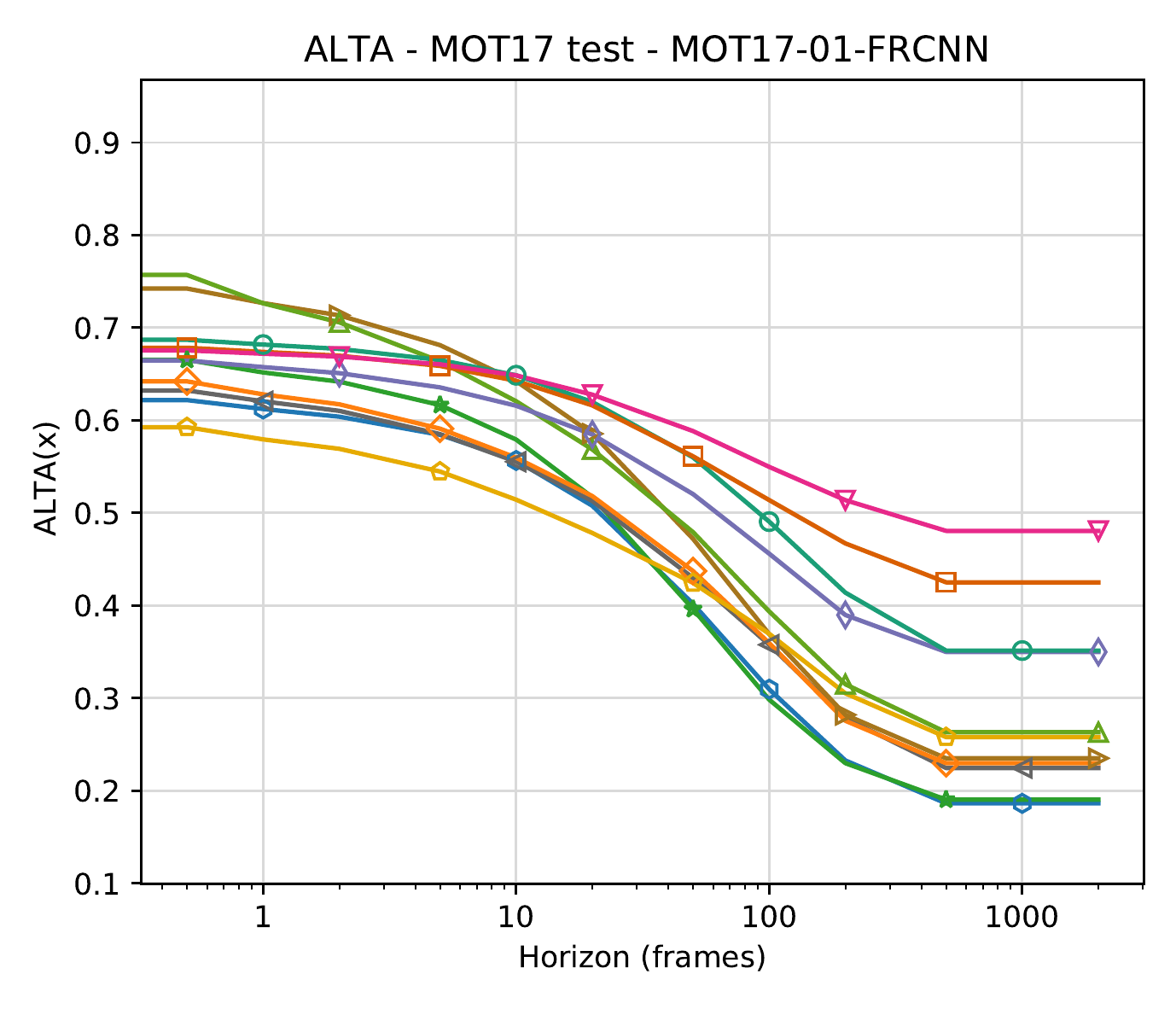}
\includegraphics[width=0.32\textwidth]{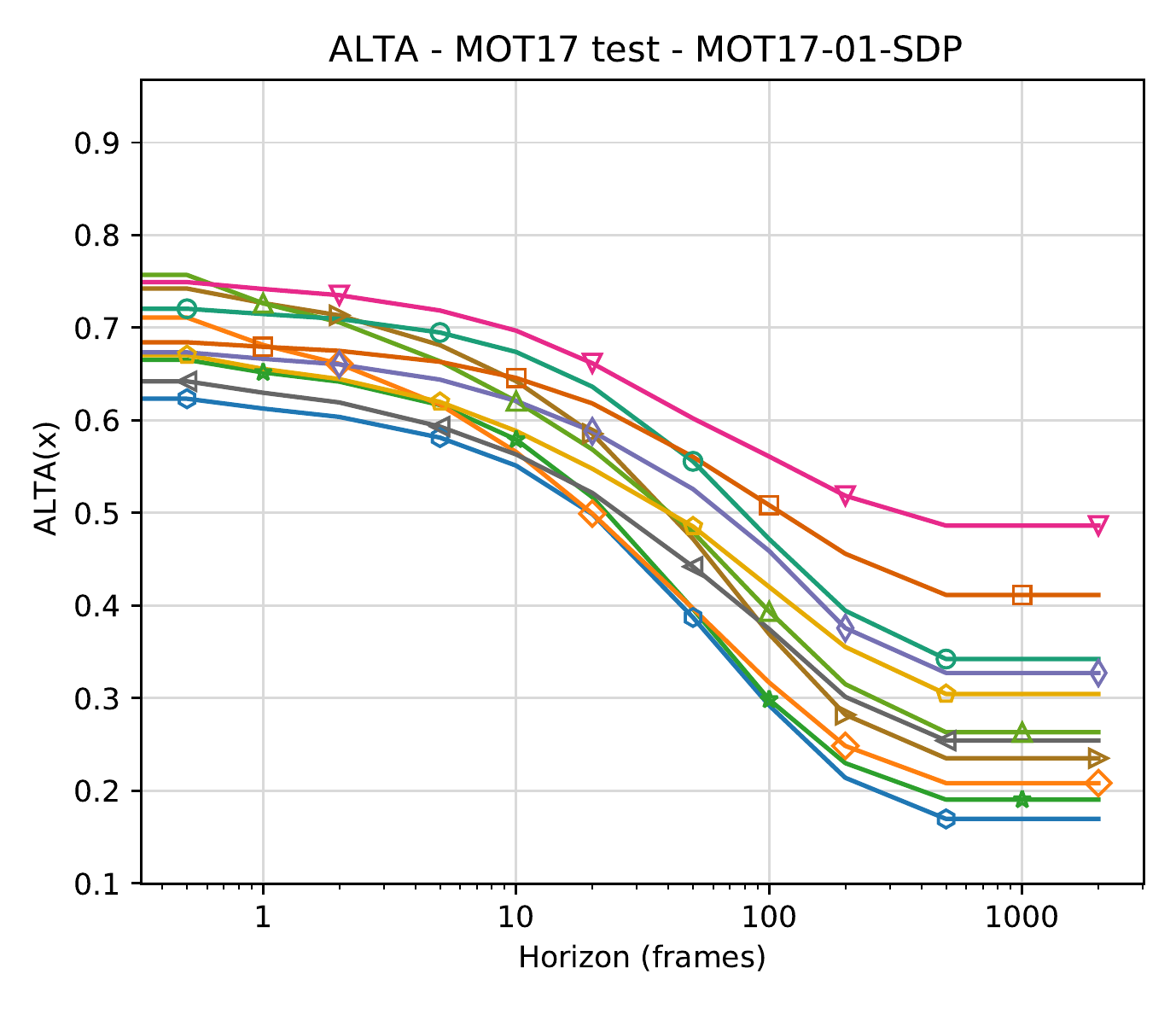}
\\
\includegraphics[width=0.32\textwidth]{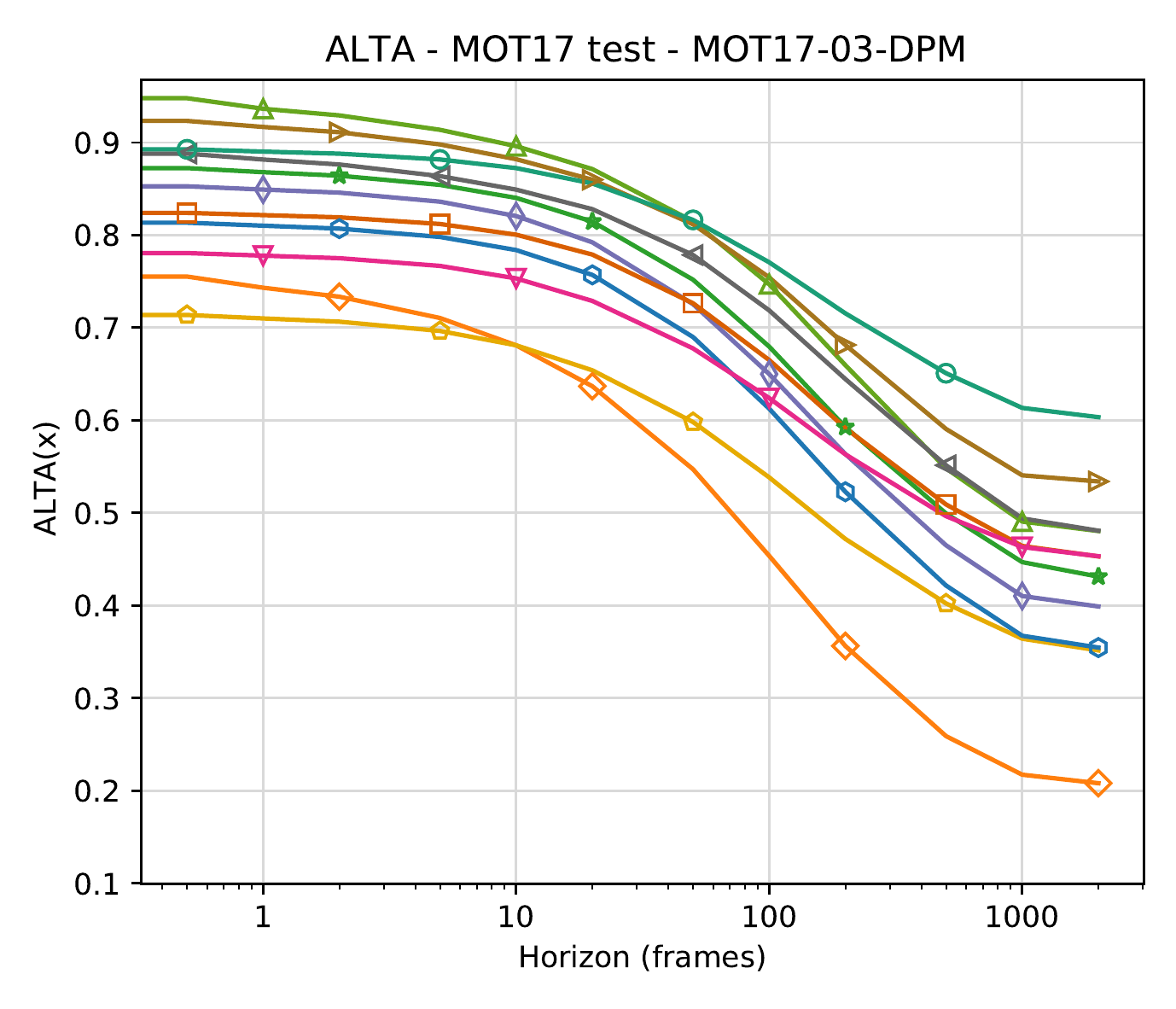}
\includegraphics[width=0.32\textwidth]{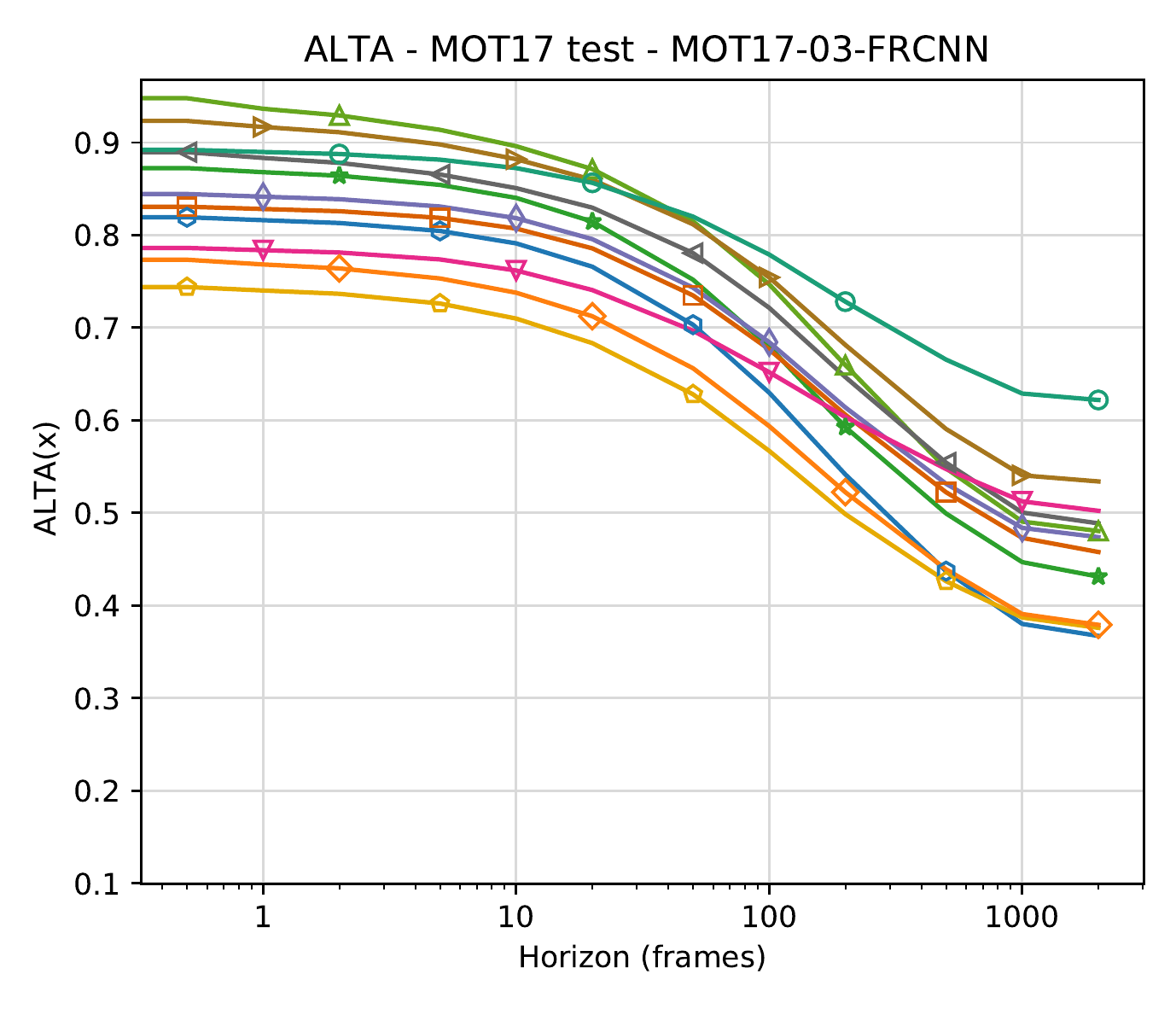}
\includegraphics[width=0.32\textwidth]{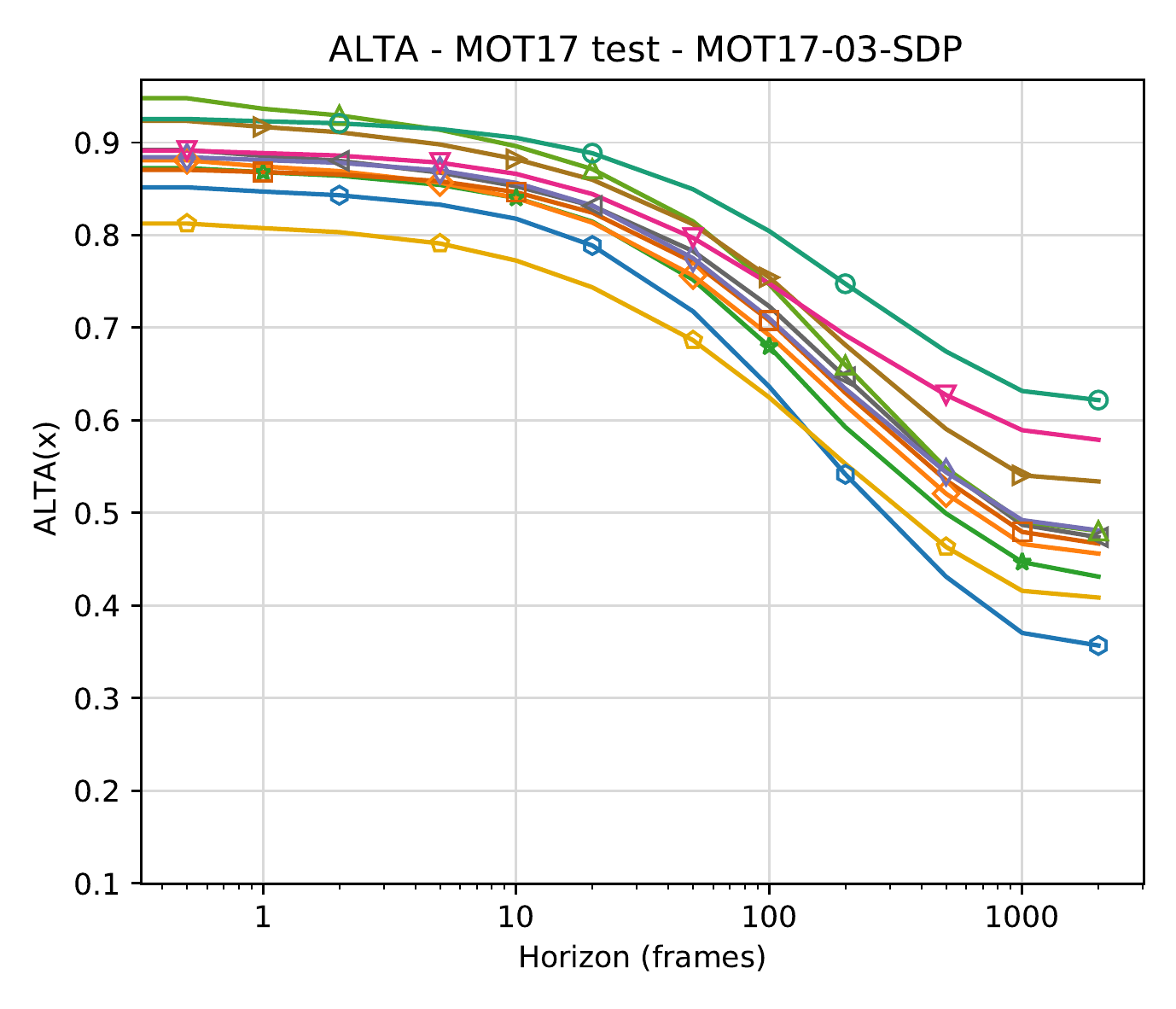}
\\
\includegraphics[width=0.32\textwidth]{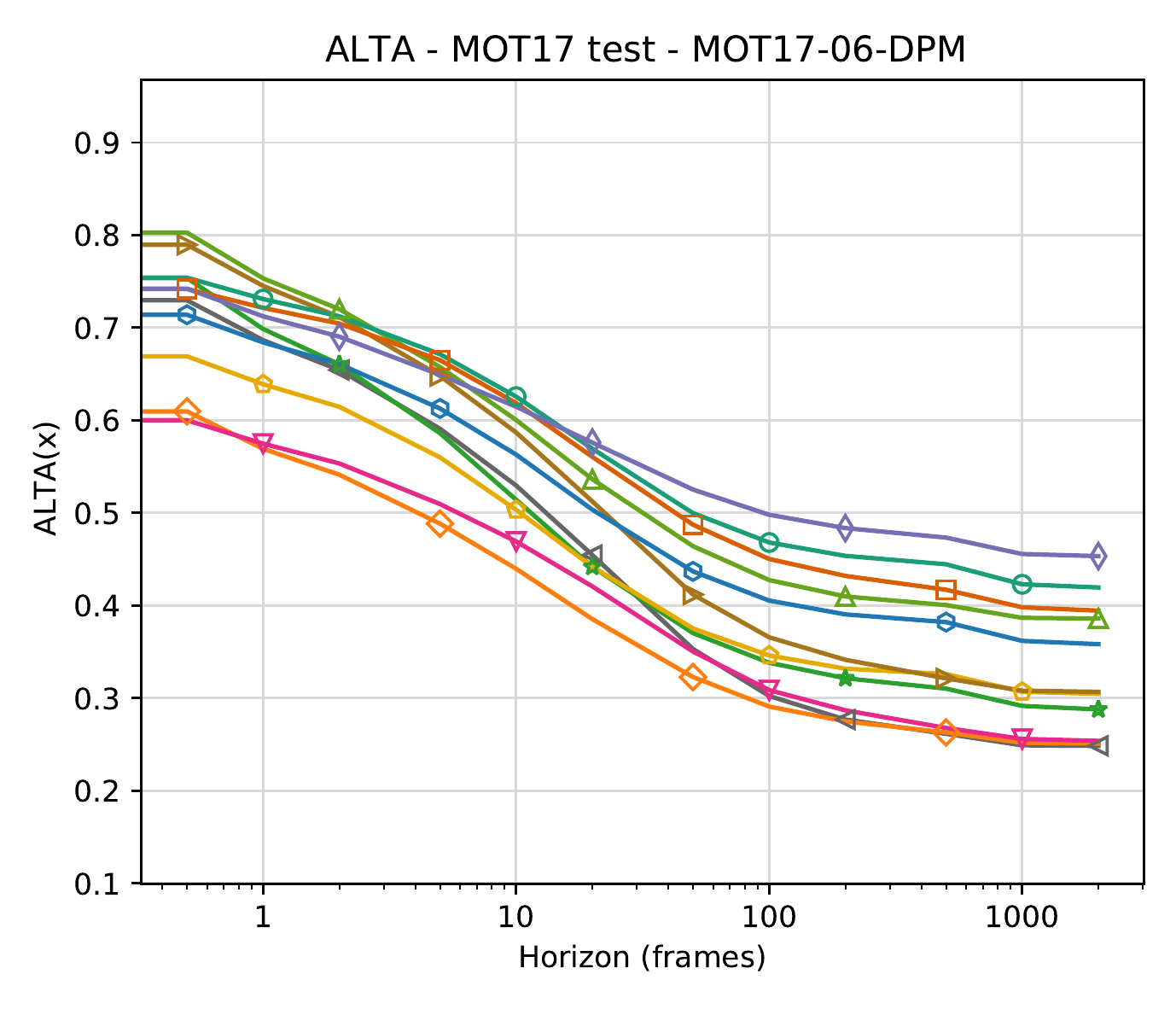}
\includegraphics[width=0.32\textwidth]{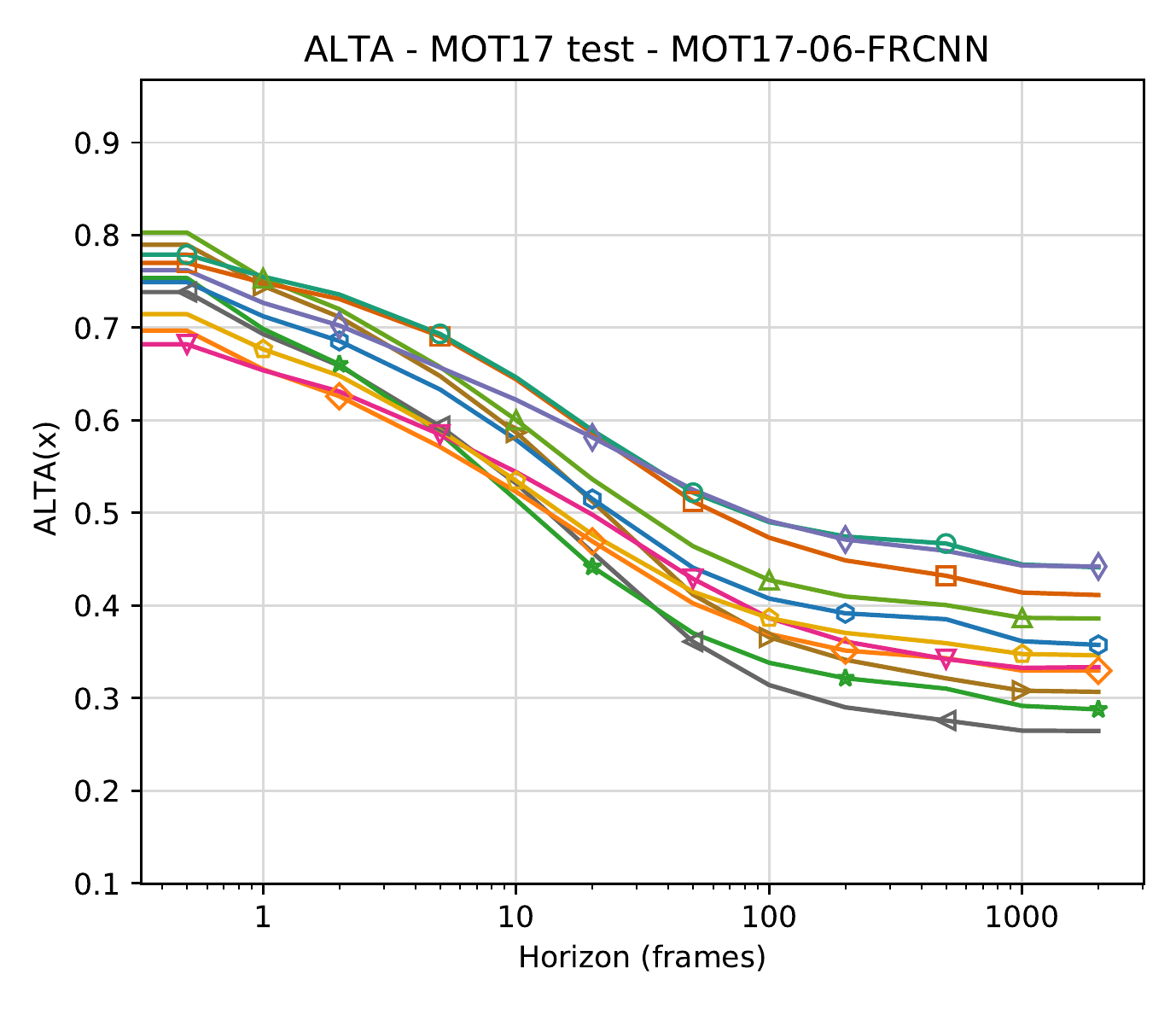}
\includegraphics[width=0.32\textwidth]{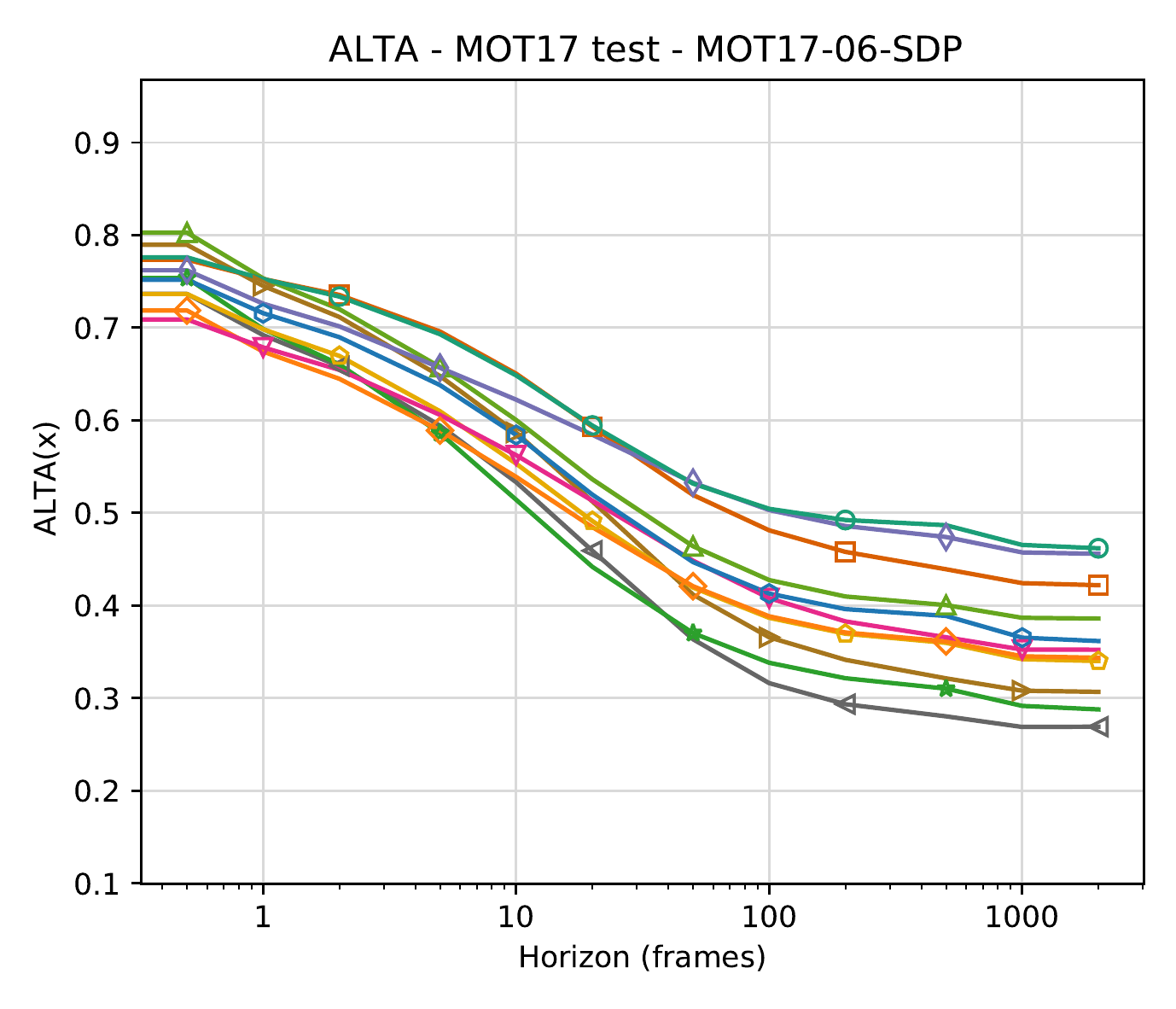}
\end{figure*}

\begin{figure*}[h!]
\centering
\includegraphics[width=0.32\textwidth]{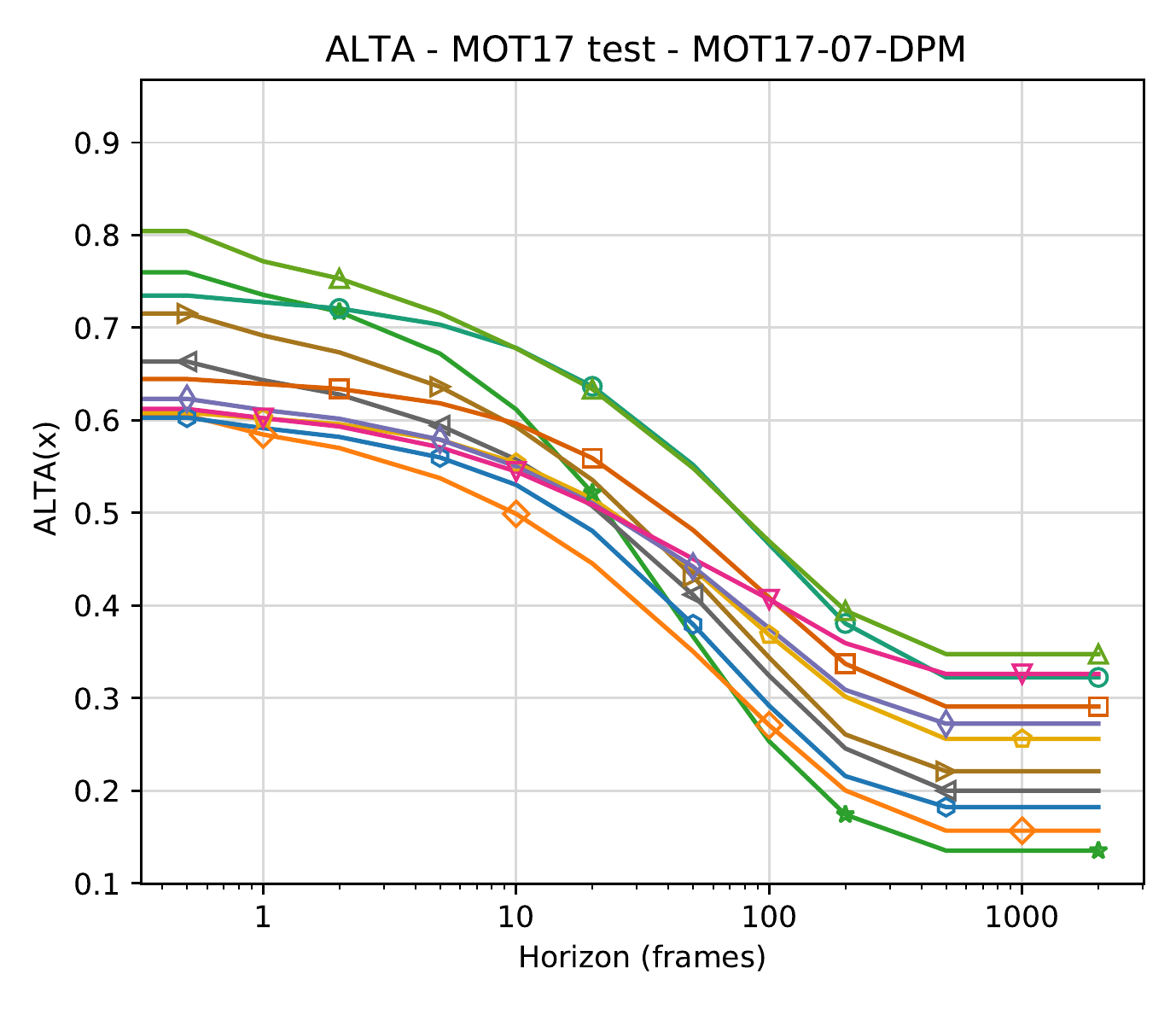}
\includegraphics[width=0.32\textwidth]{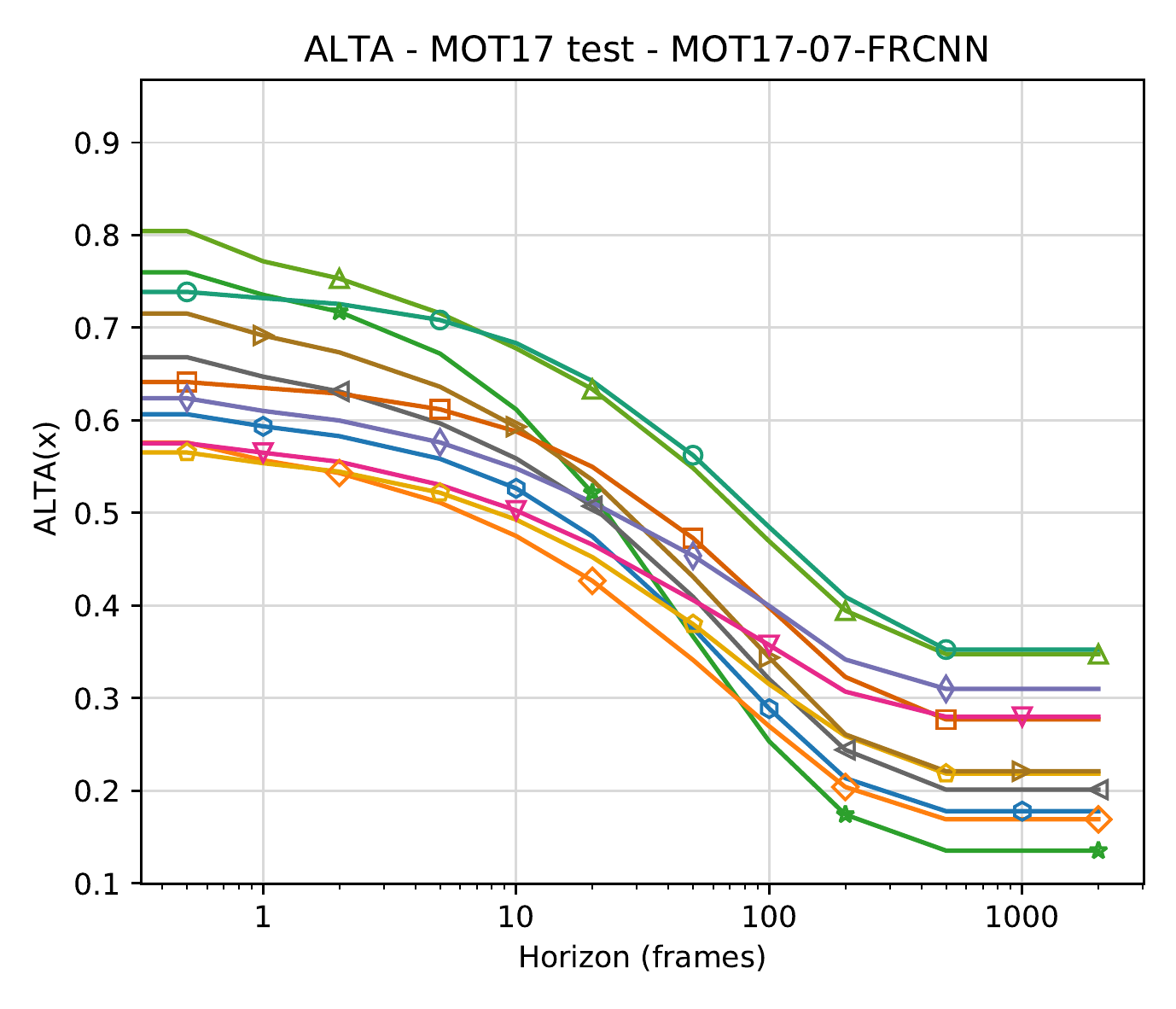}
\includegraphics[width=0.32\textwidth]{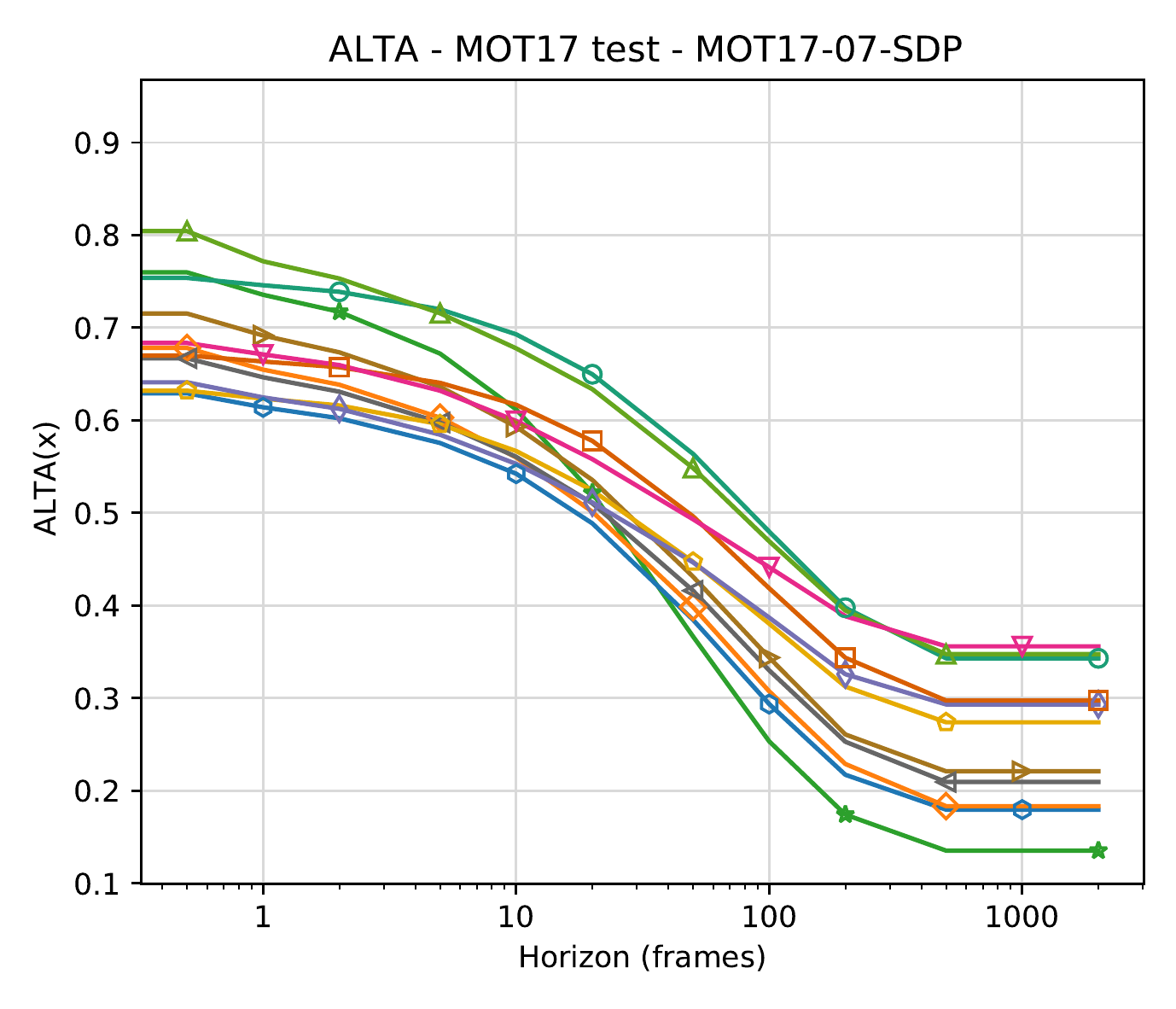}
\\
\includegraphics[width=0.32\textwidth]{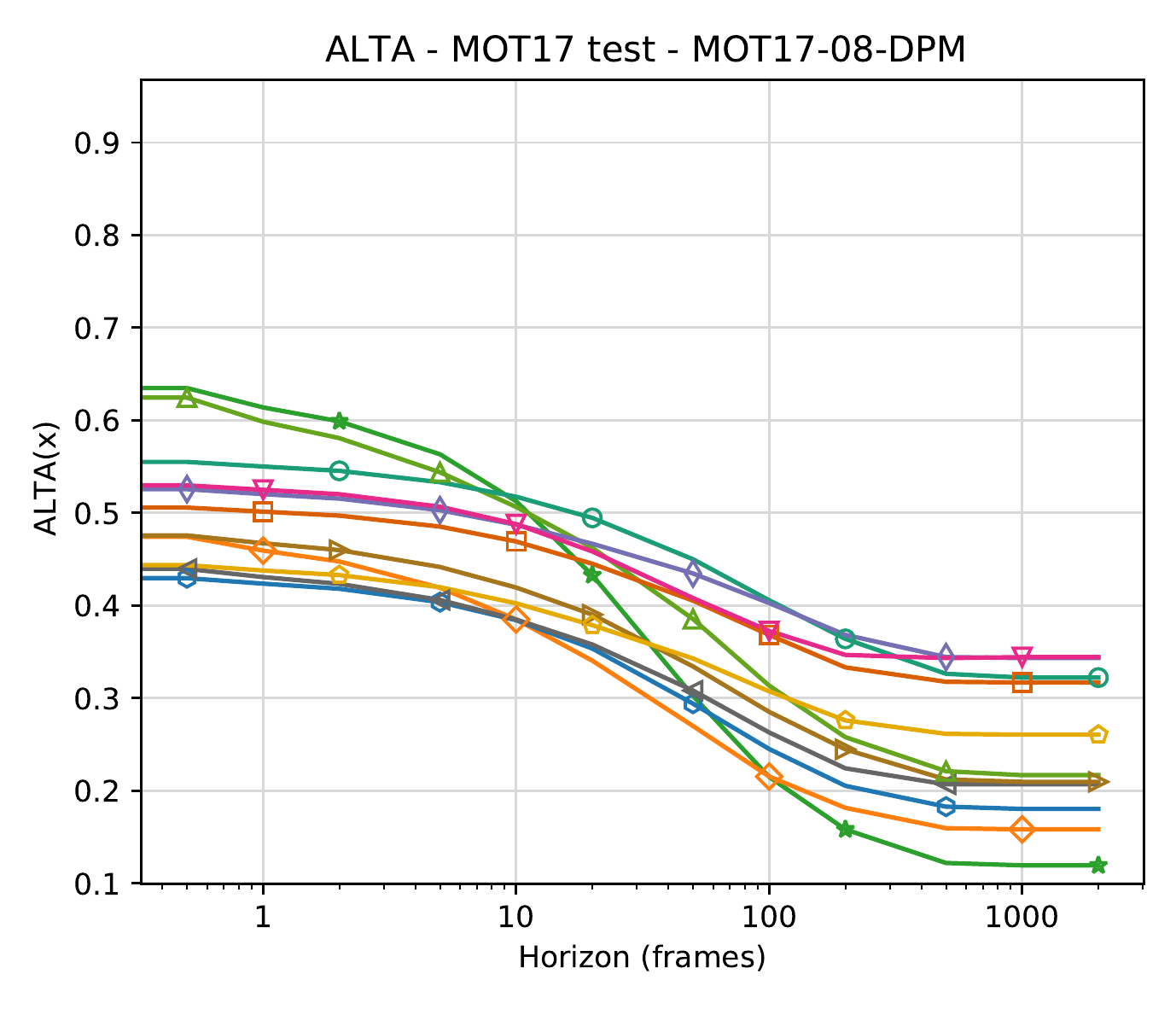}
\includegraphics[width=0.32\textwidth]{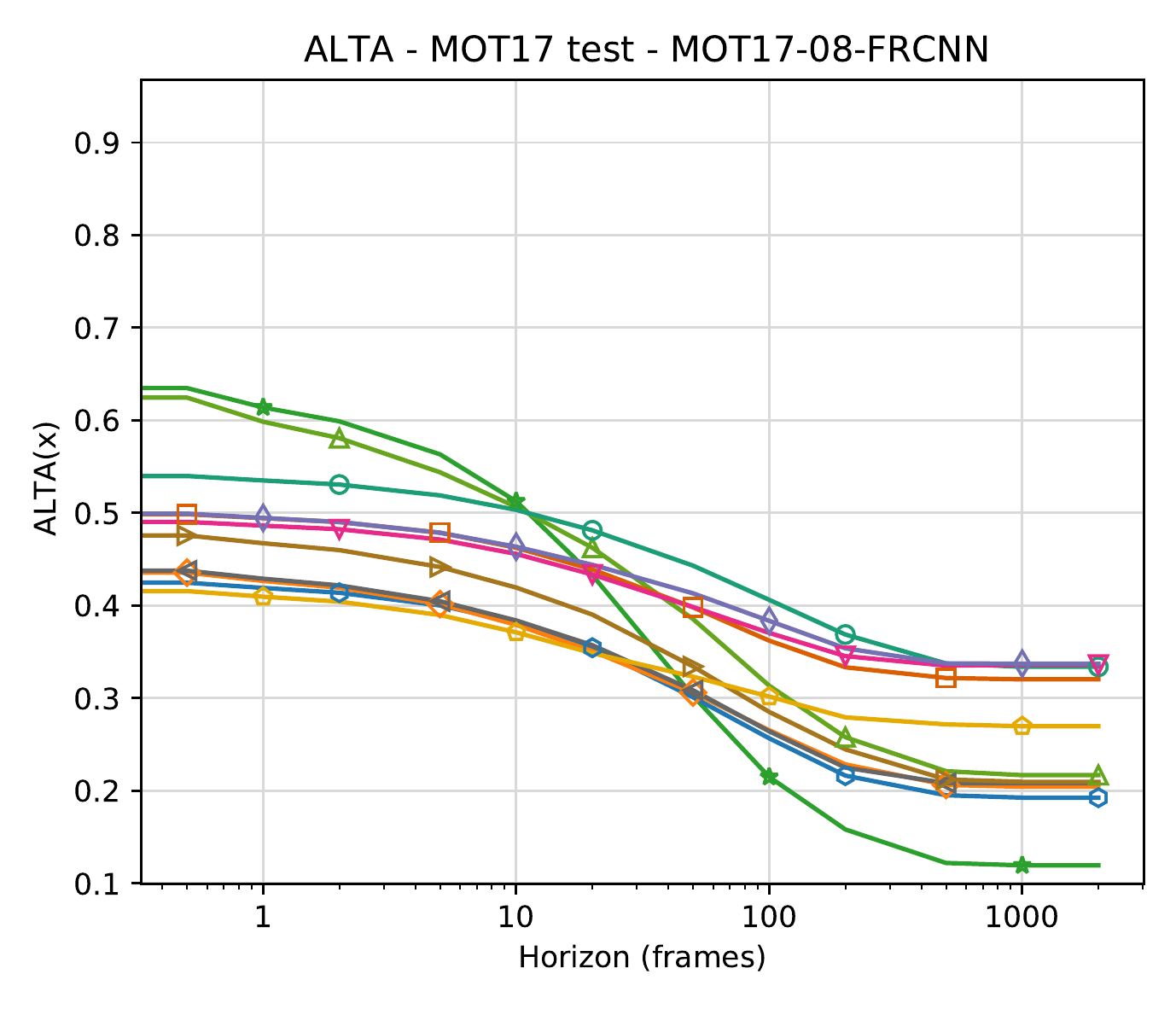}
\includegraphics[width=0.32\textwidth]{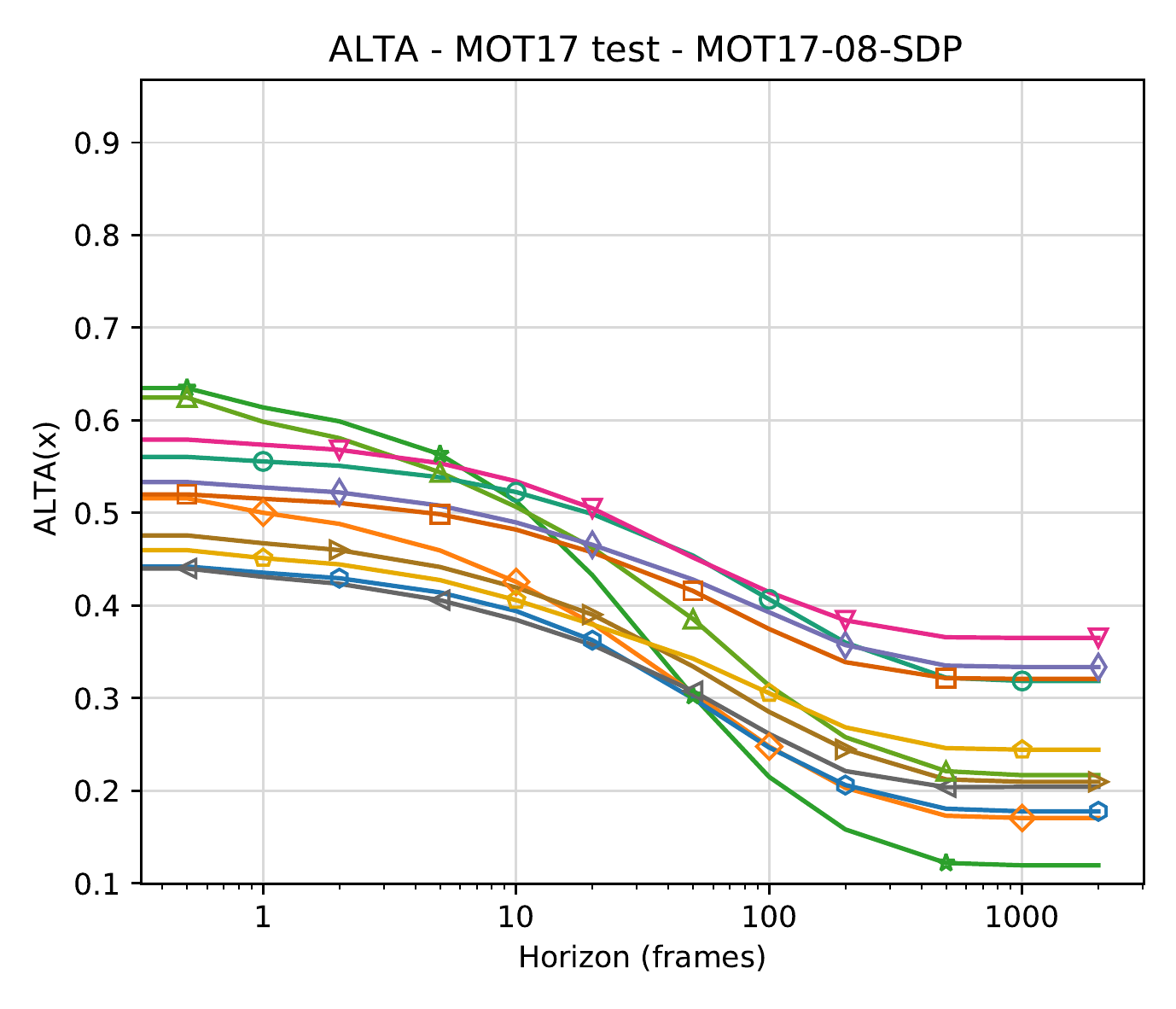}
\\
\includegraphics[width=0.32\textwidth]{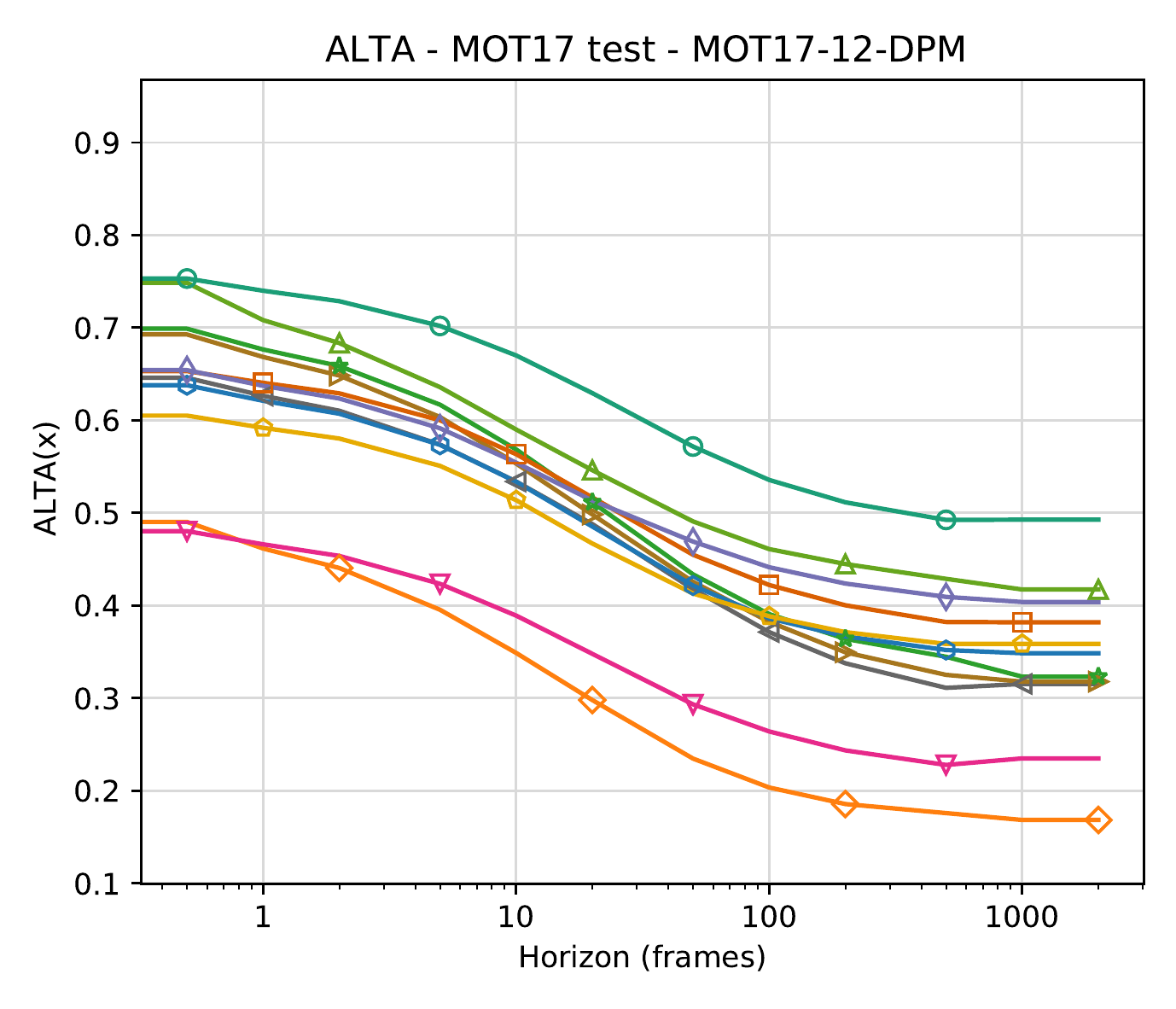}
\includegraphics[width=0.32\textwidth]{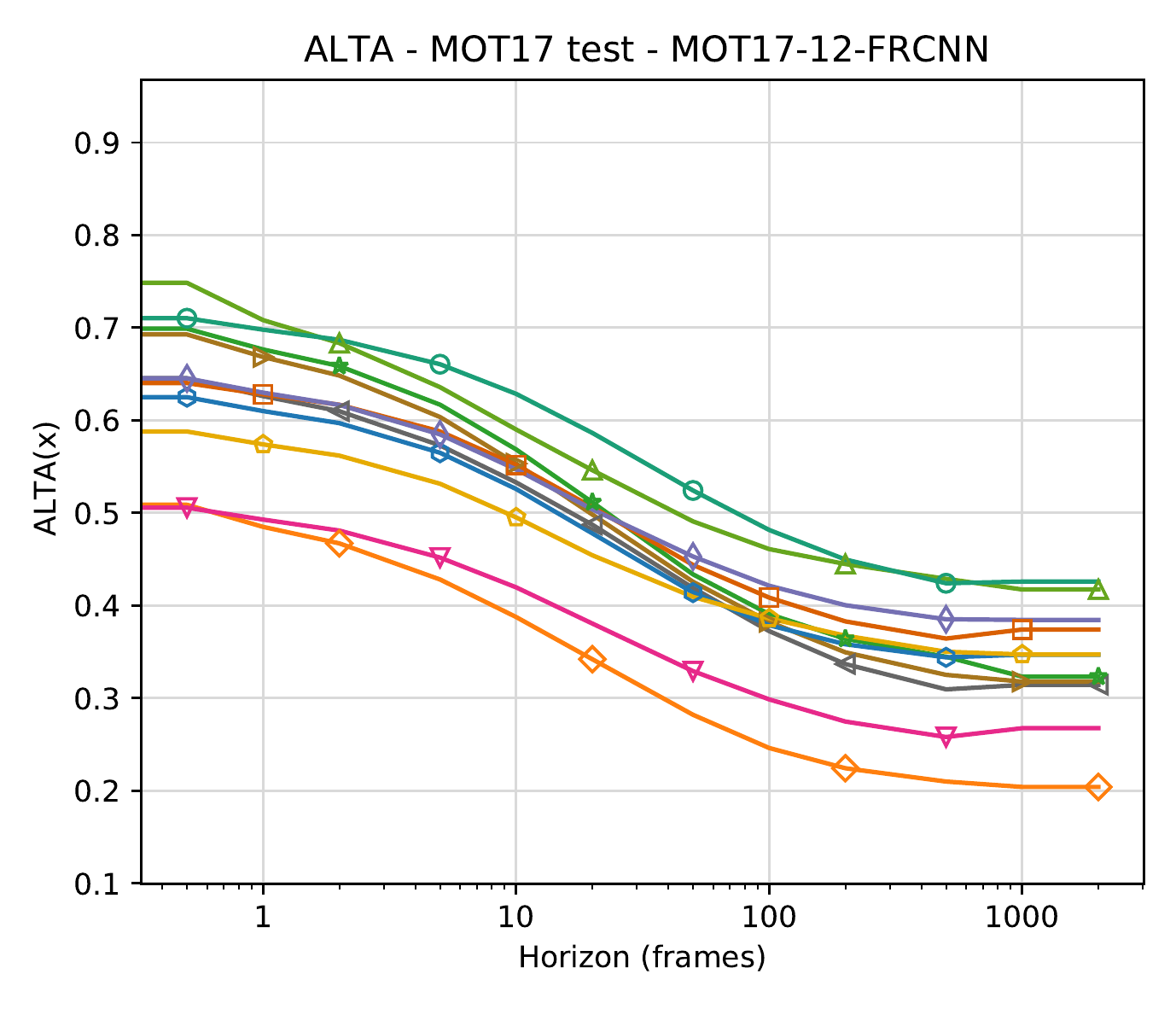}
\includegraphics[width=0.32\textwidth]{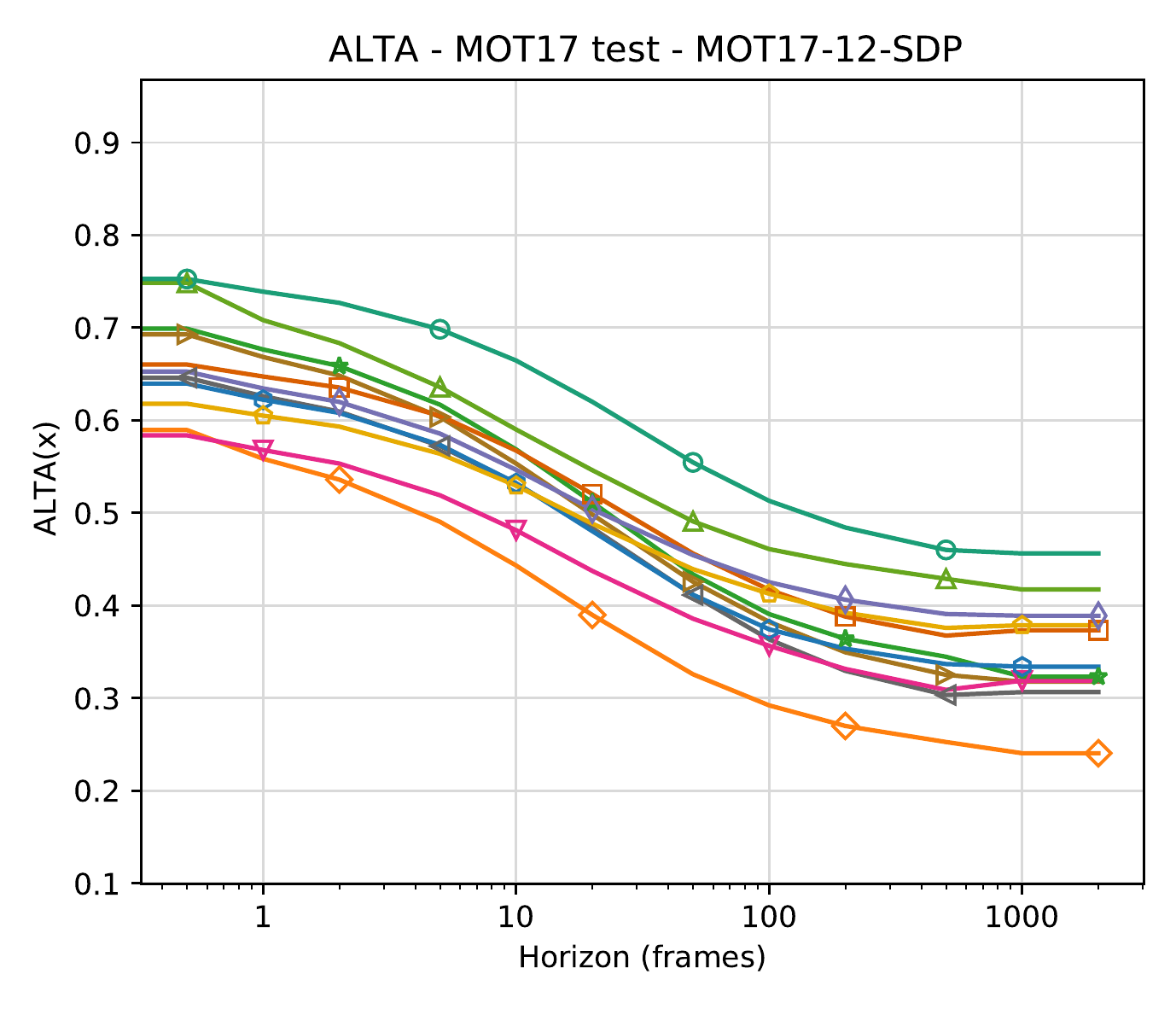}
\\
\includegraphics[width=0.32\textwidth]{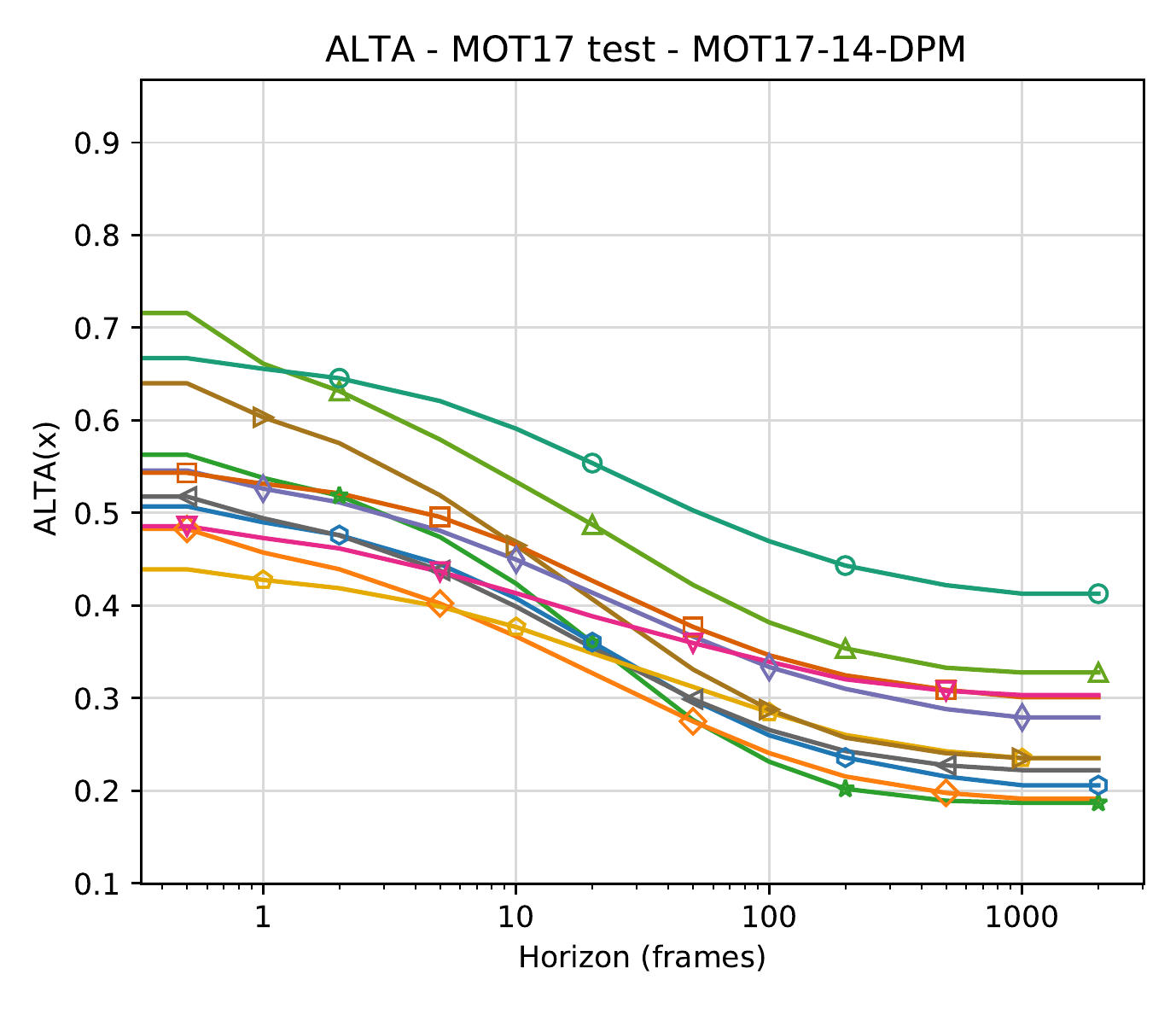}
\includegraphics[width=0.32\textwidth]{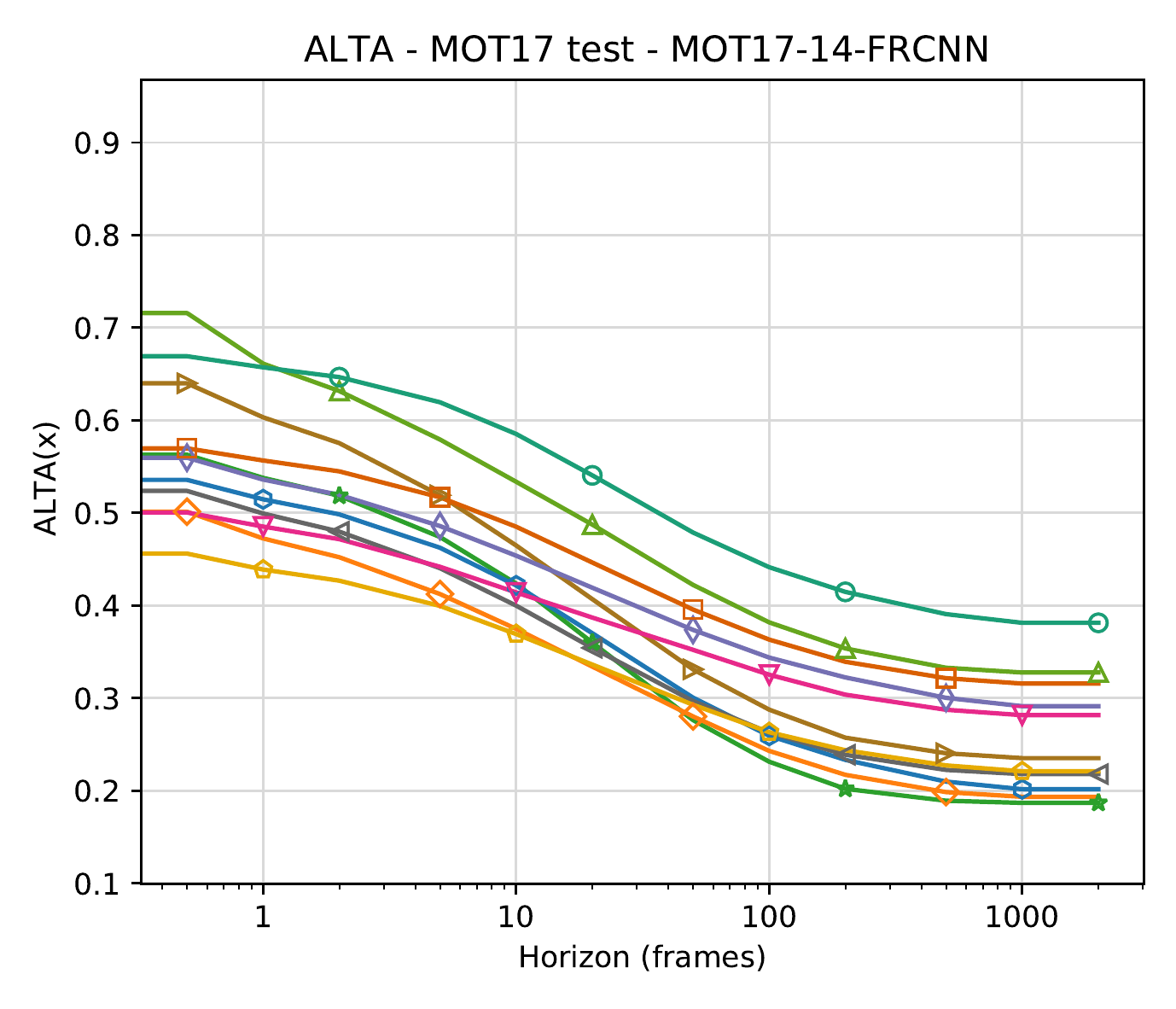}
\includegraphics[width=0.32\textwidth]{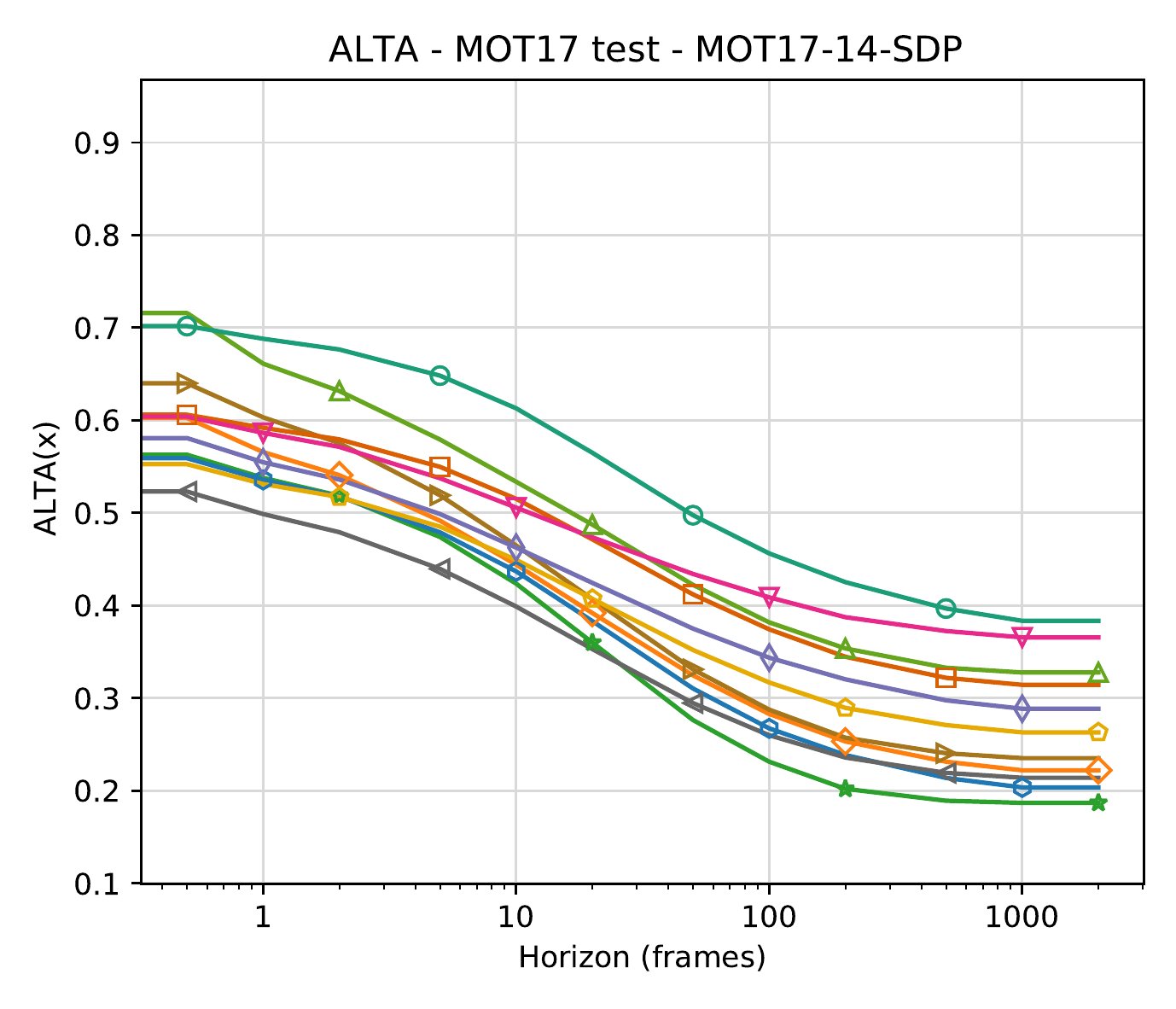}
\caption{ALTA versus horizon for each of the 21 sequences in the MOT 2017 test set.}
\end{figure*}

\clearpage\twocolumn
\section{Extended results: Waymo Open Dataset}

\subsection{Tracker descriptions}

We briefly describe the trackers which were evaluated on the Waymo Open Dataset 2D tracking benchmark.
Of the nine trackers, two were submitted by the original authors of a formal paper (Quasi-Dense R101~\cite{pang2020quasi} and Online V-IOU~\cite{bochinski2017high}), three more were accompanied by a technical report (HorizonMOT~\cite{wang20201st}, CascadeRCNN-SORT v2~\cite{Xu2020}, and DSTNet~\cite{thuync2020dstnet}) and the remaining four provided only a short description.
For these four trackers (denoted by~$\dagger$), we cite the algorithm from the description with the caveat that the submission was not made by its original authors.

\textbf{Quasi-Dense R101}~\cite{pang2020quasi} is a joint detection and embedding approach that associates detections over time without taking into account location or motion.
Association is performed by bidirectional matching and low-scoring region proposals (``backdrops'') are retained during training and testing to make the method ``quasi-dense''.
The detector is a Feature Pyramid Network~\cite{lin2017feature} Faster R-CNN~\cite{ren2015faster} and the embedding is trained with the non-parametric softmax loss~\cite{wu2018unsupervised}.
Unmatched tracks are kept alive for up to ten frames.
\textbf{Online V-IOU} is an online version of the Visual IOU tracker~\cite{bochinski2018extending}, which enhances a simplistic IOU tracker by employing a single-object tracker (KCF~\cite{henriques2014high}) to extend tracks that would otherwise have been terminated.
Detections are obtained using Cascade R-CNN~\cite{cai2019cascade} with a HRNetV2 backbone~\cite{wang2020deep}.
\textbf{HorizonMOT}~\cite{wang20201st} (no relation to our temporal horizon) is a tracking-by-detection framework that uses CenterNet~\cite{zhou2019objects} for detections and includes a deep re-ID framework based on~\cite{Wojke2017simple} to match object appearances via the cosine distance.
A fallback matching mechanism based on IOU is used for detections that could not be matched by appearance. %
\textbf{CascadeRCNN-SORT v2}~\cite{Xu2020} combines a Cascade R-CNN detector~\cite{cai2019cascade} with SORT-based tracking~\cite{bewley2016simple} a grid search was performed to find optimal SORT parameters for each class independently.
The submission named simply \textbf{Sort}$^{\dagger}$ uses the same detection architecture and tracking algorithm.
However, the different characteristics in both our analysis and the official MOTA ranking indicate that different hyper-parameters must have been used.
Yet another submission based on SORT is \textbf{dereyly\_alex}$^{\dagger}$, which obtains detections using an ensemble of two detectors from the MMDetection toolbox~\cite{chen2019mmdetection}.
The final three trackers were submitted by the same set of authors.
\textbf{FCTrack}$^{\dagger}$ uses FairMOT~\cite{zhang2020fairmot}, which is a joint detection and re-ID network based on CenterNet with a DLA-34 backbone~\cite{zhou2019objects}, \textbf{DSTNet}~\cite{thuync2020dstnet} is described as a unified Detection-Segmentation-Tracking approach with a softmax-based cross entropy loss for learning track embeddings, while \textbf{ATSS-Track}$^{\dagger}$ is simply stated to use Adaptive Training Sample Selection~\cite{zhang2020bridging} to train its detector with no additional information.

\subsection{Detailed results}

The main paper presented the ALTA plot for the vehicle class only.
This section presents further results for the pedestrian and cyclist classes, as well as for all classes combined.
While the paper did not consider the problem of how to combine metrics for multiple classes, for these experiments we follow the Waymo Open Dataset in simply combining the different classes as different sequences. %
For each tracker, we selected a per-class score threshold for the detections to maximise the detection $F_{1}$-score (the official evaluation server tries several thresholds and adopts that which maximises MOTA).

\begin{figure*}
\centering
\includegraphics[width=0.49\textwidth]{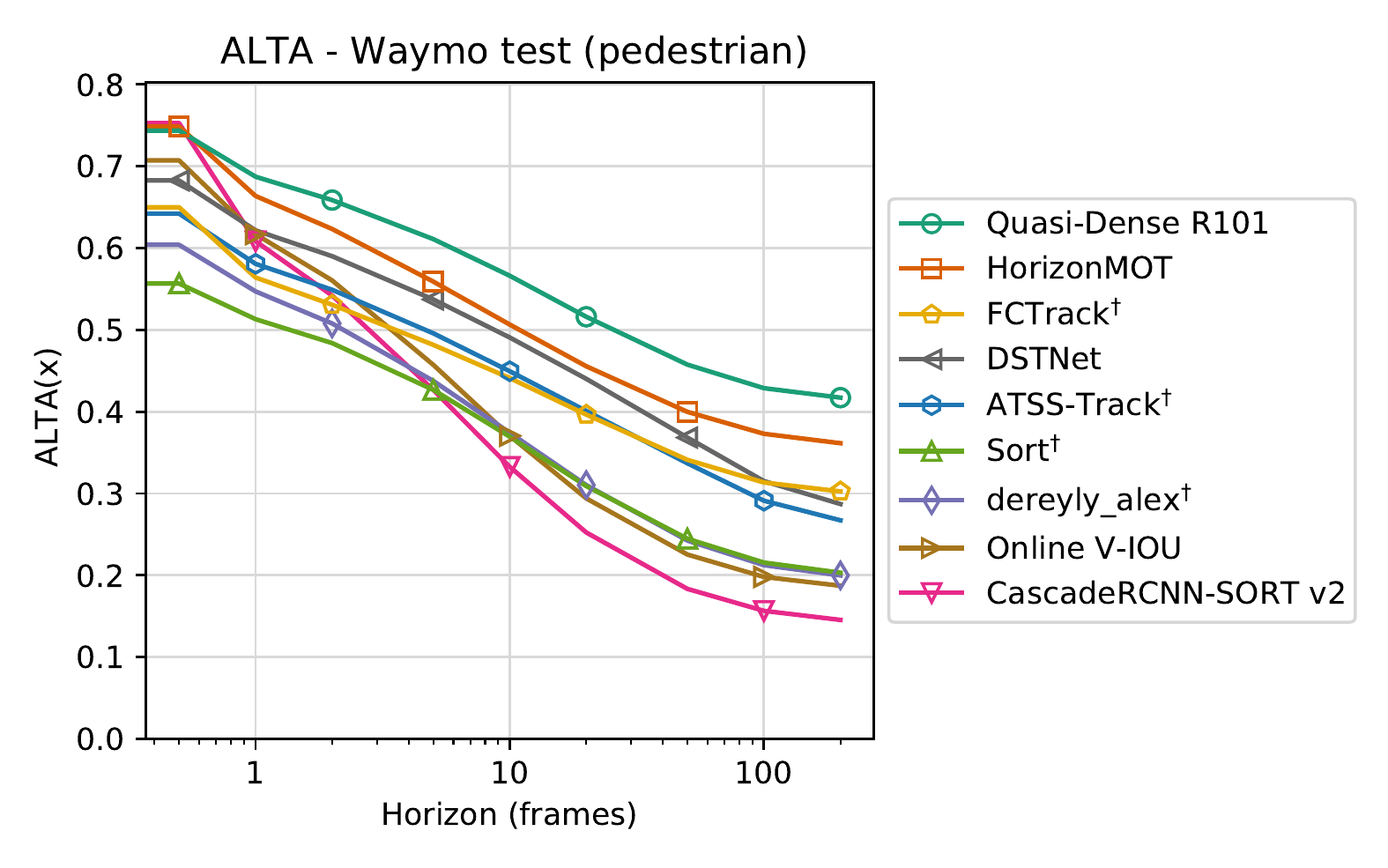}
\includegraphics[width=0.49\textwidth]{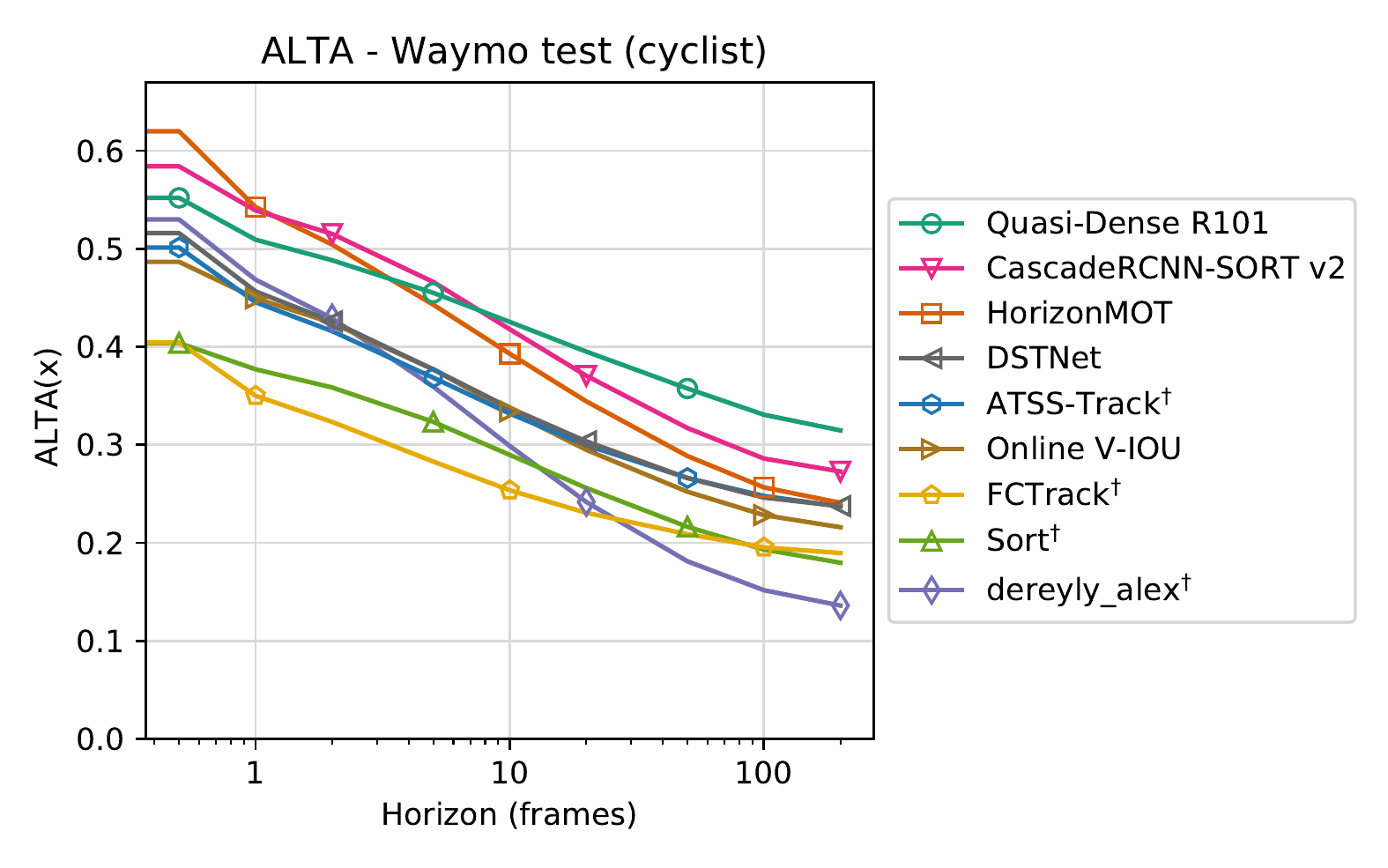}
\caption{ALTA versus horizon for pedestrians and cyclists on the Waymo Open Dataset.}
\label{fig:alta-waymo-cls2-cls4}
\end{figure*}

\begin{figure*}
\centering
\includegraphics[width=0.49\textwidth]{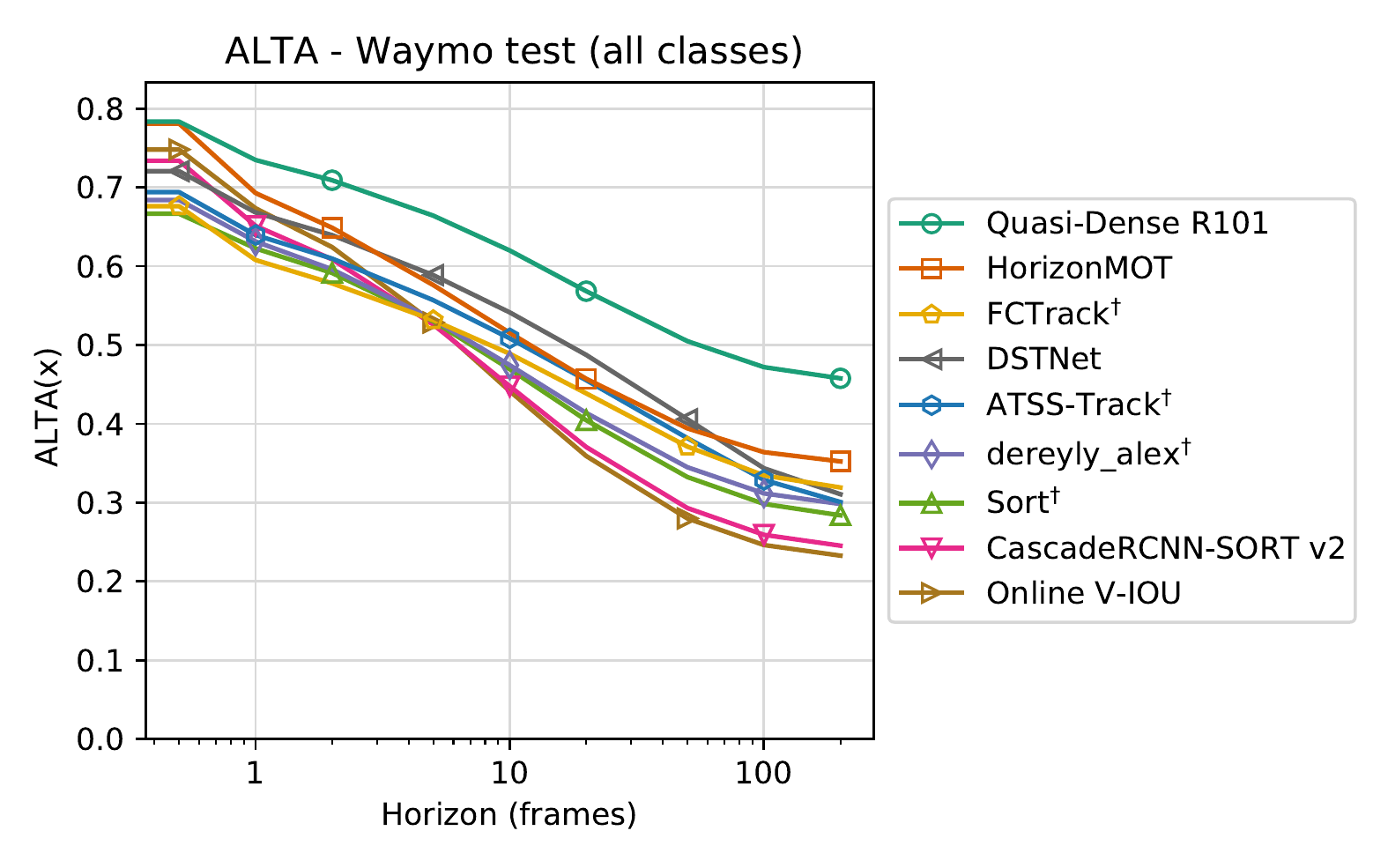}
\includegraphics[width=0.49\textwidth]{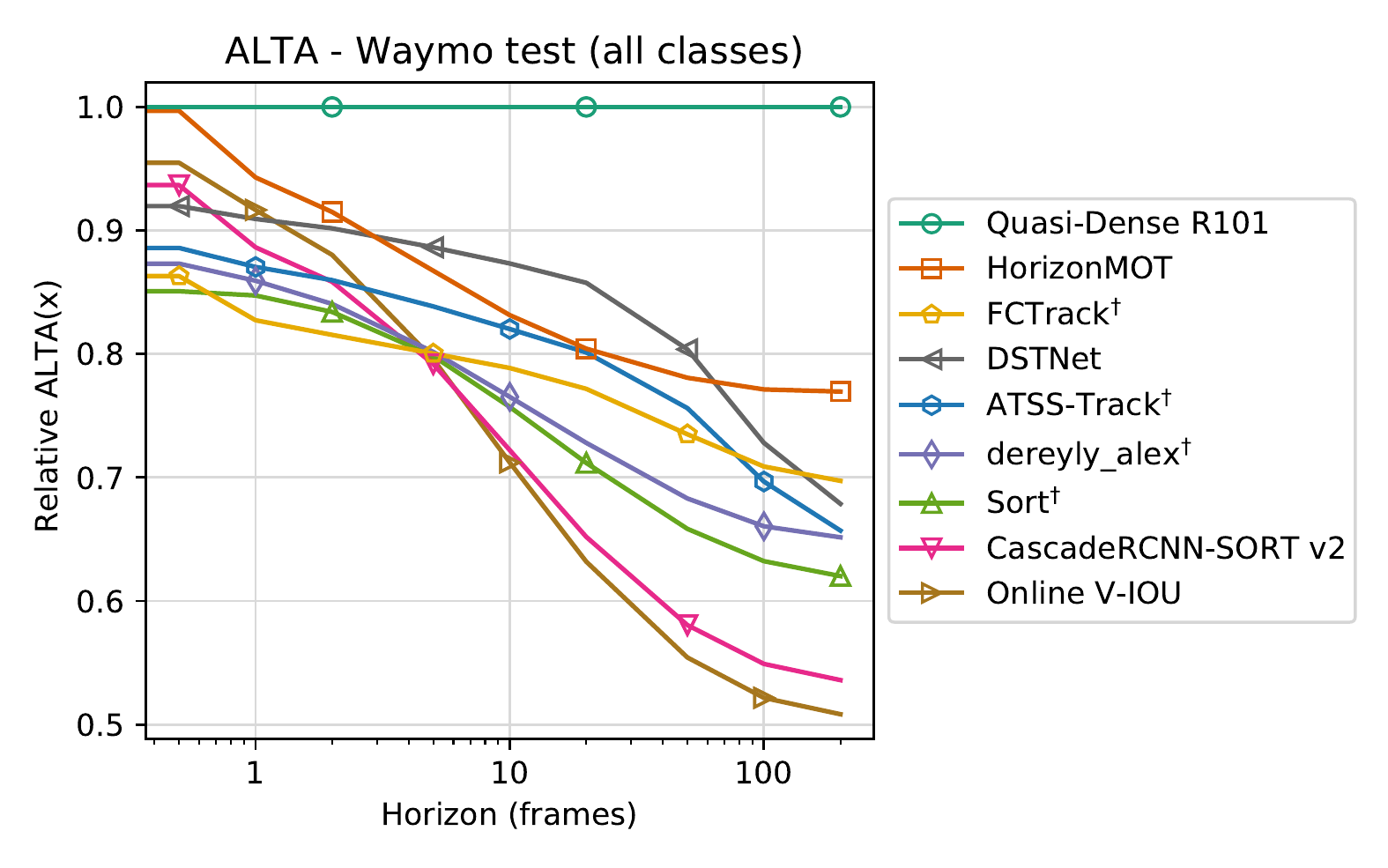} \\
\includegraphics[width=0.49\textwidth]{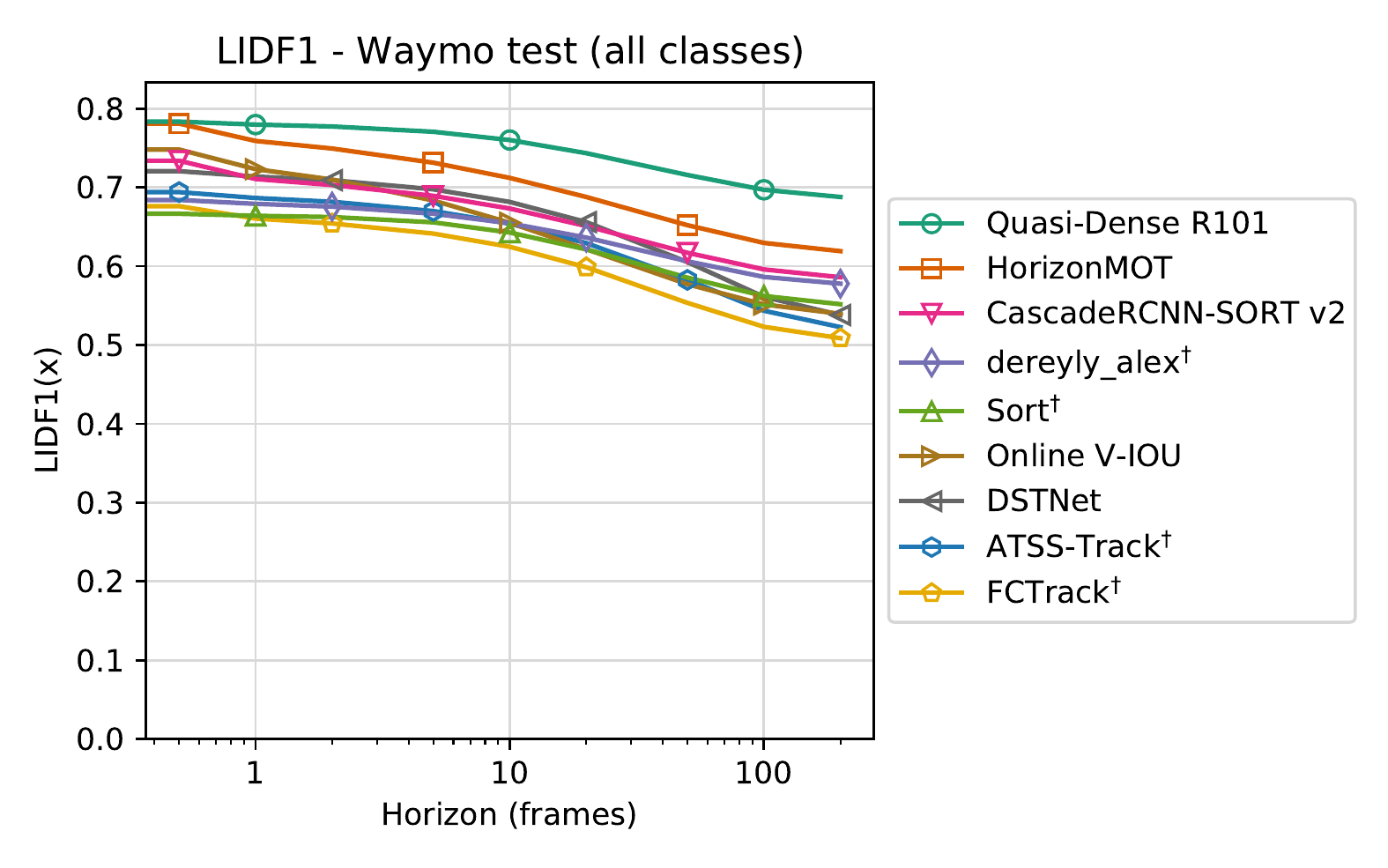}
\includegraphics[width=0.49\textwidth]{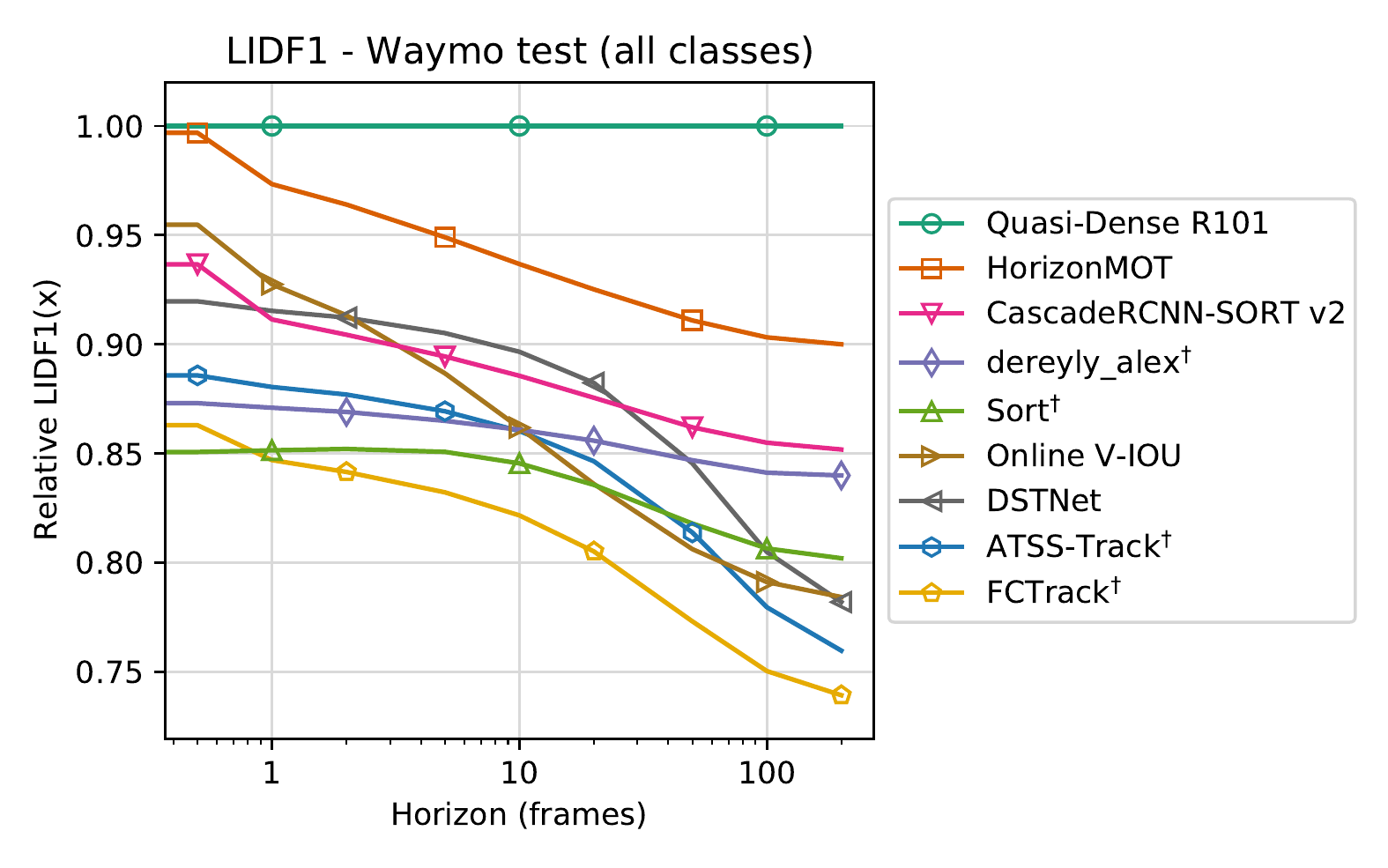}
\caption{
  Comparison of ALTA (top) and LIDF1 (bottom) as a function of horizon for all classes in the Waymo benchmark.
  The two metrics coincide in a detection metric at zero horizon and association has a much greater influence on ALTA than LIDF1.
  The plots on the right show the metric relative to the best tracker at each horizon for easier comparison.
}
\label{fig:waymo-alta-vs-lidf1}
\end{figure*}

\begin{figure*}
\centering
\makebox[\textwidth][c]{
    \includegraphics[width=1.05\columnwidth]{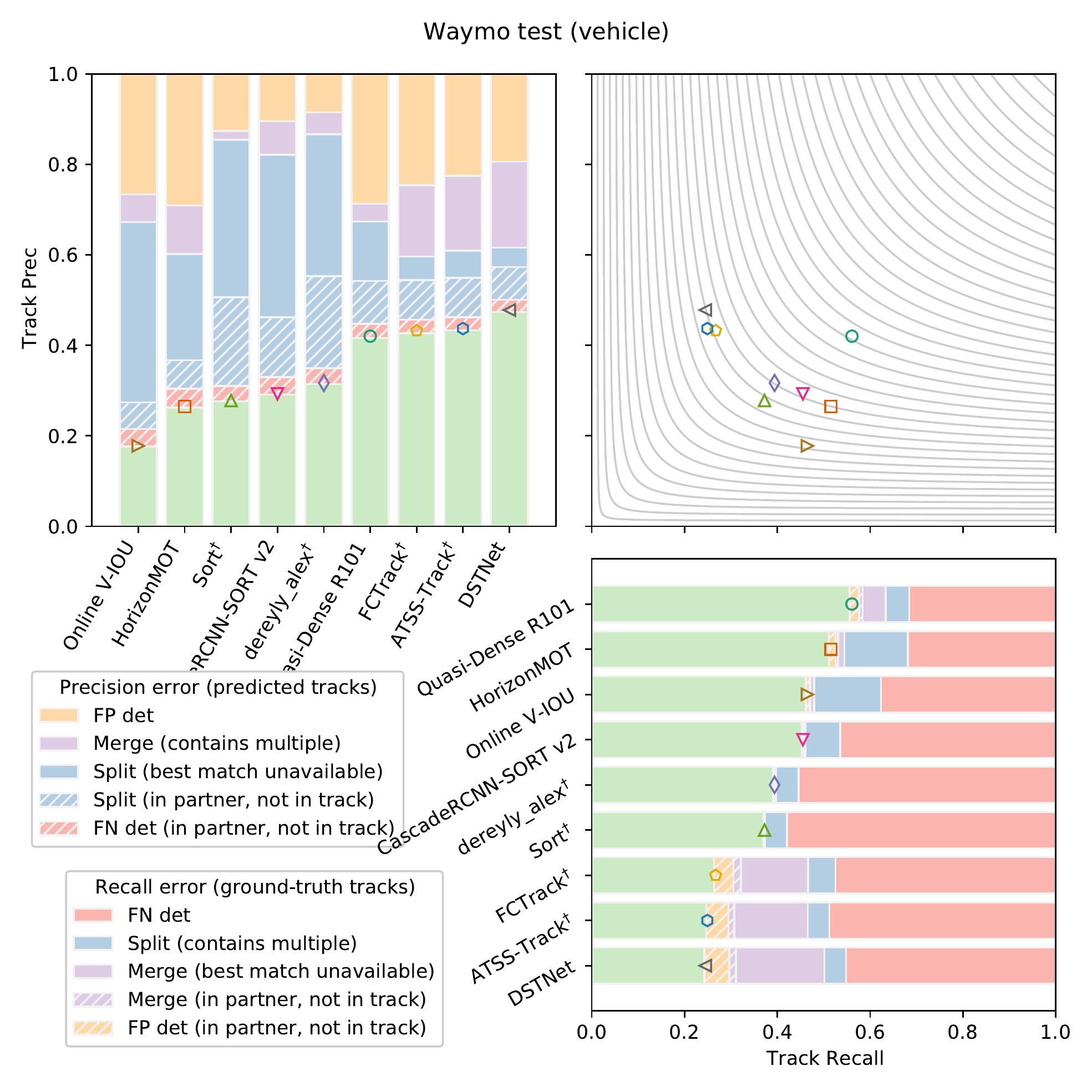}
    \includegraphics[width=1.05\columnwidth]{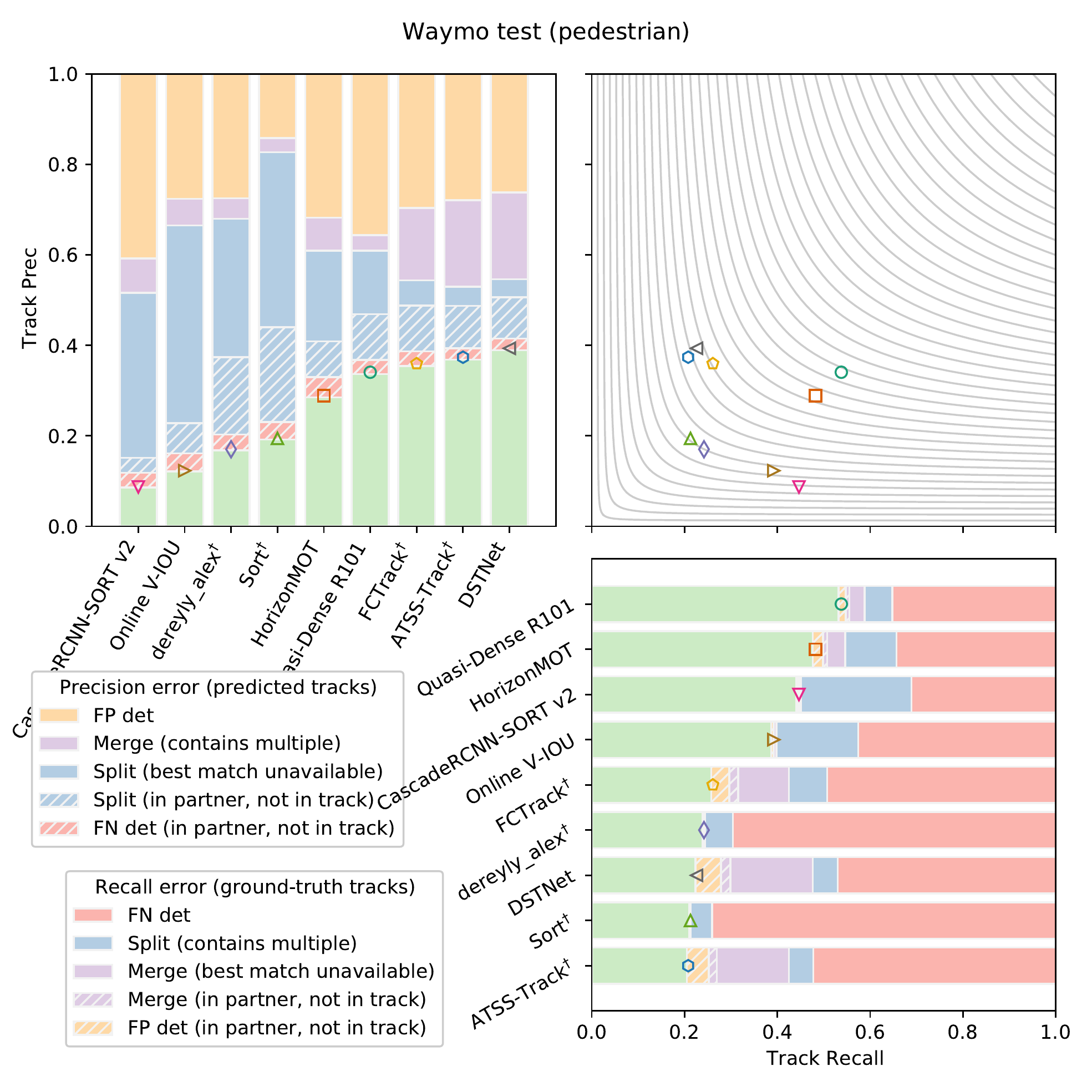}
}
\makebox[\textwidth][c]{
    \includegraphics[width=1.05\columnwidth]{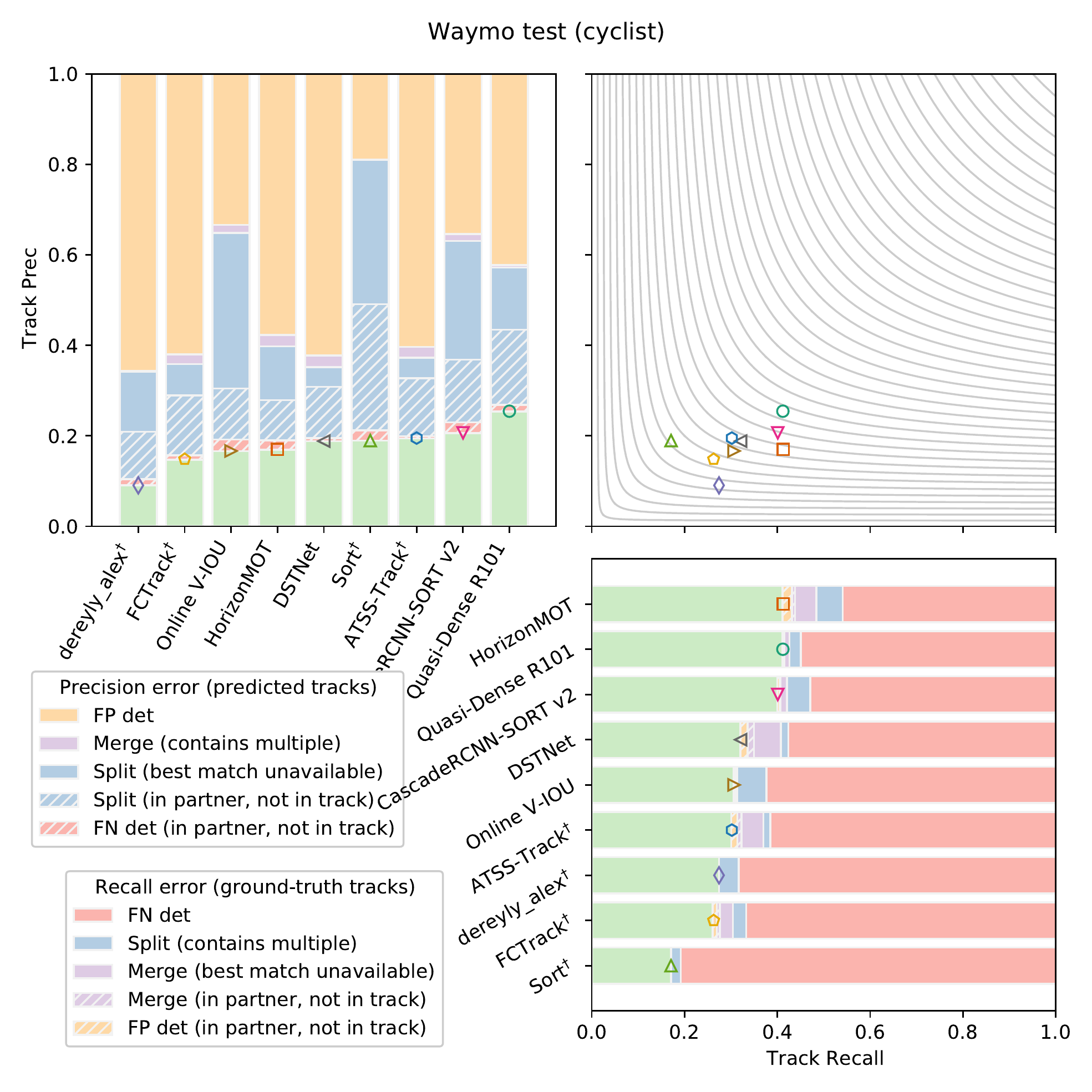}
    \includegraphics[width=1.05\columnwidth]{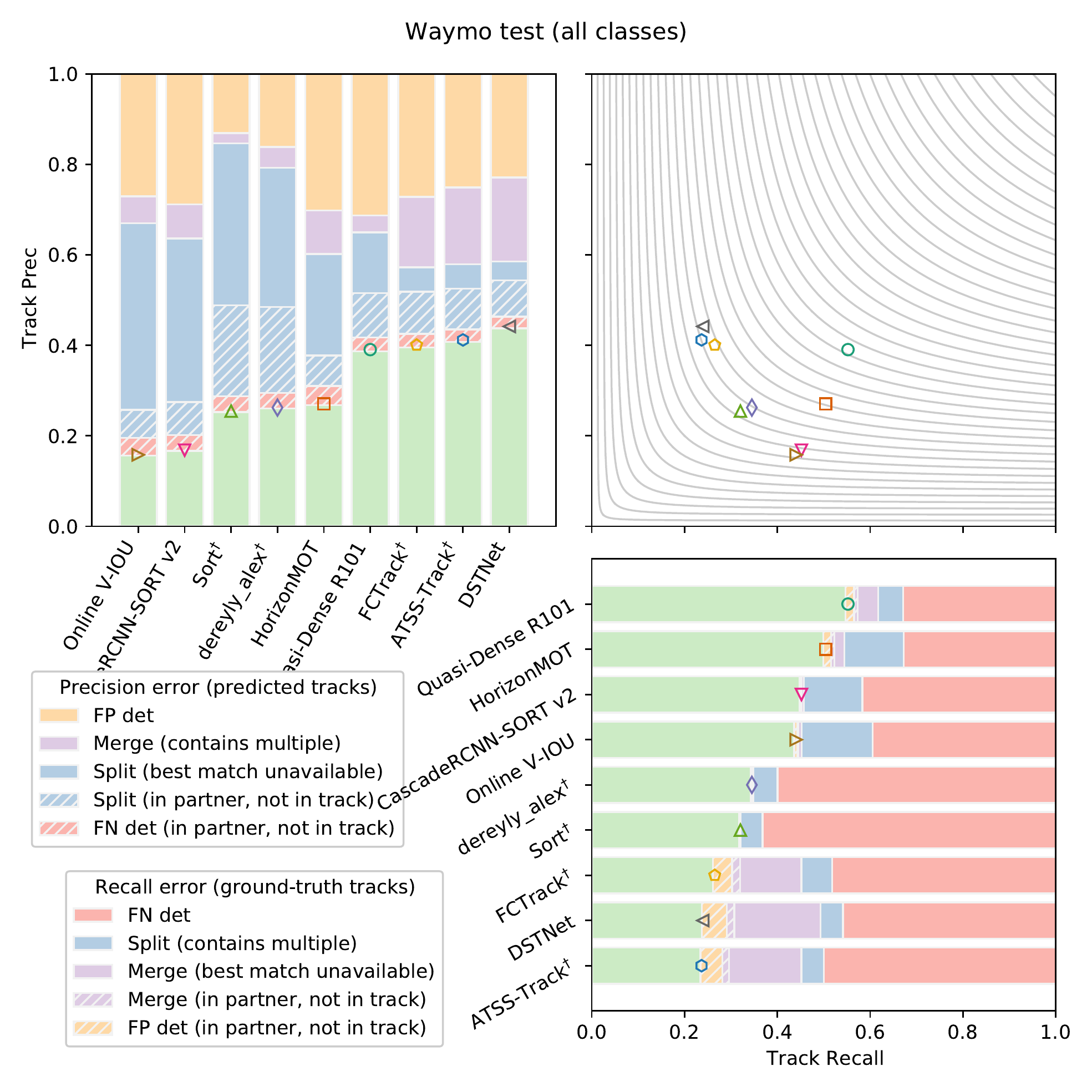}
}
\caption{Decomposition of tracker error types for each class in the Waymo Open Dataset.}
\label{fig:waymo-decompose}
\end{figure*}

Figure~\ref{fig:alta-waymo-cls2-cls4} shows the temporal characterisation of ALTA for the pedestrian and cyclist classes.
Figure~\ref{fig:waymo-alta-vs-lidf1} compares ALTA and LIDF1 for all classes.
Figure~\ref{fig:waymo-decompose} shows the decompositions of track precision and recall for each individual class and all classes combined.
Tables~\ref{tab:waymo-extended-cls1}, \ref{tab:waymo-extended-cls2}, \ref{tab:waymo-extended-cls4} and~\ref{tab:waymo-extended-all} present the full metrics for completeness, including a comparison to MOTA and IDF1.
A few key observations are as follows.
\begin{itemize}[\textendash]
\item As before, ATA is less correlated with detection and more correlated with identity switches compared to MOTA and IDF1.
\item All methods are significantly worse for cyclist than for other classes, both in terms of ATR and ATP, and this is mostly due to detection errors.
\item Trackers that do not use an appearance embedding and focus on association using motion, such as `Online V-IOU' and the three trackers based on SORT, have their track precision severely impacted by split errors.
\item The trackers which are most impacted by merge errors are the joint detection-and-embedding network `FCTrack' and the two other trackers from the same team, `DSTNet' and `ATSS-Track'.
These achieve the best precision and the worst recall.
\item `Quasi-Dense R101', which also uses instance embeddings, achieves higher recall by reducing both false-negative detections and merge errors.
\end{itemize}

It is particularly interesting that `Quasi-Dense R101' achieves the best ATA without making use of spatial location or velocity.
This is the reverse of the MOT17 challenge, where the highest-ranking tracker, MAT, used \emph{only} spatial information to perform association.
This may be due to the relatively low frame-rate and large camera motion in the Waymo dataset, although it is difficult to draw a conclusion without evaluating both methods on the same dataset.

\clearpage

\begin{table*}
\centering
\caption{
  Metric comparison for \emph{vehicle} class on the Waymo 2D tracking benchmark.
}
\scalebox{0.85}{
\begin{tabular}{l r@{~~}l r@{~~}l |  r@{~~}l r@{~~}l r@{~~}l r@{~~}l r@{~~}l}
\toprule
Submission & \multicolumn{2}{c}{MOTA} & \multicolumn{2}{c}{IDF1} & \multicolumn{2}{c}{DetF1} & \multicolumn{2}{c}{ALTA(1s)} & \multicolumn{2}{c}{ALTA(5s)} & \multicolumn{2}{c}{\textbf{ATA}} & \multicolumn{2}{c}{ATA$/$DetF1} \\
\midrule
             Quasi-Dense R101 \cite{pang2020quasi} &  .492 & {\bf (1)} &  .713 & {\bf (1)} & .802 & {\bf (1)} &     .646 &    {\bf (1)} &     .530 &    {\bf (1)} & .481 & {\bf (1)} &       .600 &      {\bf (1)} \\
                 CascadeRCNN-SORT v2 \cite{Xu2020} &  .474 & {\bf (3)} &  .615 & {\bf (4)} & .727 & {\bf (5)} &     .533 &    {\bf (4)} &     .404 &    {\bf (4)} & .356 & {\bf (2)} &       .490 &      {\bf (2)} \\
 dereyly\_alex$^{\dagger}$ \cite{bewley2016simple} &  .413 & {\bf (5)} &  .622 & {\bf (3)} & .717 &       (6) &     .521 &    {\bf (5)} &     .397 &    {\bf (5)} & .351 & {\bf (3)} &       .490 &      {\bf (3)} \\
                     HorizonMOT \cite{wang20201st} &  .472 & {\bf (4)} &  .629 & {\bf (2)} & .796 & {\bf (2)} &     .521 &          (6) &     .393 &          (6) & .350 & {\bf (4)} &       .440 &            (7) \\
       FCTrack$^{\dagger}$ \cite{zhang2020fairmot} &  .373 &       (8) &  .527 &       (9) & .690 &       (9) &     .514 &          (7) &     .389 &          (7) & .330 & {\bf (5)} &       .479 &      {\bf (4)} \\
       DSTNet \cite{thuync2020dstnet} &  .410 &       (6) &  .565 &       (7) & .738 & {\bf (4)} &     .567 &    {\bf (2)} &     .426 &    {\bf (2)} & .324 &       (6) &       .439 &            (8) \\
          Sort$^{\dagger}$ \cite{bewley2016simple} &  .372 &       (9) &  .596 & {\bf (5)} & .708 &       (8) &     .510 &          (8) &     .369 &          (8) & .318 &       (7) &       .450 &      {\bf (5)} \\
   ATSS-Track$^{\dagger}$ \cite{zhang2020bridging} &  .385 &       (7) &  .557 &       (8) & .716 &       (7) &     .536 &    {\bf (3)} &     .404 &    {\bf (3)} & .317 &       (8) &       .443 &            (6) \\
        Online V-IOU \cite{bochinski2018extending} &  .480 & {\bf (2)} &  .571 &       (6) & .767 & {\bf (3)} &     .475 &          (9) &     .309 &          (9) & .257 &       (9) &       .336 &            (9) \\

\bottomrule
\end{tabular}
}
\label{tab:waymo-extended-cls1}
\end{table*}

\begin{table*}
\centering
\caption{
  Metric comparison for \emph{pedestrian} class on the Waymo 2D tracking benchmark.
}
\scalebox{0.85}{
\begin{tabular}{l r@{~~}l r@{~~}l |  r@{~~}l r@{~~}l r@{~~}l r@{~~}l r@{~~}l}
\toprule
Submission & \multicolumn{2}{c}{MOTA} & \multicolumn{2}{c}{IDF1} & \multicolumn{2}{c}{DetF1} & \multicolumn{2}{c}{ALTA(1s)} & \multicolumn{2}{c}{ALTA(5s)} & \multicolumn{2}{c}{\textbf{ATA}} & \multicolumn{2}{c}{ATA$/$DetF1} \\
\midrule
             Quasi-Dense R101 \cite{pang2020quasi} &  .529 & {\bf (1)} &  .628 & {\bf (1)} & .744 & {\bf (3)} &     .566 &    {\bf (1)} &     .458 &    {\bf (1)} & .417 & {\bf (1)} &       .561 &      {\bf (1)} \\
                     HorizonMOT \cite{wang20201st} &  .517 & {\bf (2)} &  .598 & {\bf (2)} & .749 & {\bf (2)} &     .507 &    {\bf (2)} &     .400 &    {\bf (2)} & .361 & {\bf (2)} &       .483 &      {\bf (2)} \\
       FCTrack$^{\dagger}$ \cite{zhang2020fairmot} &  .396 &       (7) &  .465 &       (6) & .650 &       (6) &     .441 &    {\bf (5)} &     .341 &    {\bf (4)} & .303 & {\bf (3)} &       .466 &      {\bf (3)} \\
       DSTNet \cite{thuync2020dstnet} &  .456 & {\bf (5)} &  .471 & {\bf (4)} & .683 & {\bf (5)} &     .491 &    {\bf (3)} &     .368 &    {\bf (3)} & .287 & {\bf (4)} &       .421 &      {\bf (4)} \\
   ATSS-Track$^{\dagger}$ \cite{zhang2020bridging} &  .411 &       (6) &  .436 &       (8) & .642 &       (7) &     .449 &    {\bf (4)} &     .337 &    {\bf (5)} & .267 & {\bf (5)} &       .416 &      {\bf (5)} \\
          Sort$^{\dagger}$ \cite{bewley2016simple} &  .357 &       (9) &  .433 &       (9) & .557 &       (9) &     .369 &          (8) &     .245 &          (6) & .203 &       (6) &       .365 &            (6) \\
 dereyly\_alex$^{\dagger}$ \cite{bewley2016simple} &  .371 &       (8) &  .467 & {\bf (5)} & .604 &       (8) &     .374 &          (6) &     .242 &          (7) & .200 &       (7) &       .331 &            (7) \\
        Online V-IOU \cite{bochinski2018extending} &  .459 & {\bf (4)} &  .461 &       (7) & .707 & {\bf (4)} &     .370 &          (7) &     .225 &          (8) & .187 &       (8) &       .265 &            (8) \\
                 CascadeRCNN-SORT v2 \cite{Xu2020} &  .485 & {\bf (3)} &  .527 & {\bf (3)} & .753 & {\bf (1)} &     .333 &          (9) &     .183 &          (9) & .145 &       (9) &       .193 &            (9) \\

\bottomrule
\end{tabular}
}
\label{tab:waymo-extended-cls2}
\end{table*}

\begin{table*}
\centering
\caption{
  Metric comparison for \emph{cyclist} class on the Waymo 2D tracking benchmark.
}
\scalebox{0.85}{
\begin{tabular}{l r@{~~}l r@{~~}l |  r@{~~}l r@{~~}l r@{~~}l r@{~~}l r@{~~}l}
\toprule
Submission & \multicolumn{2}{c}{MOTA} & \multicolumn{2}{c}{IDF1} & \multicolumn{2}{c}{DetF1} & \multicolumn{2}{c}{ALTA(1s)} & \multicolumn{2}{c}{ALTA(5s)} & \multicolumn{2}{c}{\textbf{ATA}} & \multicolumn{2}{c}{ATA$/$DetF1} \\
\midrule
             Quasi-Dense R101 \cite{pang2020quasi} &  .332 & {\bf (3)} &  .501 & {\bf (2)} & .552 & {\bf (3)} &     .425 &    {\bf (1)} &     .357 &    {\bf (1)} & .315 & {\bf (1)} &       .570 &      {\bf (1)} \\
                 CascadeRCNN-SORT v2 \cite{Xu2020} &  .365 & {\bf (1)} &  .481 & {\bf (3)} & .584 & {\bf (2)} &     .418 &    {\bf (2)} &     .317 &    {\bf (2)} & .273 & {\bf (2)} &       .467 &      {\bf (4)} \\
                     HorizonMOT \cite{wang20201st} &  .365 & {\bf (2)} &  .507 & {\bf (1)} & .620 & {\bf (1)} &     .393 &    {\bf (3)} &     .289 &    {\bf (3)} & .241 & {\bf (3)} &       .389 &            (8) \\
       DSTNet \cite{thuync2020dstnet} &  .274 & {\bf (4)} &  .450 & {\bf (5)} & .516 & {\bf (5)} &     .337 &    {\bf (4)} &     .266 &    {\bf (5)} & .237 & {\bf (4)} &       .460 &      {\bf (5)} \\
   ATSS-Track$^{\dagger}$ \cite{zhang2020bridging} &  .272 & {\bf (5)} &  .448 &       (6) & .501 &       (6) &     .332 &          (6) &     .266 &    {\bf (4)} & .237 & {\bf (5)} &       .473 &      {\bf (2)} \\
        Online V-IOU \cite{bochinski2018extending} &  .264 &       (6) &  .385 &       (7) & .487 &       (7) &     .334 &    {\bf (5)} &     .252 &          (6) & .216 &       (6) &       .444 &            (7) \\
       FCTrack$^{\dagger}$ \cite{zhang2020fairmot} &  .182 &       (9) &  .351 &       (9) & .404 &       (8) &     .254 &          (9) &     .209 &          (8) & .190 &       (7) &       .469 &      {\bf (3)} \\
          Sort$^{\dagger}$ \cite{bewley2016simple} &  .230 &       (7) &  .351 &       (8) & .404 &       (9) &     .290 &          (8) &     .216 &          (7) & .180 &       (8) &       .445 &            (6) \\
 dereyly\_alex$^{\dagger}$ \cite{bewley2016simple} &  .227 &       (8) &  .451 & {\bf (4)} & .530 & {\bf (4)} &     .299 &          (7) &     .181 &          (9) & .136 &       (9) &       .257 &            (9) \\

\bottomrule
\end{tabular}
}
\label{tab:waymo-extended-cls4}
\end{table*}

\begin{table*}
\centering
\caption{
  Combined results for all classes on the Waymo 2D tracking benchmark.
}
\scalebox{0.85}{
\begin{tabular}{l r@{~~}l r@{~~}l |  r@{~~}l r@{~~}l r@{~~}l r@{~~}l r@{~~}l}
\toprule
Submission & \multicolumn{2}{c}{MOTA} & \multicolumn{2}{c}{IDF1} & \multicolumn{2}{c}{DetF1} & \multicolumn{2}{c}{ALTA(1s)} & \multicolumn{2}{c}{ALTA(5s)} & \multicolumn{2}{c}{\textbf{ATA}} & \multicolumn{2}{c}{ATA$/$DetF1} \\
\midrule
             Quasi-Dense R101 \cite{pang2020quasi} &  .451 & {\bf (2)} &  .688 & {\bf (1)} & .784 & {\bf (1)} &     .620 &    {\bf (1)} &     .505 &    {\bf (1)} & .458 & {\bf (1)} &       .584 &      {\bf (1)} \\
                     HorizonMOT \cite{wang20201st} &  .451 & {\bf (1)} &  .619 & {\bf (2)} & .781 & {\bf (2)} &     .515 &    {\bf (3)} &     .394 &    {\bf (3)} & .352 & {\bf (2)} &       .451 &      {\bf (3)} \\
       FCTrack$^{\dagger}$ \cite{zhang2020fairmot} &  .317 &       (9) &  .509 &       (9) & .676 &       (8) &     .489 &    {\bf (5)} &     .371 &    {\bf (5)} & .319 & {\bf (3)} &       .472 &      {\bf (2)} \\
       DSTNet \cite{thuync2020dstnet} &  .380 & {\bf (5)} &  .538 &       (7) & .721 & {\bf (5)} &     .541 &    {\bf (2)} &     .406 &    {\bf (2)} & .311 & {\bf (4)} &       .431 &            (6) \\
   ATSS-Track$^{\dagger}$ \cite{zhang2020bridging} &  .356 &       (6) &  .523 &       (8) & .694 &       (6) &     .508 &    {\bf (4)} &     .382 &    {\bf (4)} & .301 & {\bf (5)} &       .433 &      {\bf (5)} \\
 dereyly\_alex$^{\dagger}$ \cite{bewley2016simple} &  .337 &       (7) &  .578 & {\bf (4)} & .684 &       (7) &     .474 &          (6) &     .345 &          (6) & .298 &       (6) &       .436 &      {\bf (4)} \\
          Sort$^{\dagger}$ \cite{bewley2016simple} &  .320 &       (8) &  .552 & {\bf (5)} & .667 &       (9) &     .469 &          (7) &     .332 &          (7) & .284 &       (7) &       .426 &            (7) \\
                 CascadeRCNN-SORT v2 \cite{Xu2020} &  .442 & {\bf (3)} &  .586 & {\bf (3)} & .734 & {\bf (4)} &     .448 &          (8) &     .293 &          (8) & .245 &       (8) &       .334 &            (8) \\
        Online V-IOU \cite{bochinski2018extending} &  .401 & {\bf (4)} &  .540 &       (6) & .748 & {\bf (3)} &     .441 &          (9) &     .280 &          (9) & .233 &       (9) &       .311 &            (9) \\

\bottomrule
\end{tabular}
}
\label{tab:waymo-extended-all}
\end{table*}

\end{document}